 \documentclass[pmlr,twocolumn,10pt]{jmlr} 

\usepackage{booktabs}
\usepackage{siunitx}

\usepackage[switch]{lineno}

\newcommand{\equal}[1]{{\hypersetup{linkcolor=black}\thanks{#1}}}

\theorembodyfont{\upshape}
\theoremheaderfont{\scshape}
\theorempostheader{:}
\theoremsep{\newline}

\usepackage{tabularray}
\usepackage{seqsplit}
\usepackage{bbm} 
\DeclareMathOperator*{\concat}{%
    \mathchoice%
        {\Big\Vert}%
        {\big\Vert}%
        {\Vert}%
        {\Vert}%
}
\usepackage{comment}
\usepackage{setspace}
\SetKwInput{KwInit}{Initialize}
\newcommand{\mc}[1]{\multicolumn{1}{c}{#1}}

\jmlrvolume{297}
\jmlryear{2025}
\jmlrworkshop{Machine Learning for Health (ML4H) 2025} 

 \title[Data-Driven Discovery of Feature Groups in Clinical Time Series]{Data-Driven Discovery of Feature Groups in Clinical Time Series}

\newif\ifanonymous

\newcommand{\textns}[1]{\textsuperscript{\textnormal{#1}}}

\ifanonymous
    \author{%
        \Name{Anonymous Authors} \Email{anonymous.authors@email.com}\\
        \addr Address \\
    }
    
    \newcommand{\printacks}{}
\else
    \author{%
        \Name{Fedor Sergeev}\textns{1,2,}\equal{Joint-corresponding authors} \Email{fedor.sergeev@inf.ethz.ch}\\
        \Name{Manuel Burger}\textns{1} \\
        \Name{Polina Leshetkina}\textns{3}\\
        \Name{Vincent Fortuin}\textns{2,4,5} \\
        \Name{Gunnar R\"atsch}\textns{1,6,7,8}  \\
        \Name{Rita Kuznetsova}\textns{1,}\footnotemark[1] \Email{rita.kuznetsova@inf.ethz.ch}\medskip\\
        \addr \textns{1}Department of Computer Science, ETH Zurich, Switzerland\\
        \addr \textns{2}Helmholtz AI, Germany \\
        \addr \textns{3}Department of Health Science and Medicine, University of Lucerne, Switzerland \\
        \addr \textns{4}TU Munich, Germany \\
        \addr \textns{5}Munich Center for Machine Learning, Germany \\
        \addr \textns{6}AI Center, ETH Zurich, Switzerland\\
        \addr \textns{7}Swiss Institute for Bioinformatics, Zurich, Switzerland\\
        \addr \textns{8}Medical Informatics Unit, University Hospital Zurich, Switzerland
    }

    \newcommand{\printacks}{\acks{We thank the anonymous reviewers for their helpful suggestions for our paper. We thank researchers, patients, and clinical staff for creating and sharing the medical datasets used in this work. Computational data analysis was performed at Leonhard Med ({\small \url{https://sis.id.ethz.ch/services/sensitiveresearchdata/}}), a secure, trusted research environment at ETH Zurich. Fedor Sergeev was supported by grant \#902 of the Strategic Focus Area ``Personalized Health and Related Technologies (PHRT)'' of the ETH Domain (Swiss Federal Institutes of Technology). Vincent Fortuin was supported by a Branco Weiss Fellowship.}}
\fi

\usepackage[normalem]{ulem}
\usepackage[svgnames]{xcolor}
\usepackage{soul}

\newif\ifhighlight
\highlightfalse 

\definecolor{tabBlue}{RGB}{31, 119, 180}
\definecolor{tabOrange}{RGB}{255, 127, 14}
\definecolor{tabGreen}{RGB}{44, 160, 44}
\definecolor{tabPurple}{RGB}{148, 103, 189}
\definecolor{tabPink}{RGB}{227, 119, 194}    

\ifhighlight
    \colorlet{revAback}{tabBlue!20}\colorlet{revAtext}{tabBlue!75!black}
    \newcommand{\revA}[1]{{\color{revAtext}\hlc[revAback]{#1}}}
    \colorlet{revBback}{tabOrange!20}\colorlet{revBtext}{tabOrange!80!black}
    \newcommand{\revB}[1]{{\color{revBtext}\hlc[revBback]{#1}}}
    \colorlet{revCback}{tabGreen!20}\colorlet{revCtext}{tabGreen!70!black}
    \newcommand{\revC}[1]{{\color{revCtext}\hlc[revCback]{#1}}}
    \colorlet{revDback}{tabPurple!20}\colorlet{revDtext}{tabPurple!70!black}
    \newcommand{\revD}[1]{{\color{revDtext}\hlc[revDback]{#1}}}
    \colorlet{revEback}{tabPink!20}\colorlet{revEtext}{tabPink!70!black}
    \newcommand{\revE}[1]{{\color{revEtext}\hlc[revEback]{#1}}}

    \newcommand{\revAdel}[1]{{\color{revAtext}\sout{#1}}}
    \newcommand{\revBdel}[1]{{\color{revBtext}\sout{#1}}}
    \newcommand{\revCdel}[1]{{\color{revCtext}\sout{#1}}}
    \newcommand{\revDdel}[1]{{\color{revDtext}\sout{#1}}}
    \newcommand{\revEdel}[1]{{\color{revEtext}\sout{#1}}}

\else
    \newcommand{\revA}[1]{#1}
    \newcommand{\revB}[1]{#1}
    \newcommand{\revC}[1]{#1}
    \newcommand{\revD}[1]{#1}
    \newcommand{\revE}[1]{#1}

    \newcommand{\revAdel}[1]{}
    \newcommand{\revBdel}[1]{}
    \newcommand{\revCdel}[1]{}
    \newcommand{\revDdel}[1]{}
    \newcommand{\revEdel}[1]{}
\fi

\begin{document}

\maketitle

\begin{abstract}
    Clinical time series data are critical for patient monitoring and predictive modeling. These time series are typically multivariate and often comprise hundreds of heterogeneous features from different data sources. The grouping of features based on similarity and relevance to the prediction task has been shown to enhance the performance of deep learning architectures. However, defining these groups a priori using only semantic knowledge is challenging, even for domain experts. To address this, we propose a novel method that learns feature groups by clustering weights of feature-wise embedding layers. This approach seamlessly integrates into standard supervised training and discovers the groups that directly improve downstream performance on clinically relevant tasks. We demonstrate that our method outperforms static clustering approaches on synthetic data and achieves performance comparable to expert-defined groups on real-world medical data. Moreover, the learned feature groups are clinically interpretable, enabling data-driven discovery of task-relevant relationships between variables.
\end{abstract}
\begin{keywords}
Deep Learning, Healthcare, Time Series, Feature Groups, Clustering.
\end{keywords}

\paragraph*{Data and Code Availability}
    We use the MIMIC-III \citep{MIMIC-III} and HiRID datasets \citep{hirid}, which are freely accessible via the \href{https://physionet.org/}{physionet.org} platform. We share our code at \href{https://github.com/ratschlab/feature-group-discovery}{github.com/ratschlab/feature-group-discovery}.

    \begin{figure*}[tb]
    \floatconts
      {fig:scheme}
      {\caption{The proposed step-wise embedding model with learned feature groups. Filled and non-filled boxes represent vectors and learnable functions, respectively.}}
      {\includegraphics[width=\linewidth,page=9,trim={0 5.23cm 0 0},clip]{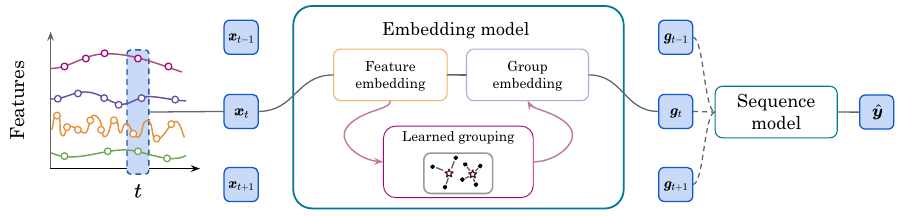}}
    \end{figure*}

\paragraph*{Institutional Review Board (IRB)}
    This research does not require IRB approval.

\section{Introduction}
\label{sec:intro}

    In medical research, considering variables through predefined semantic groups rather than individually often helps to reduce data heterogeneity and capture the interactions between features~\citep{kelly2009multiple, meira2001cancer}. This approach is especially relevant for Electronic Health Record (EHR) data, which consists of multivariate time series that describe complex and dynamic patient trajectories. Previous research~\citep{kuznetsova2023importance} has shown that grouping EHR variables by their measurement type (vitals, labs, treatments) and organ systems helps improve downstream performance and explainability of deep learning models.
    
    However, when the time series contains dozens or even hundreds of variables, manually defining feature groups that are medically meaningful and improve model performance becomes a challenge even for medical professionals. To address this challenge, we propose a novel algorithm that learns feature groups in a data-driven manner to directly improve model performance during supervised training. Specifically, we focus on the \emph{global} feature groups (the same across all patients and time points)~\citep{chormunge2018correlation}. These, unlike the sample-wise groups~\citep{masoomi2020}, are specific to each task and dataset.
    
    We build directly on previous work that has improved deep learning architectures on EHR data~\citep{tomavsev2019clinically,tomavsev2021use,yeche2021,kuznetsova2023importance} and tabular data ~\citep{gorishniy2022embeddings,gorishniy2023revisitingdeeplearning}. Specifically, we use an expressive step-wise embedding approach that has shown excellent performance on tabular and time series data by encoding features with neural networks before applying a backbone model. This approach can also be naturally extended to embed features by groups that are defined a priori by domain experts, further improving the performance on EHR data~\citep{kuznetsova2023importance}. \revE{It is based on learning embeddings first for each feature, then for each group, combining group embeddings, and passing them to the sequence model.}
    
    We further improve this approach by enabling the model to learn global feature groups during normal supervised training (see \figureref{fig:scheme}). Specifically, we characterize each feature by the corresponding feature-wise embedding weights of the step-wise embedding model and cluster them using a centroid-based algorithm. The feature-wise embedding weights are consistent across samples and time steps, guaranteeing that they encode global information about the variables. The weights are updated through gradient descent, which means that the clustering can be seamlessly integrated into the normal training loop. To the best of our knowledge, this is the first work to consider and demonstrate the impact of globally learned feature groups on prediction performance for EHR data.
    
    \paragraph{Contributions} (1) We propose a novel method for learning feature groups in time series data by clustering weights of feature-wise embeddings. (2) We show that our method recovers ground-truth feature groups on synthetic data and matches the performance of state-of-the-art models on real Intensive Care Unit (ICU) data without relying on predefined feature groups. (3) We demonstrate that the learned feature groups are novel and interpretable by medical experts, enabling data-driven discovery of task-relevant relationships between clinical variables.
    
\section{Method}
\label{sec:method}

    We consider a supervised learning problem: given a time series $\boldsymbol{x}$, the task is to predict a label $\boldsymbol{y}$. The time series is represented by a uniform grid: $\boldsymbol{x} = \left(\boldsymbol{\boldsymbol{x}_1, \ldots, \boldsymbol{x}_T}\right)$, where $T$ is the length of the series.\footnote{Notation is provided in \appendixref{app:notation}.}
    
    The step-wise embedding model (see \figureref{fig:scheme}) consists of three blocks: the feature embedding model, the group embedding model, and the sequence model \citep{kuznetsova2023importance}. The feature embedding model transforms an input time step $\boldsymbol{x}_t$ using a feature-specific linear layer with weights $\boldsymbol{W}_f$ followed by a more expressive nonlinear function $\psi$: 
    \begin{equation}\label{eq:feat_emb}
        \boldsymbol{h}_t = \psi \left(\left\{\boldsymbol{W}_f \boldsymbol{x}_{t,f}\right\}_{f=1}^F \right),
    \end{equation}
    Here, $F$ is the number of features. The group embedding model then splits the feature embeddings $\boldsymbol{h}_t$ into $K$ groups, processes them separately, and aggregates them back together into an embedding $\boldsymbol{g}_t$ (see \sectionref{sec:group-embedding}). That embedding is then passed on to the sequence model, which makes the final prediction $\hat{\boldsymbol{y}}$. The model is trained back-to-back using a supervised loss and, in the case of learned groups, a regularization term (see \sectionref{sec:training}).
    
    The feature groups are defined by a binary membership matrix $\boldsymbol{M}\in \{0,1\}^{F\times K}$, where $\boldsymbol{M}_{f,k}=1$ if the feature $f$ belongs to the group $k$, and $\boldsymbol{M}_{f,k}=0$ otherwise. Although these matrices can be given a priori, we instead propose learning them by clustering the feature-wise embedding weights $\boldsymbol{W}_f$ (see \sectionref{sec:learned-group-embedding}).

\subsection{Group Embedding}
\label{sec:group-embedding}

    Before describing the group embedding model, we formally define feature groups. A \emph{feature group} $k$ is a set of feature indices $G_k \subset \{1, \ldots, F\}$. In practice, it is convenient to represent groups using the membership matrix introduced above: $\boldsymbol{M}_k = \left\{\mathbbm{1}(f \in G_k)\right\}_{f=1}^F$. Each feature must belong to at least one group: $\forall f \ \exists k: \ \boldsymbol{M}_{f,k} = 1$. Following \citet{kuznetsova2023importance}, we initially require each feature to belong to exactly one group: $\forall f \ \exists! \ k: \ \boldsymbol{M}_{f,k} = 1$. For now, we assume that feature groups are given a priori.
        
    We now introduce the group embedding model. First, a \emph{split function} $S$ partitions the feature embeddings $\boldsymbol{h}_t$ into $K$ matrices $\boldsymbol{M}_k \boldsymbol{h}_t$, each corresponding to a group. Then a \emph{group-specific embedding function} $\phi_k$ independently processed each group (it can be parameterized, for example, by a small Transformer). Finally, an \emph{aggregation function} $A$ combines the resulting group embeddings $\phi_k\left(\boldsymbol{M}_k \boldsymbol{h}_t\right)$ into a single step-wise embedding $\boldsymbol{g}_t$. The aggregation function can be implemented as a concatenation, mean, or attention-pooling, with the choice treated as a hyperparameter.

    The complete group embedding procedure is given by:
    \begin{equation}\label{eq:group_emb}
    \begin{aligned}
        \boldsymbol{h}'_{t,k} &= S(\boldsymbol{h}_t), &\text{split into groups;}\\
        \boldsymbol{g}'_{t,k} &= \phi_k\left(h'_{t,k}\right), &\text{embed the groups;}\\
        \boldsymbol{g}_t &= A\left(g'_{t,k}\right), &\text{aggregate the groups.}
    \end{aligned}
    \end{equation}
    The sketch of the architecture described above is shown in \figureref{fig:learned_splitting}.

\subsection{Learned Groups}
\label{sec:learned-group-embedding}

    \paragraph{Hard groups} In the previous section, we assume access to a prior grouping of features. To relax this assumption, we introduce a feature clustering algorithm that produces the membership matrix $\boldsymbol{M}$ used by the splitting function $S$.

    To perform the clustering, the algorithm requires input vectors that characterize each feature. We propose using the feature-wise embedding weights $\{\boldsymbol{W}_f\}$ as these vectors.

    However, the dimensions of these matrices may vary between features (e.g., for numerical and one-hot encoded categorical features), making them unsuitable for direct comparison (see \sectionref{app:catnum}).

    To resolve this, we introduce a \emph{unification function} $U$ that maps (unifies) these matrices $\boldsymbol{W}_f = \left[\boldsymbol{w}_1, \ldots, \boldsymbol{w}_{C_f}, \boldsymbol{w}_{\text{bias}}\right]$, \revD{where $C_f$ is the size of encoding for feature $f$}. Specifically, $U$ combines the weight columns and concatenates them to the bias column. The function $U$ can then be written as:
    \begin{equation}\label{eq:unify}
        U\left(\boldsymbol{W}_f\right) =  \left[\boldsymbol{w}_{\text{bias}}^\top, \ \text{Combine}\left(\left\{\boldsymbol{w}_{i}\right\}_{i=1}^{C_f}\right)^\top\right]^\top .
    \end{equation}
    \noindent The combination function can sum the weight columns, take their mean, or ignore them entirely (producing an empty vector). The choice of this operation is treated as a hyperparameter (see \appendixref{app:unification}).
    
    Now, we can use any clustering algorithm on the set of vectors $\{U(\boldsymbol{W}_f)\}_{f=1}^F$ that characterize the features (see \figureref{fig:learned_splitting}). The algorithm then outputs the membership matrices $\{\boldsymbol{M}_k\}_{k=1}^K$.
    
    In this work, we consider centroid-based clustering algorithms: \emph{K-means}, \emph{Fuzzy K-means}, and Gaussian mixture model (\emph{GMM}) clustering algorithms \citep{hartigan1979algorithm, bezdek1984fcm, dempster1977maximum}. \revE{We focus on these algorithms because we find that controlling the dynamics of the centroid $\mu_k$ update is a useful form of regularization }(see \sectionref{sec:discandlims} for further discussion).
    
    \begin{figure}[tb]
    \floatconts
      {fig:learned_splitting}
      {\caption{Learned feature clustering scheme.}}
      {\includegraphics[width=\linewidth,page=10,trim={0 0.6cm 2.8cm 0},clip]{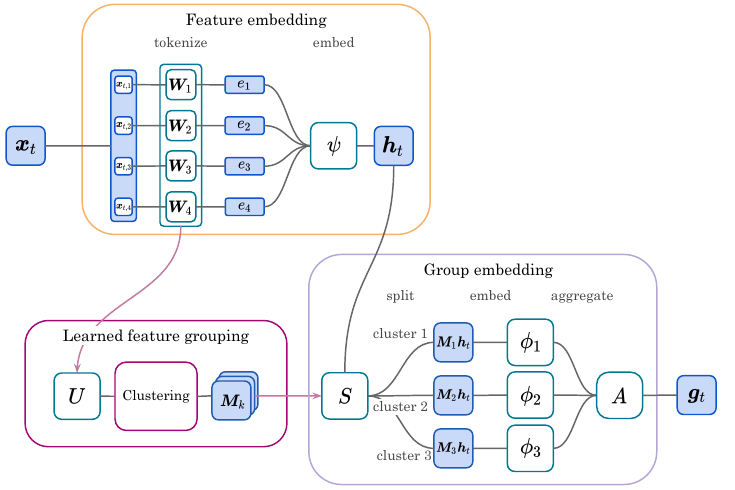}}
    \end{figure}

    \paragraph{Soft groups}  In the previous sections, following \citet{kuznetsova2023importance}, we assumed a hard clustering, where each feature belongs to exactly one cluster. However, in real data, it is natural for a variable to interact with multiple groups of features and, therefore, to belong to multiple clusters. For example, in a medical setting, variables that describe body fluid levels can be relevant to both the circulatory and renal systems.
    
    To account for this, we relax the hard clustering assumption by introducing a soft membership model. Instead of producing an indicator assignment vector for each feature, the algorithm produces a vector $\boldsymbol{p}_f$ that scores the ``probability'' of feature $f$ belonging to each group. 
    \begin{equation*}
        \boldsymbol{p}_f^k = \mathbb{P}(\text{feature $f$ is assigned to cluster $k$}),
    \end{equation*}
    where the score values are normalized but are not required to sum to one. The cluster membership matrix is then defined as
    \begin{equation}\label{eq:soft_membership}
        \boldsymbol{M}_{f,k} = \mathbbm{1} \left(\frac{\boldsymbol{p}_f^k}{\max_l \boldsymbol{p}_f^l} > \Delta_k\right),
    \end{equation}
    where $\Delta_k \in [0, 1)$ is the threshold on the membership score for assigning a feature to the cluster $k$. Here, division of the score vector by its maximum element prevents the clusters from being empty. Note that hard clustering is essentially an edge case of soft clustering when $\Delta \rightarrow 1$. 
    
    The thresholds $\Delta_k$ are treated as hyperparameters. For simplicity, we set them equal to $\Delta$ for all $k$. We discuss practical implementation and other options for the membership function in \appendixref{app:soft-details}.

\subsection{Training and Regularization}
\label{sec:training}

    The resulting model is trained using gradient descent with an additional update to the clustering (see \algorithmref{alg:featclu}). Specifically, every feature $f$ is assigned to a specific cluster $k$ based on $\boldsymbol{W}_f$. During the backpropagation phase, the weight matrices $\{\boldsymbol{W}_f\}$ are updated based on the \revE{loss gradient}. The feature clustering can then be periodically recomputed independently of the backpropagation. For example, every $P$ iterations, where $P$ is a hyperparameter. For simplicity, we set $P=1$.
    
    Since $\{\boldsymbol{W}_f\}$ are updated during training, clusters could fail to form if weight matrices change much from one epoch to another. To promote the formation of clusters within the latent feature space and improve training stability, we introduce a \emph{regularization loss term} $\mathcal L_{\text{reg}}$. Denoting cluster centroids (or means in the GMM case) $\{\boldsymbol{\mu}_k\}_{k=1}^K$: 
    \begin{equation}\label{eq:reg_loss}
    \begin{gathered}
        \mathcal L_{\text{reg}}\left(\left\{U(\boldsymbol{W}_f)\right\}_{f=1}^F\right) = \dfrac{\frac{1}{K}\sum_{k=1}^K\mathcal{L}^{\text{intra}}_k}{\frac{1}{K}\sum_{k=1}^K\mathcal{L}^{\text{inter}}_k},
    \end{gathered}
    \end{equation}
    
    \noindent Here $\mathcal{L}^{\text{inter}}$ is an \emph{intercluster loss} and $\mathcal{L}^{\text{intra}}$ is an \emph{intracluster loss} defined as 
    \begin{equation*}
        \mathcal L_k^{\text{intra}} = \sum_{i \in G_k} \lVert U(\boldsymbol W_i) - \boldsymbol \mu_k \rVert, ~~ \mathcal L_k^{\text{inter}} = \sum_{k' = 1}^K \lVert \boldsymbol \mu_k - \boldsymbol \mu_{k'} \rVert.
    \end{equation*}
    \noindent The former is responsible for decreasing the distances between features within the assigned cluster (``tightening'' the clusters). The latter is responsible for increasing the distances between different clusters (``pushing'' them apart).

   \begin{algorithm2e}[tb]
        \SetAlgoLined
        \DontPrintSemicolon
        \small
        \caption{Training learned feature groups}
        \label{alg:featclu}
        \KwIn{Num.\ of groups $K$, reg.\ coef.\ $\lambda$, EMA coef.\ $\alpha$}
        \KwInit{Feature groups $\{\boldsymbol{M}_k\}$, centroids $\{\boldsymbol{\mu}_k\}$}
        \While{training}{
            \For{each $t$}{
                $\boldsymbol{h}_t \gets \text{Feature embedding}(\boldsymbol{x}_t)$ \tcp*{\eqref{eq:feat_emb}}
                $\boldsymbol{g}_t \gets \text{Group embedding}\left(\boldsymbol{h}_t, \{\boldsymbol{M}_k\}\right)$ \tcp*{\eqref{eq:group_emb}}
            }
            $\hat{\boldsymbol{y}} \gets \text{Sequence model}(\boldsymbol{g})$ \;
            $\mathcal{L} \gets \mathcal{L}(\hat{\boldsymbol{y}}, \boldsymbol{y}) + \lambda \mathcal{L}_{\text{reg}}\left(\{\boldsymbol{\mu}_k\}, \{\boldsymbol{W}_f\}\right)$ \tcp*{\eqref{eq:loss}}
            Gradient step \;
            $\{\boldsymbol{M}_k\} \gets \text{Clustering}_{\{\boldsymbol{\mu}_k\}}\left(\{U(\boldsymbol{W}_f)\}\right)$ \tcp*{\eqref{eq:unify}}
            \If{$\{\boldsymbol{M}_k\}$ was updated}{
                $\boldsymbol\mu_k \gets \text{EMA}_{\alpha}(\boldsymbol\mu_k, \boldsymbol\mu_{k}')$ \tcp*{\eqref{eq:decay}} 
                $\{\boldsymbol{M}_k\} \gets \text{Clustering}_{\{\boldsymbol{\mu}_k\}}\left(\{U(\boldsymbol{W}_f)\}\right)$ \tcp*{\eqref{eq:unify}}
            }
        }
    \end{algorithm2e}

    This regularization term is important for ensuring the establishment of a robust clustering structure, especially in the early stages of training. Note that the gradient is only taken with respect to $\boldsymbol{W}_f$, so this function is indeed differentiable and only affects the feature embedding function defined in \equationref{eq:feat_emb}.

    The final loss then takes the form
    \begin{equation}\label{eq:loss}
        \mathcal{L} = \mathcal{L}(\boldsymbol{y}, \hat{\boldsymbol{y}}) + \lambda \mathcal{L}_{\text{reg}}\left(\{\boldsymbol{\mu}_k\}_{k=1}^K, \left\{\boldsymbol{W}_f\right\}_{f=1}^F\right),
    \end{equation}
    where $\mathcal{L}(\boldsymbol{y}, \hat{\boldsymbol{y}})$ is the supervision loss and $\lambda \in [0,1)$ is the weight of the clustering loss. Combining supervised task objective and feature group learning enables grouping of features based on their mutual relationships and relevance to the task label.
    
    To further improve training stability, we modify the update of $\boldsymbol{\mu}$ to take into account the previous centroid position by calculating an \emph{expected moving average} (EMA). This provides an additional degree of control over the dynamics of centroid movement. 

    Specifically, for K-means algorithms, given the centroids from the previous iteration $\boldsymbol{\mu}_k$ and the newly calculated centroids $\boldsymbol{\mu}_k'$, the update is
    \begin{equation}\label{eq:decay}
        \boldsymbol\mu_k = \alpha \boldsymbol\mu_k + (1 - \alpha) \boldsymbol\mu_k',
    \end{equation}
    where $\alpha \in [0,1)$ is the \emph{decay parameter}. 
    
    Similarly, for GMMs we can compute an ``EMA'' between the Gaussian distributions using moment matching (see \appendixref{app:gmm-ema} for a discussion),
    \begin{equation}
        \begin{aligned}
            \boldsymbol\mu_k &= \alpha \boldsymbol\mu_k + (1 - \alpha) \boldsymbol\mu_k', \\
            \boldsymbol\Sigma_k &= \alpha \boldsymbol\Sigma_k + (1 - \alpha) \boldsymbol\Sigma_k' + \\
            &\quad + \alpha (1 - \alpha) (\boldsymbol\mu_k - \boldsymbol\mu_k') \times (\boldsymbol\mu_k - \boldsymbol\mu_k')^T.
        \end{aligned}
        \label{eq:decay-gmm}
    \end{equation}

\section{Experiments}

\subsection{Synthetic Data}

\subsubsection{Setup}

    \begin{figure}[tb]
    \floatconts
      {fig:synthetic_data}
      {\caption{Synthetic data labeling procedure. Feature values are sampled from Gaussian Processes with various parameters.}}
      {\includegraphics[width=1.00\linewidth,page=11,trim={0.1cm 2.15cm 1.74cm 0},clip]{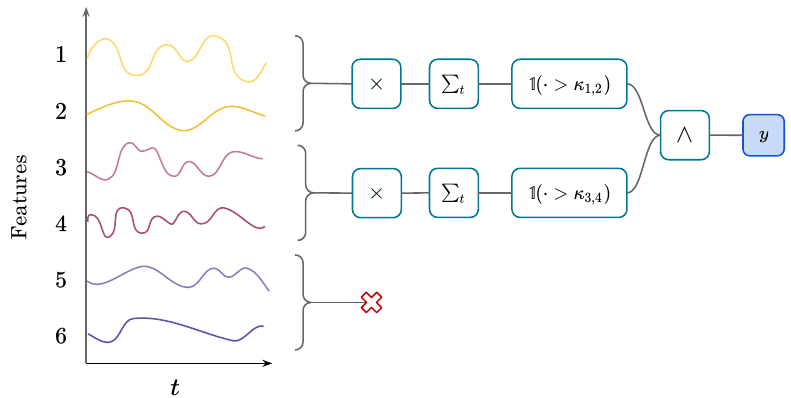}}
    \end{figure}

    \revA{First, we study the performance of our method on a synthetic dataset. While on real data, the target underlying feature groups are not known, we design the synthetic dataset where the true grouping is fixed. The dataset is simple, enabling comparison to usual clustering methods that use simple aggregated information to characterize the features.}

    \revD{To study the difference between these data representations, we use the same clustering algorithm for all settings. For simplicity, we choose K-means as other clustering methods can be seen as its generalization.}

    We generate a synthetic multivariate time series dataset with $6$ features of length $20$, where the feature values are sampled from Gaussian processes (GP) with individual parameters (see \figureref{fig:synthetic_data}). We define the ground-truth feature groups $\left\{\{1, 2\}, \{3, 4\}, \{5, 6\}\right\}$. Using these groups, we assign a binary label to each time series as follows. First, we sum the values of the feature products of $1$, $2$ and $3$, $4$ over all time steps. Then we calculate the indicator function to determine whether the aggregated values are greater than the thresholds $\kappa_{1,2}$ and $\kappa_{3,4}$. The thresholds are chosen as medians of the respective sums over the dataset. Finally, the two indicators are combined into a single label $y$ via a logical ``and'' operation. The features $5$ and $6$ form a group, as they both do not correlate with the label (see \appendixref{app:training_details_synth} for further details).

    \begin{table}[tb]
        \floatconts
        {tab:synth_results}%
        {\caption{Clustering quality on synthetic data. Means and standard deviations are computed over 5 runs. Best non-oracle performance is highlighted in bold. Dynamic clustering outperforms static clustering, which fails to learn ground-truth labels.}}
        {\resizebox{\linewidth}{!}{%
        \begin{tabular}{lrrr}
            \toprule
            Algorithm / Input & \multicolumn{1}{c}{ARI (\textuparrow)} & \multicolumn{1}{c}{NMI (\textuparrow)} & \multicolumn{1}{c}{Silhouette} \\ \midrule
            \multicolumn{4}{l}{\textbf{Reference models}} \\ 
            Random                           & -0.00 \textpm \ 0.57 & 0.40 \textpm \ 0.37 & -0.001 \textpm \  0.002 \\
            Oracle                           &  1.00 \textpm \ 0.00 & 1.00 \textpm \ 0.00 & -0.003 \textpm \ 0.000 \\ \addlinespace
            \multicolumn{4}{l}{\textbf{Static K-means}} \\ 
            $\mathcal{D}_{\text{flat}}$       & -0.24 \textpm \ 0.10 & 0.38 \textpm \ 0.05 & 0.002 \textpm \ 0.003 \\
            $\mathcal{D}_{\text{time mean}}$  & -0.14 \textpm \ 0.20 & 0.39 \textpm \ 0.11 & 0.004 \textpm \ 0.005 \\
            $\mathcal{D}_{\text{sample mean}}$& -0.03 \textpm \ 0.06 & 0.43 \textpm \ 0.04 & \textbf{0.110} \textpm \ 0.060 \\
            $\mathcal{D}_{\text{full mean}}$  &  0.04 \textpm \ 0.06 & 0.49 \textpm \ 0.05 & -0.001 \textpm \ 0.002 \\ \addlinespace
            \multicolumn{4}{l}{\textbf{Dynamic K-means}} \\ 
            $\mathcal{D}$                     &  \textbf{0.23} \textpm \ 0.51 & \textbf{0.59} \textpm \ 0.29 & -0.001 \textpm \ 0.002 \\ 
            \bottomrule
        \end{tabular}}}
    \end{table}

    Classical clustering algorithms require an input vector that characterizes the features and their relationship to the sample label $y$. Therefore, the input dataset $\mathcal{D}$ must be converted to a suitable format. \revD{We construct the new dataset as a concatenation of the label values $y$ with a vector representing the feature values. For the latter, we} consider the following options: (1) concatenation of all feature values across all samples $i$ and time steps $t$ ($\mathcal{D}_{\text{flat}}$); (2) concatenation of the feature values averaged over time ($\mathcal{D}_{\text{time mean}}$); (3) concatenation of the feature values averaged over the samples ($\mathcal{D}_{\text{sample mean}}$); (4) concatenation of the feature values averaged over both samples and time ($\mathcal{D}_{\text{full mean}}$). The formal definitions of these transformations are provided in \appendixref{app:training_details_synth}.
    
    A more sophisticated approach would involve training a sequence model to capture the interaction between features and the target label. This is precisely what our proposed algorithm does, without requiring any modification of the input dataset $\mathcal{D}$.

    We compare the ability of this static approach and our proposed dynamic algorithm to recover the target feature groups. For reference, we also report results for random clustering and an oracle.

    \revA{Performance is evaluated using the Adjusted Rand Index (ARI), Normalized Mutual Information (NMI), and the silhouette score. ARI and NMI are metrics that measure how close the learned clustering is to the true feature groups} \citep{luecken2022benchmarking, hassan2021novel}\revA{. They are in range [-1,1] and [0,1] respectively, with higher values indicating higher similarity. ARI measures how many permutations it would take to transform one into another. NMI represents the mutual information between true and predicted clusterings. The silhouette score, unlike ARI and NMI, does not rely on true feature groups. Instead, it measures how well the clusters are formed in a geometric sense using distances in the embedding space (see Appendix C.1 for details).}
    
\subsubsection{Results}

    We present the results in \tableref{tab:synth_results}. We see that the static K-means fails to recover the feature groups for all data representations, performing no better, and often worse, than random clustering. In contrast, dynamic K-means can recover the correct feature groups for at least some initializations. 
    
    We also notice that the silhouette score for the oracle is lower than for the random clustering and most static K-means assignments. This indicates that optimizing an internal clustering metric, such as the silhouette score, does not necessarily correspond to recovering the true feature groups. This finding highlights the importance of using a downstream metric that ensures that the learned groups actually benefit the target application.

\subsection{Medical Data}
\label{sec:med_results}

\subsubsection{Setup}

    \begin{table*}[htb]
        \floatconts
        {tab:icu_auprc_hard}%
        {\caption{\textbf{Performance benchmark for different grouping schemes} (AUPRC \textuparrow) on online and offline classification tasks using ICU datasets. Mean and standard deviation are reported over three training runs with different random seeds. Best results are highlighted in bold. Reference results for prior groups (Prior) and models without feature grouping (None) are taken from~\citet{kuznetsova2023importance}. Learned groupings perform comparably to expert-defined ones.}}%
        {
            {%
                \small
                \begin{tabular}{@{}ccccccc@{}}
                    \toprule
                    \multicolumn{3}{c}{Feature Grouping} & \multicolumn{2}{c}{HIRID} & \multicolumn{2}{c}{MIMIC-III} \\ \cmidrule(lr){1-3} \cmidrule(lr){4-5} \cmidrule(lr){6-7}
                    \mc{Type} & \mc{Membership} & \mc{Algorithm} & \mc{Circ} & \mc{Mort}  & \mc{Decomp} & \mc{Mort} \\ \midrule
                    None      & --   & --            & 0.388 \textpm \ 0.006 & 0.605 \textpm \ 0.006 & \textbf{0.387} \textpm \ 0.003 & 0.512 \textpm \ 0.008 \\
                    Prior     & Hard & Type          & \textbf{0.402} \textpm \ 0.004 & \textbf{0.627} \textpm \ 0.019 & 0.380 \textpm \ 0.004 & \revB{0.521} \textpm \ 0.001 \\
                    Prior     & Hard & Organ         & \textbf{0.406} \textpm \ 0.004 & \textbf{0.623} \textpm \ 0.012 & 0.374 \textpm \ 0.001 & \textbf{0.526} \textpm \ 0.006 \\\addlinespace
                    Learned   & Hard & K-means       & 0.393 \textpm \ 0.006 & \textbf{0.631} \textpm \ 0.009 & 0.381 \textpm \ 0.002 & \textbf{0.525} \textpm \ 0.006 \\
                    Learned   & Hard & Fuzzy K-means & 0.382 \textpm \ 0.010 & \textbf{0.629} \textpm \ 0.012 & \textbf{0.386} \textpm \ 0.007 & \textbf{0.523} \textpm \ 0.003 \\
                    Learned   & Hard & GMM           & 0.399 \textpm \ 0.001 & \textbf{0.625} \textpm \ 0.006 & 0.381 \textpm \ 0.001 & \textbf{0.519} \textpm \ 0.011 \\\addlinespace
                    Learned   & Soft & Fuzzy K-means & 0.395 \textpm \ 0.002 & 0.615 \textpm \ 0.001 & 0.377 \textpm \ 0.006 & \textbf{0.521} \textpm \ 0.009 \\
                    Learned   & Soft & GMM           & \textbf{0.403} \textpm \ 0.004 & \revB{0.610} \textpm \ 0.022 & 0.380 \textpm \ 0.003 & \textbf{0.520} \textpm \ 0.006 \\ \hline
                \end{tabular}
            }
        }
    \end{table*}
    
    Next, we evaluate our method's ability to improve downstream performance on real-world data. We consider two widely used Intensive Care Unit (ICU) datasets: HiRID \citep{hirid} and MIMIC-III \citep{MIMIC-III}. We follow the pre-processing and labeling protocol introduced by \citet{yeche2021} and subsequently adopted by \citet{yeche2024dsaforeep, kuznetsova2023importance}. The technical details and the full training setup are provided in \appendixref{app:setup}.

    We consider mortality and circulatory failure prediction on HiRID, and mortality and decompensation prediction on MIMIC-III. Mortality prediction (\emph{Mort}) is an offline binary classification task: given a time series, the model predicts whether the patient will survive at the end of their stay. Circulatory failure (\emph{Circ}) and decompensation (\emph{Decomp}) are early event prediction tasks (online binary classification): at each time step, the model predicts whether an event will occur in the next few hours (see \appendixref{app:tasks} for more details). 

    Given the strong class imbalance across all tasks, we evaluate performance using the Area Under the Precision-Recall Curve (AUPRC). We also report the Area Under the Receiver Operator Curve (AUROC), although it has been shown to tend to saturate in this setting~\citep{saito2015precision}, so we prioritize AUPRC (see \appendixref{app:metrics} for more details).

    We compare models that learn feature groups with models that use groups defined a priori. Following \citet{kuznetsova2023importance}, previous groups are defined based on (1) measurement type (e.g., vital signs, laboratory tests) and (2) the organ system to which the variables belong (e.g., circulatory, nervous). We also report reference results for models that do not use any feature grouping, with numbers taken directly from \citet{kuznetsova2023importance}, where the FT-Transformer~\citep{gorishniy2023revisitingdeeplearning} is used as the base architecture. Specific assignments are listed in \appendixref{app:prior_groups}.

\subsubsection{Results}

    \paragraph{Downstream performance} The results of different clustering algorithms along with models using prior groups (Prior) and models without feature grouping (None) are reported in \tableref{tab:icu_auprc_hard}. Our models generally outperform architectures without grouping and match the best results achieved by using the prior grouping in \citet{kuznetsova2023importance}. Additional results, including AUROC and p-values, are reported in \appendixref{app:add_res}.

    \paragraph{Interpretation of learned groups} The learned groups, analyzed in detail with the help of a medical doctor in \appendixref{app:learned_groups}, are generally clinically explainable. Models trained on the same task tend to discover related groups, frequently capturing either the baseline state of the patient or specific types of shock. However, larger clusters sometimes contain heterogeneous sets of variables, which reduces explainability and suggests that controlling for cluster size may be beneficial. These findings indicate potential for refining expert-defined feature groups by tailoring them to specific prediction tasks and patient cohorts.

    For the decompensation prediction on MIMIC-III, we observe that models without grouping outperform those with expert-defined groups. Examining the clustering dynamics during training (\figureref{fig:sankey_md_kf}), we find that the model learns to assign all features to a single group. This behavior suggests that the method can correct a suboptimal initialization, rediscovering the most suitable grouping strategy (see \appendixref{app:initialization} for further discussion). \revD{We discuss the possible reason behind the no-grouping outperforming algorithms with grouping on this task in} \appendixref{app:learned_groups}\revD{.}

    At the same time, the feature groups discovered in other tasks are novel. In \figureref{fig:learned_groups_color}, we visualize the embedding space of the model trained on HiRID for mortality prediction using t-SNE \citep{vandermaaten08a_tsne}, coloring the features according to the learned and expert-defined groupings. The learned clusters are well-formed, yet distinct from the expert-defined groups, indicating that the method uncovers new, task-relevant structures.

    \paragraph{Ablations} To better understand how the model hyperparameters affect its performance, we perform ablation studies and report their results in \appendixref{app:ablations}. They show that regularization and EMA improve performance on several tasks. However, there is no single best choice for the initialization method, number of clusters, unification function, or merge function, as the optimal configuration varies between tasks.

    \begin{figure*}[htb]
        \floatconts
        {fig:learned_groups_color}
        {\caption{Embedding space of learned features colored according to learned and prior groupings (K-means,  HiRID mortality). The embedding space is visualized using t-SNE of Euclidean distances between centroids and features. The model produces novel feature clusters that differ from prior groupings.}}
        {%
            \subfigure[By learned cluster]{%
                \includegraphics[width=0.32\linewidth]{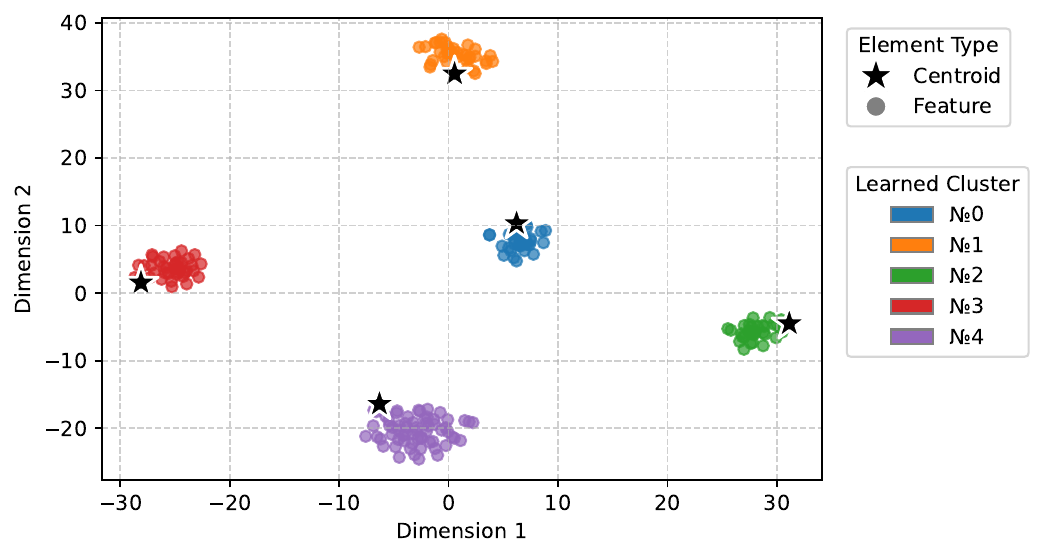}
            }
            \hfill
            \subfigure[By organ system]{%
                \includegraphics[width=0.32\linewidth]{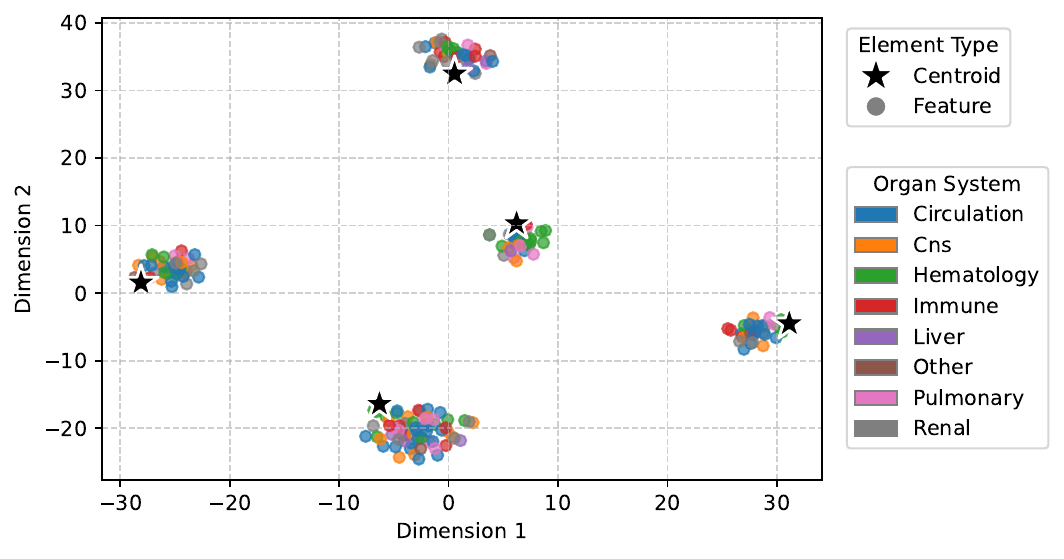}
            }
            \hfill
            \subfigure[By measurement type]{%
                \includegraphics[width=0.32\linewidth]{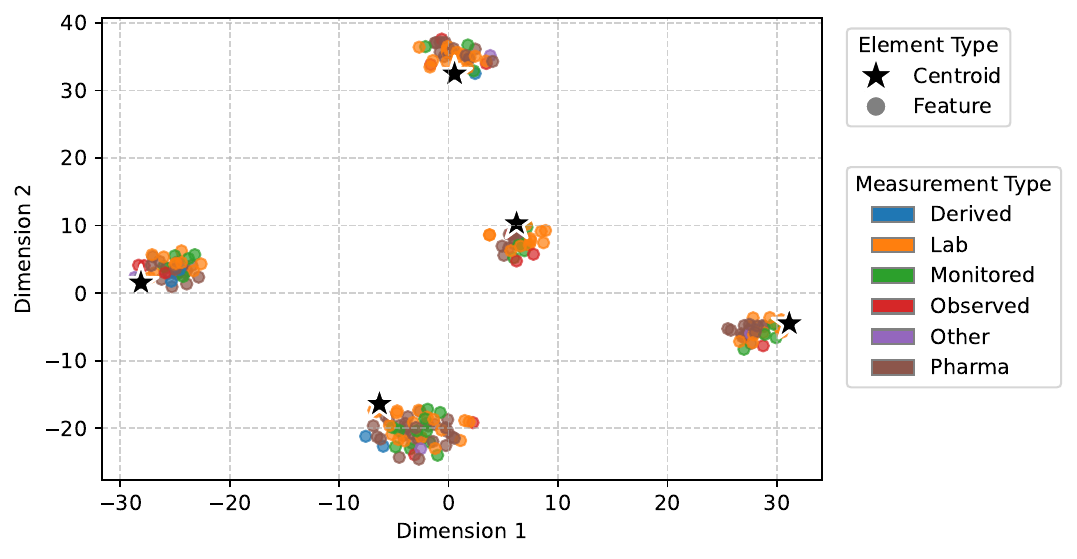}
            }
        }
    \end{figure*}

\section{Discussion and Limitations}
\label{sec:discandlims}
    
    \revA{Our method has shown comparable downstream performance to the a priori defined feature groups. This indicates that the feature groups proposed by} \citet{kuznetsova2023importance} \revA{are indeed a strong baseline. However, our method is able to match this baseline while providing greater flexibility. For example, on new data or tasks, it could automatically discover feature groups that lead to strong downstream performance, which is not guaranteed for prior groupings.}

    This flexibility comes at the cost of increased complexity in two respects, (1) computational cost and (2) additional hyperparameters. The computational cost arises from feature clustering, but since this step does not require GPU resources, we found that training times with prior and learned clustering were comparable in practice \revE{(see} \appendixref{app:ablations}\revE{)}. The additional hyperparameters include the number of clusters, the EMA coefficient, and the type and strength of regularization. However, given that base architectures already involve dozens of hyperparameters and require extensive tuning \citep{yeche2021, kuznetsova2023importance}, the additional hyperparameters might not pose a significant limitation in practice.
    
    In this work, we focused on centroid-based clustering algorithms, which allow additional regularization through the EMA of centroids. We did not explore centroid-less methods such as DBSCAN \citep{dbscan} and Gumbel-Softmax NN-based approaches \citep{imrie2022composite, jang2016categorical}. Since these methods lack centroid regularization, they may face challenges in dynamic training settings. Similarly, we have not considered updating centroids using gradients, as this might make the clustering less stable due to the stochasticity of the gradient descent. We leave the investigation of these methods to future work.
    
    Our model currently relies on clustering algorithms with a fixed number of clusters, specified as a hyperparameter. Although in practice models may use fewer clusters (e.g., K-means for decompensation prediction), automatic discovery of the optimal number of clusters remains an open direction for future research \revB{(we discuss this problem in more depth in} \sectionref{app:adaptive_cl_num} \revB{and provide initial results for the Elbow method in} \sectionref{app:elbow}\revB{).}

    Finally, we have limited the scope of our experiments to in-distribution evaluation. In the real world, generalization across hospitals is a critical challenge \citep{moor2023foundation, burger2024towards, burger2025foundation, van2023yet}. Our model is naturally adaptable to such settings and could potentially reveal how feature groups vary across institutions, thereby providing insights into distribution shifts.

\section{Related Work}

    \paragraph{ML for ICU} Large EHR datasets have enabled the rapid development of machine learning models for ICU time series. \citet{harutyunyan2019multitask,hyland2020early,yeche2021} formalized clinical prediction tasks and developed strong tree-based and deep learning architectures for MIMIC-III and HiRID datasets. \citet{yeche2023temporal, yeche2024dsaforeep,kuznetsova2023importance} further improved deep learning models by modifying the loss function and architecture. We build directly on this line of research, adopting the same datasets and task definitions.
    
    \paragraph{Feature embedding in time series} Out-of-the-box deep neural networks often struggle to match the performance of tree-based architectures on tabular data. To address this shortcoming, \citet{gorishniy2022embeddings} proposed various strategies for feature embedding and incorporated them into ResNet \citep{he2016deep} and Transformer \citep{vaswani2017attention} models. \citet{tomavsev2021use, tomavsev2019clinically, kuznetsova2023importance} adapted these innovations to ICU time series with various embedding architectures.

    \paragraph{Feature grouping} Predefined feature groups have been shown to improve the downstream performance and explainability of time series models. \citet{kuznetsova2023importance, tomavsev2019clinically, tomavsev2021use} employed expert-defined feature groups to improve downstream performance on ICU time series data. \citet{swamy2024interpretcc} used predefined feature groups to design an inherently interpretable neural network. We extend these ideas by \emph{learning} feature groups to directly improve downstream performance while enhancing model explainability. 

    Several works have explored data-driven feature grouping before, but with key differences from our approach. \citet{masoomi2020instance} considered \emph{local} (or instance-wise) feature grouping, while we focus on \emph{global} feature groups that could help predictive modeling for a set clinical task in all patients. \citet{imrie2022composite} also considered global feature grouping, learning feature groups for static data using an ensemble-based method. Our method is tailored for time series and learns groups by clustering embedding weights directly within a deep learning architecture.
        
    In a related context, the same clustering algorithms have been applied to time series in the context of sample grouping \citep{ma2019learning}, while our focus is on grouping features.
    Similarly, interdependencies between features can be expressed via attention weights \citep{ma2020concare}, while our method produces explicit and global groups. Finally, feature grouping has also been used to achieve other goals, such as improving the computational efficiency of the attention mechanism \citep{roy2021efficient, vyas2020fast}, tailoring it to specific downstream tasks \citep{li2021groupformer}, or phenotyping patients via tensor decomposition \citep{becker2023unsupervised}.

\section{Conclusion}

    In this work, we introduce a novel method for learning feature groups in clinical time series by clustering weights of feature-wise embeddings. Our experiments show that this dynamic model \revE{is capable of recovering} ground-truth feature groups on synthetic data. On real-world ICU datasets, our method produces novel feature groups that match the performance of expert-defined groups. We show that the model produces interpretable clusters and is capable of rediscovering the best-known groupings. Looking ahead, this framework could be used to explore feature groups across tasks and hospitals more broadly, offering valuable data-driven insights for building more robust and explainable models.

\printacks

\bibliography{paper}

@article{vandermaaten08a_tsne,
  author  = {Laurens van der Maaten and Geoffrey Hinton},
  title   = {Visualizing Data using t-{SNE}},
  journal = {Journal of Machine Learning Research},
  year    = {2008},
  volume  = {9},
  number  = {86},
  pages   = {2579--2605},
  url     = {http://jmlr.org/papers/v9/vandermaaten08a.html}
}

@article{moor2023foundation,
  title={Foundation models for generalist medical artificial intelligence},
  author={Moor, Michael and Banerjee, Oishi and Abad, Zahra Shakeri Hossein and Krumholz, Harlan M and Leskovec, Jure and Topol, Eric J and Rajpurkar, Pranav},
  journal={Nature},
  volume={616},
  number={7956},
  pages={259--265},
  year={2023},
  publisher={Nature Publishing Group UK London}
}

@misc{MIMIC-III,
  doi = {10.13026/C2XW26},
  url = {https://physionet.org/content/mimiciii/1.4/},
  author = {Johnson, Alistair and Pollard, Tom and Roger, Mark},
  title = {{MIMIC-III Clinical Database}},
  publisher = {PhysioNet},
  year = {2016}
}

@article{hirid,
  author = {Faltys, M. and Zimmermann, M. and Lyu, X. and Hüser, M. and Hyland, S. and Rätsch, G. and Merz, T.},
  journal = {PhysioNet},
  year = {2021},
  title = {{HiRID}, a high time-resolution ICU dataset (version 1.1.1).},
  doi = {10.13026/nkwc-js72},
  url = {https://doi.org/10.13026/nkwc-js72}
}

@article{van2023yet,
  title={Yet another {ICU} benchmark: A flexible multi-center framework for clinical {ML}},
  author={Van De Water, Robin and Schmidt, Hendrik and Elbers, Paul and Thoral, Patrick and Arnrich, Bert and Rockenschaub, Patrick},
  journal={arXiv preprint arXiv:2306.05109},
  year={2023}
}

@inproceedings{yeche2021,
 author = {Y\`{e}che, Hugo and Kuznetsova, Rita and Zimmermann, Marc and H\"{u}ser, Matthias and Lyu, Xinrui and Faltys, Martin and R\"{a}tsch, Gunnar},
 booktitle = {Proceedings of the Neural Information Processing Systems Track on Datasets and Benchmarks},
 editor = {J. Vanschoren and S. Yeung},
 pages = {},
 title = {{HiRID}-{ICU}-Benchmark --- A Comprehensive Machine Learning Benchmark on High-resolution {ICU} Data},
 ur ={https://datasets-benchmarks-proceedings.neurips.cc/paper_files/paper/2021/file/5878a7ab84fb43402106c575658472fa-Paper-round1.pdf},
 volume = {1},
 year = {2021}
}

@article{harutyunyan2019multitask,
  title={Multitask learning and benchmarking with clinical time series data},
  author={Harutyunyan, Hrayr and Khachatrian, Hrant and Kale, David C and Ver Steeg, Greg and Galstyan, Aram},
  journal={Scientific data},
  volume={6},
  number={1},
  pages={1--18},
  year={2019},
  publisher={Nature Publishing Group}
}

@article{hyland2020early,
  title={Early prediction of circulatory failure in the intensive care unit using machine learning},
  author={Hyland, Stephanie L and Faltys, Martin and H{\"u}ser, Matthias and Lyu, Xinrui and Gumbsch, Thomas and Esteban, Crist{\'o}bal and Bock, Christian and Horn, Max and Moor, Michael and Rieck, Bastian and others},
  journal={Nature medicine},
  volume={26},
  number={3},
  pages={364--373},
  year={2020},
  publisher={Nature Publishing Group}
}

@article{vaswani2017attention,
  title={Attention is all you need},
  author={Vaswani, Ashish and Shazeer, Noam and Parmar, Niki and Uszkoreit, Jakob and Jones, Llion and Gomez, Aidan N and Kaiser, Lukasz and Polosukhin, Illia},
  journal={arXiv preprint arXiv:1706.03762},
  year={2017}
}

@misc{gorishniy2023revisitingdeeplearning,
      title={Revisiting Deep Learning Models for Tabular Data}, 
      author={Yury Gorishniy and Ivan Rubachev and Valentin Khrulkov and Artem Babenko},
      year={2023},
      eprint={2106.11959},
      archivePrefix={arXiv},
      primaryClass={cs.LG},
      url={https://arxiv.org/abs/2106.11959}, 
}

@misc{yeche2024dsaforeep,
      title={Dynamic Survival Analysis for Early Event Prediction}, 
      author={Hugo Yèche and Manuel Burger and Dinara Veshchezerova and Gunnar Rätsch},
      year={2024},
      eprint={2403.12818},
      archivePrefix={arXiv},
      primaryClass={cs.LG},
      url={https://arxiv.org/abs/2403.12818}, 
}

@article{paszke2019pytorch,
  title={Pytorch: An imperative style, high-performance deep learning library},
  author={Paszke, Adam and Gross, Sam and Massa, Francisco and Lerer, Adam and Bradbury, James and Chanan, Gregory and Killeen, Trevor and Lin, Zeming and Gimelshein, Natalia and Antiga, Luca and others},
  journal={Advances in neural information processing systems},
  volume={32},
  year={2019}
}

@article{diederik2014adam,
  title={Adam: A method for stochastic optimization},
  author={Diederik, P Kingma},
  journal={(No Title)},
  year={2014}
}

@article{scikit-learn,
  title={Scikit-learn: Machine Learning in {P}ython},
  author={Pedregosa, F. and Varoquaux, G. and Gramfort, A. and Michel, V.
          and Thirion, B. and Grisel, O. and Blondel, M. and Prettenhofer, P.
          and Weiss, R. and Dubourg, V. and Vanderplas, J. and Passos, A. and
          Cournapeau, D. and Brucher, M. and Perrot, M. and Duchesnay, E.},
  journal={Journal of Machine Learning Research},
  volume={12},
  pages={2825--2830},
  year={2011}
}

@inproceedings{kuznetsova2023importance,
  title={On the importance of step-wise embeddings for heterogeneous clinical time-series},
  author={Kuznetsova, Rita and Pace, Aliz{\'e}e and Burger, Manuel and Y{\`e}che, Hugo and R{\"a}tsch, Gunnar},
  booktitle={Machine Learning for Health (ML4H)},
  pages={268--291},
  year={2023},
  organization={PMLR}
}

@article{swamy2024interpretcc,
  title={Interpret{CC}: intrinsic user-centric interpretability through global mixture of experts},
  author={Swamy, Vinitra and Montariol, Syrielle and Blackwell, Julian and Frej, Jibril and Jaggi, Martin and K{\"a}ser, Tanja},
  journal={Network},
  volume={2},
  pages={G1},
  year={2024}
}

@article{roy2021efficient,
  title={Efficient content-based sparse attention with routing transformers},
  author={Roy, Aurko and Saffar, Mohammad and Vaswani, Ashish and Grangier, David},
  journal={Transactions of the Association for Computational Linguistics},
  volume={9},
  pages={53--68},
  year={2021},
  publisher={MIT Press One Rogers Street, Cambridge, MA 02142-1209, USA journals-info~…}
}

@inproceedings{li2021groupformer,
  title={Groupformer: Group activity recognition with clustered spatial-temporal transformer},
  author={Li, Shuaicheng and Cao, Qianggang and Liu, Lingbo and Yang, Kunlin and Liu, Shinan and Hou, Jun and Yi, Shuai},
  booktitle={Proceedings of the IEEE/CVF International Conference on Computer Vision},
  pages={13668--13677},
  year={2021}
}

@article{vyas2020fast,
  title={Fast transformers with clustered attention},
  author={Vyas, Apoorv and Katharopoulos, Angelos and Fleuret, Fran{\c{c}}ois},
  journal={Advances in Neural Information Processing Systems},
  volume={33},
  pages={21665--21674},
  year={2020}
}

@article{becker2023unsupervised,
  title={Unsupervised {EHR}-based phenotyping via matrix and tensor decompositions},
  author={Becker, Florian and Smilde, Age K and Acar, Evrim},
  journal={Wiley Interdisciplinary Reviews: Data Mining and Knowledge Discovery},
  volume={13},
  number={4},
  pages={e1494},
  year={2023},
  publisher={Wiley Online Library}
}

@article{gorishniy2022embeddings,
  title={On embeddings for numerical features in tabular deep learning},
  author={Gorishniy, Yury and Rubachev, Ivan and Babenko, Artem},
  journal={Advances in Neural Information Processing Systems},
  volume={35},
  pages={24991--25004},
  year={2022}
}

@article{imrie2022composite,
  title={Composite feature selection using deep ensembles},
  author={Imrie, Fergus and Norcliffe, Alexander and Li{\`o}, Pietro and van der Schaar, Mihaela},
  journal={Advances in Neural Information Processing Systems},
  volume={35},
  pages={36142--36160},
  year={2022}
}

@article{masoomi2020instance,
  title={Instance-wise feature grouping},
  author={Masoomi, Aria and Wu, Chieh and Zhao, Tingting and Wang, Zifeng and Castaldi, Peter and Dy, Jennifer},
  journal={Advances in Neural Information Processing Systems},
  volume={33},
  pages={13374--13386},
  year={2020}
}

@article{ma2019learning,
  title={Learning representations for time series clustering},
  author={Ma, Qianli and Zheng, Jiawei and Li, Sen and Cottrell, Gary W},
  journal={Advances in neural information processing systems},
  volume={32},
  year={2019}
}

@inproceedings{ma2020concare,
  title={Concare: Personalized clinical feature embedding via capturing the healthcare context},
  author={Ma, Liantao and Zhang, Chaohe and Wang, Yasha and Ruan, Wenjie and Wang, Jiangtao and Tang, Wen and Ma, Xinyu and Gao, Xin and Gao, Junyi},
  booktitle={Proceedings of the AAAI Conference on Artificial Intelligence},
  volume={34}, 
  pages={833--840},
  year={2020}
}

@article{burger2024towards,
  title={Towards Foundation Models for Critical Care Time Series},
  author={Burger, Manuel and Sergeev, Fedor and Londschien, Malte and Chopard, Daphn{\'e} and Y{\`e}che, Hugo and Gerdes, Eike and Leshetkina, Polina and Morgenroth, Alexander and Bab{\"u}r, Zeynep and Bogojeska, Jasmina and others},
  journal={arXiv preprint arXiv:2411.16346},
  year={2024}
}

@article{hartigan1979algorithm,
  title={Algorithm AS 136: A k-means clustering algorithm},
  author={Hartigan, John A. and Wong, Manchek A.},
  journal={Journal of the royal statistical society. series c (applied statistics)},
  volume={28},
  number={1},
  pages={100--108},
  year={1979},
  publisher={JSTOR}
}

@article{bezdek1984fcm,
  title={{FCM}: The fuzzy c-means clustering algorithm},
  author={Bezdek, James C. and Ehrlich, Robert and Full, William},
  journal={Computers \& geosciences},
  volume={10},
  number={2-3},
  pages={191--203},
  year={1984},
  publisher={Elsevier}
}

@article{dempster1977maximum,
  title={Maximum likelihood from incomplete data via the {EM} algorithm},
  author={Dempster, Arthur P. and Laird, Nan M. and Rubin, Donald B.},
  journal={Journal of the royal statistical society: series B (methodological)},
  volume={39},
  number={1},
  pages={1--22},
  year={1977},
  publisher={Wiley Online Library}
}

@article{runnalls2007kullback,
  title={{K}ullback-{L}eibler approach to {G}aussian mixture reduction},
  author={Runnalls, Andrew R.},
  journal={IEEE Transactions on Aerospace and Electronic Systems},
  volume={43},
  number={3},
  pages={989--999},
  year={2007},
  publisher={IEEE}
}

@article{alspach2003nonlinear,
  title={Nonlinear {B}ayesian estimation using {G}aussian sum approximations},
  author={Alspach, Daniel and Sorenson, Harold},
  journal={IEEE transactions on automatic control},
  volume={17},
  number={4},
  pages={439--448},
  year={2003},
  publisher={IEEE}
}

@article{cao2014generalized,
  title={Generalized product of experts for automatic and principled fusion of {G}aussian process predictions},
  author={Cao, Yanshuai and Fleet, David J.},
  journal={arXiv preprint arXiv:1410.7827},
  year={2014}
}

@article{hinton2002training,
  title={Training products of experts by minimizing contrastive divergence},
  author={Hinton, Geoffrey E.},
  journal={Neural computation},
  volume={14},
  number={8},
  pages={1771--1800},
  year={2002},
  publisher={MIT Press}
}

@article{agueh2011barycenters,
  title={Barycenters in the {W}asserstein space},
  author={Agueh, Martial and Carlier, Guillaume},
  journal={SIAM Journal on Mathematical Analysis},
  volume={43},
  number={2},
  pages={904--924},
  year={2011},
  publisher={SIAM}
}

@inproceedings{cuturi2014fast,
  title={Fast computation of {W}asserstein barycenters},
  author={Cuturi, Marco and Doucet, Arnaud},
  booktitle={International conference on machine learning},
  pages={685--693},
  year={2014},
  organization={PMLR}
}

@article{burger2025foundation,
  title={A Foundation Model for Intensive Care Unlocking Generalization across Tasks and Domains at Scale},
  author={Burger, Manuel and Chopard, Daphn{\'e} and Londschien, Malte and Sergeev, Fedor and Y{\`e}che, Hugo and Kuznetsova, Rita and Faltys, Martin and Gerdes, Eike and Leshetkina, Polina and B{\"u}hlmann, Peter and others},
  journal={medRxiv},
  pages={2025--07},
  year={2025},
  publisher={Cold Spring Harbor Laboratory Press}
}

@inproceedings{he2016deep,
  title={Deep residual learning for image recognition},
  author={He, Kaiming and Zhang, Xiangyu and Ren, Shaoqing and Sun, Jian},
  booktitle={Proceedings of the IEEE conference on computer vision and pattern recognition},
  pages={770--778},
  year={2016}
}

@article{kelly2009multiple,
  title={Multiple mutations in genetic cardiovascular disease: a marker of disease severity?},
  author={Kelly, Matthew and Semsarian, Christopher},
  journal={Circulation: Cardiovascular Genetics},
  volume={2},
  number={2},
  pages={182--190},
  year={2009},
  publisher={Lippincott Williams \& Wilkins}
}

@article{meira2001cancer,
  title={Cancer predisposition in mutant mice defective in multiple genetic pathways: uncovering important genetic interactions},
  author={Meira, Lisiane B. and Reis, Antonio MC. and Cheo, David L. and Nahari, Dorit and Burns, Dennis K. and Friedberg, Errol C.},
  journal={Mutation Research/Fundamental and Molecular Mechanisms of Mutagenesis},
  volume={477},
  number={1-2},
  pages={51--58},
  year={2001},
  publisher={Elsevier}
}

@article{tomavsev2019clinically,
  title={A clinically applicable approach to continuous prediction of future acute kidney injury},
  author={Toma{\v{s}}ev, Nenad and Glorot, Xavier and Rae, Jack W and Zielinski, Michal and Askham, Harry and Saraiva, Andre and Mottram, Anne and Meyer, Clemens and Ravuri, Suman and Protsyuk, Ivan and others},
  journal={Nature},
  volume={572},
  number={7767},
  pages={116--119},
  year={2019},
  publisher={Nature Publishing Group UK London}
}

@article{tomavsev2021use,
  title={Use of deep learning to develop continuous-risk models for adverse event prediction from electronic health records},
  author={Toma{\v{s}}ev, Nenad and Harris, Natalie and Baur, Sebastien and Mottram, Anne and Glorot, Xavier and Rae, Jack W and Zielinski, Michal and Askham, Harry and Saraiva, Andre and Magliulo, Valerio and others},
  journal={Nature protocols},
  volume={16},
  number={6},
  pages={2765--2787},
  year={2021},
  publisher={Nature Publishing Group UK London}
}

@article{rousseeuw1987silhouettes,
  title={Silhouettes: a graphical aid to the interpretation and validation of cluster analysis},
  author={Rousseeuw, Peter J.},
  journal={Journal of computational and applied mathematics},
  volume={20},
  pages={53--65},
  year={1987},
  publisher={Elsevier}
}

@article{hubert1985comparing,
  title={Comparing partitions},
  author={Hubert, Lawrence and Arabie, Phipps},
  journal={Journal of classification},
  volume={2},
  number={1},
  pages={193--218},
  year={1985},
  publisher={Springer}
}

@article{vinh2009information,
  title={Information theoretic measures for clusterings comparison: Variants},
  author={Vinh, NX. and Epps, J. and Bailey, J.},
  journal={Properties, normalization and correction for chance},
  volume={18},
  year={2009}
}

@Article{dbscan,
    title = {{dbscan}: Fast Density-Based Clustering with {R}},
    author = {Michael Hahsler and Matthew Piekenbrock and Derek Doran},
    journal = {Journal of Statistical Software},
    year = {2019},
    volume = {91},
    number = {1},
    pages = {1--30},
    doi = {10.18637/jss.v091.i01},
  }

@techreport{arthur2006k,
  title={k-means\texttt{++}: The advantages of careful seeding},
  author={Arthur, David and Vassilvitskii, Sergei},
  year={2006},
  institution={Stanford}
}

@article{jang2016categorical,
  title={Categorical reparameterization with gumbel-softmax},
  author={Jang, Eric and Gu, Shixiang and Poole, Ben},
  journal={arXiv preprint arXiv:1611.01144},
  year={2016}
}

@article{chormunge2018correlation,
  title={Correlation based feature selection with clustering for high dimensional data},
  author={Chormunge, Smita and Jena, Sudarson},
  journal={Journal of Electrical Systems and Information Technology},
  volume={5},
  number={3},
  pages={542--549},
  year={2018},
  publisher={Elsevier}
}

@inproceedings{masoomi2020,
 author = {Masoomi, Aria and Wu, Chieh and Zhao, Tingting and Wang, Zifeng and Castaldi, Peter and Dy, Jennifer},
 booktitle = {Advances in Neural Information Processing Systems},
 editor = {H. Larochelle and M. Ranzato and R. Hadsell and M.F. Balcan and H. Lin},
 pages = {13374--13386},
 publisher = {Curran Associates, Inc.},
 title = {Instance-wise Feature Grouping},
 url = {https://proceedings.neurips.cc/paper_files/paper/2020/file/9b10a919ddeb07e103dc05ff523afe38-Paper.pdf},
 volume = {33},
 year = {2020}
}

@article{saito2015precision,
  title={The precision-recall plot is more informative than the ROC plot when evaluating binary classifiers on imbalanced datasets},
  author={Saito, Takaya and Rehmsmeier, Marc},
  journal={PloS one},
  volume={10},
  number={3},
  pages={e0118432},
  year={2015},
  publisher={Public Library of Science San Francisco, CA USA}
}

@inproceedings{yeche2023temporal,
  title={Temporal label smoothing for early event prediction},
  author={Y{\`e}che, Hugo and Pace, Aliz{\'e}e and Ratsch, Gunnar and Kuznetsova, Rita},
  booktitle={International Conference on Machine Learning},
  pages={39913--39938},
  year={2023},
  organization={PMLR}
}

@article{student1908probable,
  title={The probable error of a mean},
  author={Student},
  journal={Biometrika},
  pages={1--25},
  year={1908},
  publisher={JSTOR}
}

@article{luecken2022benchmarking,
  title={Benchmarking atlas-level data integration in single-cell genomics},
  author={Luecken, Malte D and B{\"u}ttner, Maren and Chaichoompu, Kridsadakorn and Danese, Anna and Interlandi, Marta and M{\"u}ller, Michaela F and Strobl, Daniel C and Zappia, Luke and Dugas, Martin and Colom{\'e}-Tatch{\'e}, Maria and others},
  journal={Nature methods},
  volume={19},
  number={1},
  pages={41--50},
  year={2022},
  publisher={Nature Publishing Group US New York}
}

@article{hassan2021novel,
  title={A novel cluster detection of COVID-19 patients and medical disease conditions using improved evolutionary clustering algorithm star},
  author={Hassan, Bryar A and Rashid, Tarik A and Hamarashid, Hozan K},
  journal={Computers in biology and medicine},
  volume={138},
  pages={104866},
  year={2021},
  publisher={Elsevier}
}

@article{lange2004stability,
  title={Stability-based validation of clustering solutions},
  author={Lange, Tilman and Roth, Volker and Braun, Mikio L and Buhmann, Joachim M},
  journal={Neural computation},
  volume={16},
  number={6},
  pages={1299--1323},
  year={2004},
  publisher={MIT Press One Rogers Street, Cambridge, MA 02142-1209, USA journals-info~…}
}

@article{ferguson1973bayesian,
  title={A Bayesian analysis of some nonparametric problems},
  author={Ferguson, Thomas S},
  journal={The annals of statistics},
  pages={209--230},
  year={1973},
  publisher={JSTOR}
}

@article{thorndike1953belongs,
  title={Who belongs in the family?},
  author={Thorndike, Robert L},
  journal={Psychometrika},
  volume={18},
  number={4},
  pages={267--276},
  year={1953},
  publisher={Springer-Verlag}
}

@article{yadav2025failure,
  title={Failure Modes of Time Series Interpretability Algorithms for Critical Care Applications and Potential Solutions},
  author={Yadav, Shashank and Subbian, Vignesh},
  journal={arXiv preprint arXiv:2506.19035},
  year={2025}
}

@inproceedings{goutte2005probabilistic,
  title={A probabilistic interpretation of precision, recall and F-score, with implication for evaluation},
  author={Goutte, Cyril and Gaussier, Eric},
  booktitle={European conference on information retrieval},
  pages={345--359},
  year={2005},
  organization={Springer}
}

@article{chicco2023matthews,
  title={The Matthews correlation coefficient (MCC) should replace the ROC AUC as the standard metric for assessing binary classification},
  author={Chicco, Davide and Jurman, Giuseppe},
  journal={BioData Mining},
  volume={16},
  number={1},
  pages={4},
  year={2023},
  publisher={Springer}
}

\newpage

\appendix

\section{Notation}
\label{app:notation}

    \begin{itemize}
        \itemsep0em 
        \item Data
            \begin{itemize}
                \itemsep0em 
                \item $\boldsymbol{x} \in \mathbb{R}^{T \times \sum_{f=1}^F C_f }$ --- time series
                \item $\boldsymbol{x}_t$ --- time step $t \in \{1, \ldots, T\}$
                \item $\boldsymbol{x}_{t,f}$ --- feature $f \in \{1, \ldots, F\}$ at time step $t$
                \item $\boldsymbol{y}$ --- true label, either $\boldsymbol{y} \in \{0, 1\}$ for binary classification or $\boldsymbol{y} \in \{0, 1\}^T$ for early event prediction
                \item $\hat{\boldsymbol{y}}$ --- predicted label
                \item $T$ --- length of the time series
                \item $F$ --- number of features
                \item $N_{\text{num}}$ --- number of numerical features
                \item $N_{\text{cat}}$ --- number of categorical features
                \item $C_f$ --- size of feature $f$ encoding, equals to one for numerical and to the number of classes for categorical features
                \item $D$ --- size of the dataset
            \end{itemize}
        \item Embedding architecture
            \begin{itemize}
                \item $H$ --- token size
                \item $\boldsymbol{W}_{f} \in \mathbb{R}^{C_f \times H}$ --- tokenization matrix
                \item $\boldsymbol{w}_i$ --- $i$-th column of $\boldsymbol{W}_{f}$
                \item $\psi$ --- nonlinear feature embedding function
                \item $K$ --- number of feature groups
                \item $G_k \in 2^F$ --- set of indices of features belonging to the group $k \in \{1, \ldots, K\}$
                \item $\boldsymbol{M}$ --- feature-group membership matrix; for hard clustering $\boldsymbol{M} \in \{0,1\}^{F,K}$ where $\boldsymbol{M}_{k,f} = 1$ iff $f \in G_i$, for soft clustering $\boldsymbol{M} \in \mathbb{R}^{F,K}$
                \item $S$ --- splitting function
                \item $\phi_k$ --- group embedding function for group $k$
                \item $A$ --- aggregation function
                \item $\boldsymbol{g}_t$ --- final embedding 
            \end{itemize}
        \item Learned grouping
            \begin{itemize}
                \item $U$ --- function converting feature embedding matrices $\boldsymbol{W}_f$ to the same dimensionality
                \item $\boldsymbol{\mu}$ --- centroid or GMM
                \item $\Sigma$ --- GMM covariance
                \item $\mathcal L_{\text{reg}}$ --- cluster regularization loss term
                \item $\mathcal L^{\text{inter}}$ --- intercluster loss, ``tightens'' the clusters
                \item $\mathcal L^{\text{intra}}$ --- intracluster loss, ``pushes'' the clusters apart
                \item $\lambda$ --- regularization loss term coefficient
                \item $\alpha$ --- centroid moving decay parameter
                \item $\boldsymbol{p}_f \in [0, 1]^K$--- cluster membership score vector for feature $f$
                \item $\Delta$ -- cluster membership threshold
            \end{itemize}
    \end{itemize}

\section{Method Details}

\subsection{Categorical and Numerical Features}
\label{app:catnum}

    In the medical time series we consider, each feature $\boldsymbol{x}_{t,f}$ can be categorical or numerical. The categorical features are considered to be one hot encoded: $\boldsymbol{x}_{t,f} \in \mathbbm{1}^{C_f}$, where $C_f$ is the number of categories for feature $f$. The numerical features are simply represented by real numbers: $\boldsymbol{x}_{t,f} \in \mathbb{R}^{C_f}$, $C_f = 1$. We denote the number of categorical and numerical features $N_{\text{cat}}$ and $N_{\text{num}}$, respectively. Without loss of generality, we will assume that numerical features are indexed $\{1, \ldots, N_{cat}\}$. Then we can write $\boldsymbol{x} \in \mathbb{R}^{T \times (N_{\text{num}} + \sum_{f=1}^{N_{\text{cat}}} C_f)}$.
    
    The embedding matrices then have the following dimensions $\boldsymbol{W}_f \in \mathbb{R}^{(C_f + 1) \times H}$ ($+1$ for the bias term). For numerical features, the dimensions of $\boldsymbol{W}_f$ are the same for any $f$: the matrix contains two columns for the weight and the bias term, respectively. For categorical variables, $\{\boldsymbol{W}_f\}$ has $C_f + 1$ columns, so the dimensions can differ for different variables. Because the dimensions can differ for different features, so we have to introduce the unification function $U$.

\subsection{Unification Function}
\label{app:unification}

    In \equationref{eq:unify} we define the unification function as 
    \begin{equation*}
        U\left(\boldsymbol{W}_f\right) = \left[\boldsymbol{w}_{\text{bias}}^\top, \ \text{Combine}\left(\left\{\boldsymbol{w}_{i}\right\}_{i=1}^{C_f}\right)^\top\right]^\top.
    \end{equation*}
    Here we more formally define \textsf{Combine} and introduce abbreviations used in the ablation tables \tableref{tab:ablation_hirid_circ_gmm-soft,tab:ablation_hirid_mort_kmeans,tab:ablation_hirid_mort_kmeans-fuzzy,tab:ablation_mimic_decomp_gmm,tab:ablation_mimic_mort_kmeans}.

    Specifically, for numerical features \textsf{Combine} returns the weight unchanged, while for categorical features it performs one of the following operations:
    \begin{itemize}
        \item ignore (\emph{bias}) returning $\emptyset$;
        \item substitute with zeros (\emph{bias\_ext\_catzero}): $\boldsymbol{0}_{C_f}$;
        \item sum (\emph{bias\_sum\_linear}): $\sum_{i=1}^{C_f} \boldsymbol{w}_i$;
        \item average (\emph{bias\_avg\_linear}): $\frac{1}{C_f} \sum_{i=1}^{C_f} \boldsymbol{w}_i$.
    \end{itemize}

\subsection{Regularization}

    Denote $\boldsymbol{u}_i = U(\boldsymbol W_i)$. The intracluster loss introduced earlier is then
    \begin{equation*}
        \mathcal L_k^{\text{intra}} = \sum_{i \in G_k} \left\lVert \boldsymbol{u}_i - \boldsymbol \mu_k \right\rVert = \sum_{i = 1}^F \left\lVert \boldsymbol{M}_{k,i} \boldsymbol{u}_i - \boldsymbol \mu_k \right\rVert.
    \end{equation*}
    This form is easy to compute in a vectorized fashion, so we use it to calculate the loss in practice. We call this version \emph{hard intracluster loss}.
    
    We introduce another, more general loss for cluster compactness that we call a \emph{soft intracluster loss}. Now, instead of calculating the average over the distances between each cluster's centroid and \emph{its} features, we calculate the average over the weighed distances between cluster centroids and \emph{all} features. This is achieved by applying a softmax to the real-valued membership matrix $\boldsymbol{M}$. For hard clustering, $\boldsymbol{M}$ contains raw distances between each centroid and feature. For soft clustering, $\boldsymbol{M}$ is just the usual real-valued membership matrix. The loss is then
    \begin{equation*}
        \mathcal L_k^{\text{intra}} = \sum_{i = 1}^F \left\lVert \operatorname{Softmax}\left[ \boldsymbol{M}_{k,i}\right] \boldsymbol{u}_i - \boldsymbol \mu_k \right\rVert.
    \end{equation*}

    We consider both regularization variants for all algorithms, with the choice treated as a hyperparameter.

\subsection{EMA for GMM}
\label{app:gmm-ema}

    The expected moving average (EMA) allows us to stabilize the training by making the centroid updates more conservative. For GMM, a similar operation would involve computing an ``average'' between two Gaussian distributions that should also be Gaussian. This can be done using several approaches.
    
    First, the resulting distribution can be viewed as a mixture of experts \citep{runnalls2007kullback, alspach2003nonlinear}. However, a linear mixture of two Gaussian distributions is not necessarily Gaussian. Instead, a new Gaussian distribution can be built to match the moments of the mixture distribution. This Gaussian distribution also happens to have a minimal KL distance to the original mixture distribution among the Gaussian distributions. Its mean and covariance are
    \begin{equation*}
        \begin{aligned}
            \boldsymbol\mu_k &= \alpha \boldsymbol\mu_k + (1 - \alpha) \boldsymbol\mu_k', \\
            \boldsymbol\Sigma_k &= \alpha \boldsymbol\Sigma_k + (1 - \alpha) \boldsymbol\Sigma_k' + \\
            &\quad + \alpha (1 - \alpha) (\boldsymbol\mu_k - \boldsymbol\mu_k') \times (\boldsymbol\mu_k - \boldsymbol\mu_k')^T.
        \end{aligned}
    \end{equation*}
    
    Second, the resulting distribution can be viewed as a (generalized) product of experts \citep{hinton2002training, cao2014generalized}. In this case, the resulting distribution is Gaussian with the following parameters
    \begin{equation*}
        \begin{aligned}
            \boldsymbol{\Sigma}_{new}^{-1} &= \alpha \boldsymbol{\Sigma}_1^{-1} + (1-\alpha) \boldsymbol{\Sigma}_2^{-1}, \\
            \boldsymbol{\mu}_{new} &= \boldsymbol{\Sigma}_{new} \left( \alpha \boldsymbol{\Sigma}_1^{-1} \boldsymbol{\mu}_1 + (1-\alpha) \boldsymbol{\Sigma}_2^{-1} \boldsymbol{\mu}_2 \right).
        \end{aligned}
    \end{equation*}
    
    Third, from the perspective of optimal transport theory, the resulting distribution can be viewed as the Wasserstein barycenter of the original Gaussian distributions \citep{agueh2011barycenters, cuturi2014fast}. In the case of Wasserstein-2 distance, it is also a Gaussian with mean and covariance
    \begin{equation*}
        \begin{aligned}
            \boldsymbol\mu_{new} &= \alpha \boldsymbol\mu_1 + (1-\alpha) \boldsymbol\mu_2,\\
            \boldsymbol\Sigma_{new} &= \left(\alpha \boldsymbol\Sigma_1^{1/2} + (1-\alpha) \boldsymbol\Sigma_2^{1/2}\right)^2.     
        \end{aligned}
    \end{equation*}

    For our application, we choose to use moment matching because it is the fastest and easiest to compute. Future works can investigate whether other approaches lead to better stabilization of the GMM components during training.

\subsection{Soft Membership Functions}
\label{app:soft-details}

    The soft membership function transforms a real-valued characterization of the cluster membership (e.g., distances to the centroids for K-means) to a binary one. Ideally, it should produce clusters that are (a) not empty and (b) not overly full. Inability to satisfy these conditions would likely lead to non-informative feature groups and worse downstream performance. Furthermore, the function should expose a parameter to control ``softness'' of the learned clusters. Designing such a function for use during training is not trivial, so we considered multiple options. 
    
    First, we tried simple thresholding with the threshold parameter $\Delta$ controlling the softness.
    \begin{equation*}
        \boldsymbol{M}_{k,f} = \mathbbm{1} \left(\boldsymbol{p}_f^k > \Delta\right).
    \end{equation*}

    \noindent However, in this case, since the absolute values of the cluster membership (e.g. distances to centroids) are not controlled, a good value of the threshold parameter would differ across clusters and training runs. 
    
    One way to address this challenge is to select feature groups based on percentiles instead of thresholds (i.e., choose the top 5 \% features for each cluster). However, in this case the sizes of the clusters end up having a fixed size, which constrains the ability of the model to learn divers groups. 
    
    Another way to address this is to softmax the membership values before applying the threshold
    \begin{equation*}
        \boldsymbol{M}_{k,f} = \mathbbm{1} \left(\operatorname{Softmax}\boldsymbol{p}_f^l > \Delta\right).
    \end{equation*}

    \noindent In practice, we found it very hard to tune with $\Delta \gtrsim 0.4$ producing empty clusters and $\Delta \approx 0.3$ producing full clusters. This is probably due to the distances being similar to each other in the early stages of training. 
    
    A different way to control the magnitude of the membership values is to normalize by the maximum as \equationref{eq:soft_membership}. We found this approach to be easier to tune but still suffer from producing overly complete clusters. An additional practical modification that we apply is to first assign every feature to a cluster using an $\operatorname{argmax}$ like in hard clustering, then assign every cluster with a feature, and finally apply the soft membership function. These additional steps ensure that every feature is assigned and that every cluster is not empty, resulting in improved stability and easier hyperparameter search.

\subsection{Cluster Initialization}
\label{app:initialization}

    One of the important design questions for clustering algorithms is their initialization. We consider initialization from k-means\texttt{++} and expert-defined feature groups. 

    k-means\texttt{++} is a classic algorithm used to select centroids for K-means \citep{arthur2006k}. First, one feature is chosen uniformly at random as a cluster centroid. Then, the next centroid is chosen randomly among the other features with a probability proportional to the inverse square distance to the closest centroid. This process is continued until all centroids are chosen. The algorithm was designed for K-means and is therefore directly applicable in our case as well. For GMM, first we use k-means\texttt{++} to initialize centroids, then we choose the closest centroid to each feature to define the initial groups, and finally we calculate the Gaussian means and covariance using these groups.
    
    Expert-defined feature groups for ICU data were proposed by \citet{kuznetsova2023importance}. In this case, the expert group features are given directly as input to the model, so we calculate the K-means centroids or GMM means and covariances directly using these splits. Using expert-defined feature groups already improves downstream performance, so initializing from them can help discover even better learned feature groups.

\subsection{Adaptive Number of Clusters}
\label{app:adaptive_cl_num}

\revB{As discussed in} \sectionref{sec:discandlims} \revB{, the number of clusters that the proposed algorithm uses is technically fixed, which is a limitation. Note, however, that the algorithm does have some flexibility: it can use fewer clusters by setting some of them to be empty like in the MIMIC Decompensation case (see }\figureref{fig:sankey_md_kf}\revB{).

To choose the best number of clusters we use a simple hyperparameter search. Other methods that could be used for that or for adjusting the number of clusters dynamically include stability-based method} \citep{lange2004stability}, the elbow method \citep{thorndike1953belongs}, and the Dirichlet processes \citep{ferguson1973bayesian}.\revB{

Stability-based methods would choose the number of clusters based on the metrics that characterize the stability of clustering between bootstrapped versions of the dataset. This does not align with our main goal of improving the downstream performance of the underlying predictive model. 

The elbow method could be used with the downstream performance metrics directly. However, in practice, it would require fixing all other hyperparameters. Our ablation study shows that performance can vary based on them (see} \appendixref{app:ablations})\revB{, so fixing one set of parameters for a different number of clusters might not lead to the best performance. At the same time, performing a sufficient hyperparameter search already allows us to draw plots similar to those used by the elbow method (subfigures (j) in }\tableref{tab:ablation_hirid_circ_gmm-soft,tab:ablation_hirid_mort_kmeans,tab:ablation_hirid_mort_kmeans-fuzzy,tab:ablation_mimic_decomp_gmm,tab:ablation_mimic_mort_kmeans}). We show results for the application of the elbow method in \appendixref{app:elbow}. \revB{

More advanced methods like Dirichlet processes would require a big modification to the pipeline and would likely be challenging to stabilize within our dynamic clustering context, which is why we leave this to future work.}

\section{Experiment Details}

\subsection{Metrics}
\label{app:metrics}

    We evaluate feature groups according to either how good the clusters are themselves or how they influence downstream performance. For the former, we use both internal metrics (do not require knowledge of the true underlying feature groups) and external metrics (vice versa).

    The true feature grouping is known only for synthetic data. In this case, we can use external metrics:
    \begin{itemize}
        \item \texttt{Adjusted rand index} (ARI) represents the number of permutations of points needed to recover the true clustering from the learned one \citep{hubert1985comparing}. It takes values between -1 and 1, higher the better (0 corresponds to random assignment and 1 corresponds to a perfect match). More formally, if the rand index $\operatorname{RI}$ is the proportion of pairs of points that are correctly assigned, and $\mathbb{E}(\operatorname{RI})$ is expected to be $\operatorname{RI}$ under a random permutation, then \begin{equation*}
            \operatorname{ARI} = \dfrac{\operatorname{RI} - \mathbb{E}(\operatorname{RI})}{\max (\operatorname{RI}) - \mathbb{E}(\operatorname{RI})}.
        \end{equation*}
        \item \texttt{Normalized mutual information} (NMI) is derived from information theory and represents the mutual information between true and predicted clusterings \citep{vinh2009information}. It takes values between zero and one; the higher, the better. Denoting $I$ mutual information, and $H$ as entropy,
        \begin{equation*}
            \operatorname{NMI}(u,v) = \dfrac{I(u,v)}{\sqrt{H(u)H(v)}}.
        \end{equation*}
    \end{itemize}

    Without access to the true feature grouping, the quality of the clusters can be evaluated based on the distances between the clusters and points (internal metrics). For example, we use intercluster and intracluster distances for regularization \equationref{eq:reg_loss}. Additionally, for synthetic data, we compute the

    \begin{itemize}
        \item \texttt{Silhoette score} \citep{rousseeuw1987silhouettes}. Denoting mean distances between a feature $f$ to the points in the same cluster $\mathcal{D}_{\text{same}}$ and to the points in the nearest cluster $\mathcal{D}_{\text{nearest}}$
        \begin{equation*}
            S = \mathbb{E}_i \left[\dfrac{\mathcal{D}_{\text{nearest}} - \mathcal{D}_{\text{same}}}{\max\left(\mathcal{D}_{\text{nearest}}, \mathcal{D}_{\text{same}}\right)}\right].
        \end{equation*}
    \end{itemize}

    For downstream performance, we consider metrics that are commonly used in the field and suit highly imbalanced tasks on ICU data \citep{yeche2021, harutyunyan2019multitask}. Specifically, for classification, we use \texttt{area under the precision-recall curve} (AUPRC) and \texttt{area under the receiver-operator curve} (AUROC).

\subsection{Synthetic Data}
\label{app:training_details_synth}
    
    We generate synthetic data by sampling 10,000 examples from Gaussian processes with $6$ features and the length of $20$. We use the RBF kernel for all features with length scales $\{1, 2, 4, 8, 1, 2\}$ and amplitudes $\{0.5, 1.0, 3.5, 0.5, 0.5, 0.5\}$.

    \begin{figure}[htb]
        \floatconts
        {fig:synth_samples}
        {\caption{Samples from the synthetic dataset.}}
        {%
            \subfigure[Sample 0]{%
                \includegraphics[width=0.3\linewidth]{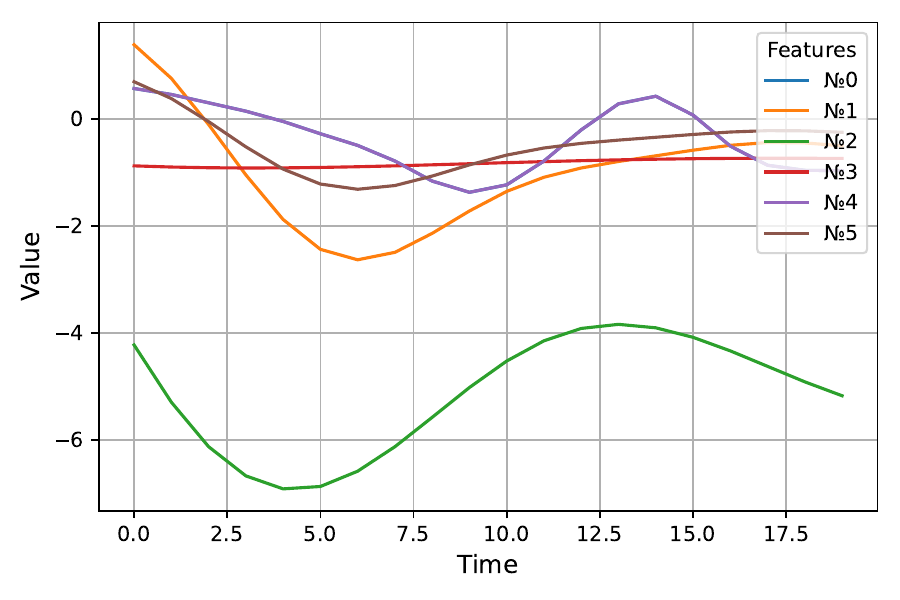}
            }
            \hfill
            \subfigure[Sample 1]{%
                \includegraphics[width=0.3\linewidth]{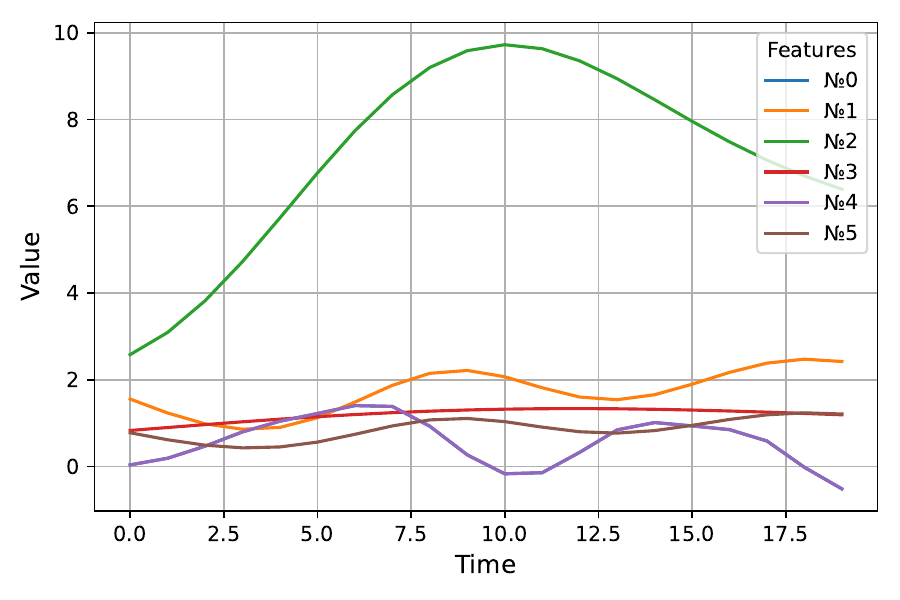}
            }
            \hfill
            \subfigure[Sample 2]{%
                \includegraphics[width=0.3\linewidth]{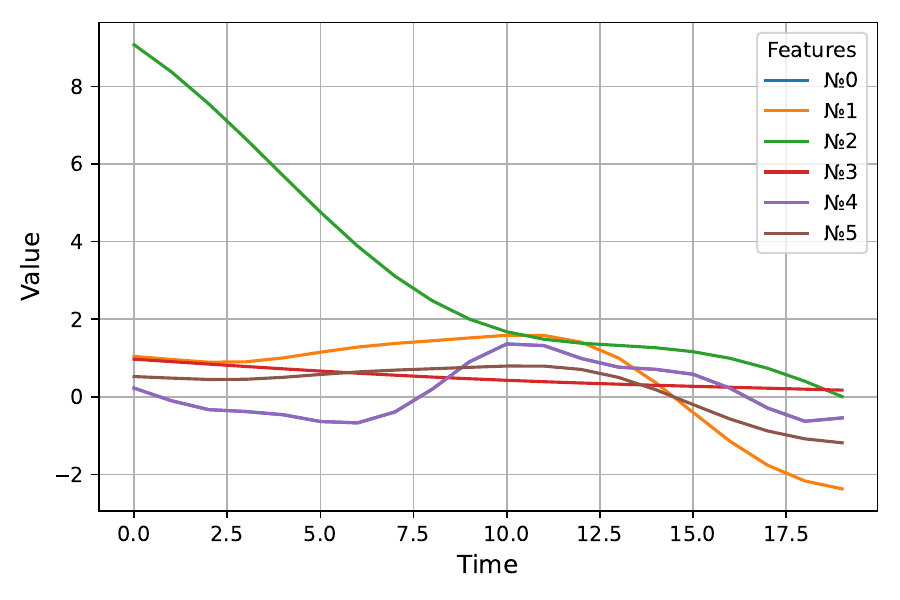}
            }
        }
    \end{figure}

    The data transformations for static clustering are formally defined as 
    \begin{gather*}
        \mathcal{D}_{\text{flat}} = \left\{\concat_{i,t} \left(\boldsymbol{x}^{i}_{t,f} \ | \ y^i\right)\right\}_{f=1}^F,\\
        \mathcal{D}_{\text{time mean}} = \left\{\concat_{i} \left(\frac{1}{D} \sum_{t} \boldsymbol{x}^{i}_{t,f} \ | \ y^i\right)\right\}_{f=1}^F,\\
        \mathcal{D}_{\text{sample mean}} = \left\{\concat_{t} \left(\frac{1}{T} \sum_{i} \boldsymbol{x}^{i}_{t,f} \ | \ y^i\right)\right\}_{f=1}^F,\\
        \mathcal{D}_{\text{full mean}} = \left\{\frac{1}{DT} \sum_{i,t} \boldsymbol{x}^{i}_{t,f} \ | \ y^i\right\}_{f=1}^F.
    \end{gather*}
    Here, $\concat$ and $|$ are concatenation operations.

    We train the model for 1000 epochs with a learning rate 0.002, an early stopping patience of 10 and a batch size of 5000. Clustering is done using Hard K-means with k-means\texttt{++} initialization and no regularization. FT-Transformer has 6 hidden units, 2 heads, and a depth of 1. Further details can be found in the source code. 

\subsection{Medical Data}
\label{app:training_details_med}

    \begin{table*}[htbp]
        \floatconts
        {tab:icu_auroc_hard}%
        {\caption{\textbf{Performance benchmark for different grouping schemes} (AUROC \textuparrow) on online and offline classification tasks using ICU datasets. Mean and standard deviation are reported over three training runs with different random seeds. Best results are highlighted in bold. Learned groupings perform comparably to expert-defined ones. Reference results for prior groups (Prior) and models without feature grouping (None) are taken from~\citet{kuznetsova2023importance}.}}%
        {
        {%
        \small
        \begin{tabular}{@{}ccccccc@{}}
            \toprule
            \multicolumn{3}{c}{Feature Grouping} & \multicolumn{2}{c}{HIRID} & \multicolumn{2}{c}{MIMIC-III} \\ \cmidrule(lr){1-3} \cmidrule(lr){4-5} \cmidrule(lr){6-7}
            \mc{Type} & \mc{Membership} & \mc{Algorithm} & \mc{Circ} & \mc{Mort}  & \mc{Decomp} & \mc{Mort} \\ \midrule
            None      & --   & --            & 0.911 \textpm \ 0.001 & 0.905 \textpm \ 0.002 & \textbf{0.916} \textpm \ 0.001 & 0.858 \textpm \ 0.002 \\
            Prior     & Hard & Type          & 0.910 \textpm \ 0.002 & 0.901 \textpm \ 0.003 & 0.914 \textpm \ 0.001 & 0.860 \textpm \ 0.002 \\
            Prior     & Hard & Organ         & \textbf{0.916} \textpm \ 0.0003 & \textbf{0.910} \textpm \ 0.003 & 0.914 \textpm \ 0.001 & \textbf{0.861} \textpm \ 0.002 \\\addlinespace
            Learned   & Hard & K-means       & 0.912 \textpm \ 0.0008 & \textbf{0.912} \textpm \ 0.004 & \textbf{0.917} \textpm \ 0.0002 & \textbf{0.863} \textpm \ 0.002 \\
            Learned   & Hard & Fuzzy K-means & 0.907 \textpm \ 0.0004 & 0.908 \textpm \ 0.003 & \textbf{0.916} \textpm \ 0.0009 & \textbf{0.861} \textpm \ 0.004 \\
            Learned   & Hard & GMM           & 0.912 \textpm \ 0.0008 & 0.900 \textpm \ 0.004 & \textbf{0.917} \textpm \ 0.0009 & 0.858 \textpm \ 0.001 \\\addlinespace
            Learned   & Soft & Fuzzy K-means & \textbf{0.911} \textpm \ 0.0022 & 0.902 \textpm \ 0.002 & \textbf{0.916} \textpm \ 0.0019 & \textbf{0.863} \textpm \ 0.002 \\
            Learned   & Soft & GMM           & \textbf{0.912} \textpm \ 0.0012 & 0.906 \textpm \ 0.001 & \textbf{0.915} \textpm \ 0.0024 & \textbf{0.861} \textpm \ 0.003 \\ \hline
        \end{tabular}
        }}
    \end{table*}
    
\subsubsection{Task Definitions}
\label{app:tasks}

    We consider clinically relevant tasks defined for the HiRID and MIMIC-III datasets \citep{yeche2021,  harutyunyan2019multitask}. These can be split into static (prediction is made once per sample) and dynamic tasks (prediction is made continuously for the sample); and into classification and regression tasks.

    On HiRID we consider
    \begin{itemize}
        \item \texttt{Circulatory failure} (Circ), a dynamic binary classification task predicting whether a circulatory failure will occur in the next 8 hours;
        \item \texttt{Mortality} (Mort), a static binary classification predicting whether the patient will pass away at the end of the admission using observation from the first 24 hours;
    \end{itemize}

    \noindent On MIMIC-III we consider
    \begin{itemize}
        \item \texttt{Decompensation} (Decomp), a dynamic binary classification task predicting whether a patient will pass away in the next 24 hours;
        \item \texttt{Mortality} (Mort), a static binary classification predicting whether the patient will pass away at the end of the admission using observation from the first 24 hours.
    \end{itemize}

    More detailed definitions can be found in \citet{kuznetsova2023importance, yeche2021, harutyunyan2019multitask}.

\subsubsection{Setup}
\label{app:setup}

    We train using 1 to 4 NVidia RTX2080TI GPUs with Xeon E5-2630v4 CPUs. The batch size ranges from 1 to 16 depending on the model size and hardware capacity. The training time is between 5--22 hours. We use Adam optimizer and early stopping with patience of 10 epochs \citep{diederik2014adam}.  The models are implemented in PyTorch \citep{paszke2019pytorch} and scikit-learn \citep{scikit-learn}.
    
    We perform a hyperparameter search across the number of clusters, type of embedding extraction function $U$, EMA coefficients, regularization type and coefficient, cluster initialization type, fuzzy coefficient (for fuzzy K-means), covariance matrix type (for GMM), membership coefficient (for soft membership models). The parameters for the underlying embedding and sequence model are taken from \citet{kuznetsova2023importance, yeche2021}. Further details can be found in the project repository.

    For t-SNE visualizations, we use a perplexity of 10 for HiRID and 5 for MIMIC with a random seed of 42 \citep{vandermaaten08a_tsne}.

\subsubsection{Additional Results}\label{app:add_res}

    \paragraph{Additional metrics} Similarly to \tableref{tab:icu_auprc_hard}, we present the downstream performance now measured by AUROC in \tableref{tab:icu_auroc_hard}. Our conclusions are consistent for both metrics. 
    
    We provide p-values for the paired Student t-test~\citep{student1908probable} on the null hypothesis that a model with grouping performs better than the reference model without grouping in terms of AUPRC in \tableref{tab:pvalues}.
    \begin{table}[htbp]
        \floatconts
        {tab:pvalues}%
        {\caption{P-values for comparison of various models to the reference model without feature grouping by AUPRC. Values under $0.05$ are highlighted in bold.}}%
        {
            \resizebox{\linewidth}{!}{
            \begin{tabular}{@{}lcccc@{}}
                \toprule
                Grouping & HiRID Circ & HiRID Mort & MIMIC Decomp & MIMIC Mort \\
                \midrule
                Prior by Type & \textbf{0.0348} & 0.1745 & 0.0774 & 0.1890 \\
                Prior by Organ & \textbf{0.0167} & 0.1045 & \textbf{0.0109} & 0.0774 \\
                Learned Hard K-means & 0.3651 & \textbf{0.0187} & 0.0530 & 0.0927 \\
                Learned Hard Fuzzy K-means & 0.4335 & 0.0548 & 0.8361 & 0.1275 \\
                Learned Hard GMM & 0.0827 & \textbf{0.0151} & 0.0620 & 0.4275 \\
                Learned Soft Fuzzy K-means & 0.1719 & 0.0982 & 0.0833 & 0.2660 \\
                Learned Soft GMM & \textbf{0.0286} & 0.7364 & \textbf{0.0460} & 0.2433 \\
                \bottomrule
            \end{tabular}
            }
        }
    \end{table}

    \revC{In addition, we provide an F-score} \citep{goutte2005probabilistic} \revC{and the Matthews correlation coefficient (MCC)} \citep{chicco2023matthews} \revC{for the mortality prediction experiments. These metrics were proposed to address limitations of AUROC and AUPRC in evaluating performance on imbalanced tasks, but are not yet widely used in ICU research} \citep{chicco2023matthews,yadav2025failure}. 
    \begin{table}[htbp]
        \floatconts
        {tab:mcc_f1_mort}
        {\caption{MCC and F1 scores for various clustering algorithms trained for mortality prediction.}}%
        {        
            \resizebox{\linewidth}{!}{%
            \begin{tabular}{@{}lllll@{}}
            \toprule
            \multicolumn{1}{c}{Algorithm} & \multicolumn{2}{c}{MCC} & \multicolumn{2}{c}{F1} \\ \cmidrule(l){2-3} \cmidrule(l){4-5}
            \multicolumn{1}{c}{} & \multicolumn{1}{c}{MIMIC Mort} & \multicolumn{1}{c}{HiRID Mort} & \multicolumn{1}{c}{MIMIC Mort} & \multicolumn{1}{c}{HiRID Mort} \\ \midrule
            K-means & 0.4467 \textpm \ 0.0097 & 0.5838 \textpm \ 0.0030 & 0.5142 \textpm \ 0.0090 & 0.6171 \textpm \ 0.0012 \\
            Fuzzy K-means & 0.4538 \textpm \ 0.0083 & 0.5837 \textpm \ 0.0071 & 0.5113 \textpm \ 0.0023 & 0.6139 \textpm \ 0.0031 \\
            GMM & 0.4458 \textpm \ 0.0070 & 0.5804 \textpm \ 0.0069 & 0.5046 \textpm \ 0.0074 & 0.6118 \textpm \ 0.0037 \\
            Soft Fuzzy GMM & 0.4453 \textpm \ 0.0041 & 0.5661 \textpm \ 0.0065 & 0.5104 \textpm \ 0.0072 & 0.5976 \textpm \ 0.0050 \\
            Soft GMM & 0.4459 \textpm \ 0.0106 & 0.5797 \textpm \ 0.0074 & 0.5110 \textpm \ 0.0075 & 0.6104 \textpm \ 0.0102 \\ \bottomrule
            \end{tabular}%
            }
        }
    \end{table}
    
    \paragraph{Learning dynamics} We investigated the cluster learning dynamics and describe them for various models using a fixed seed in \figureref{fig:sankey_hc_gs,fig:sankey_hm_k,fig:sankey_hm_kf,fig:sankey_md_kf,fig:sankey_mm_k,fig:sankey_mm_kf}. Generally, we observe that in several cases the models keep the groupings constant from initialization. We believe this to be the case of the exploration--exploitation trade-off: the earlier the clustering is set, the more time and signal passes through the other parts of the architecture, potentially improving the downstream performance. To control the trade-off, we can adjust the regularization and EMA coefficients. It also seems that the Fuzzy K-means model is particularly suitable for exploration.

    \begin{figure}[htbp]
        \floatconts
        {fig:cl_sizes_att}
        {\caption{Dynamics of cluster sizes for the Fuzzy K-means trained for circulatory prediction on HiRID.}}
        {\includegraphics[width=0.65\linewidth]{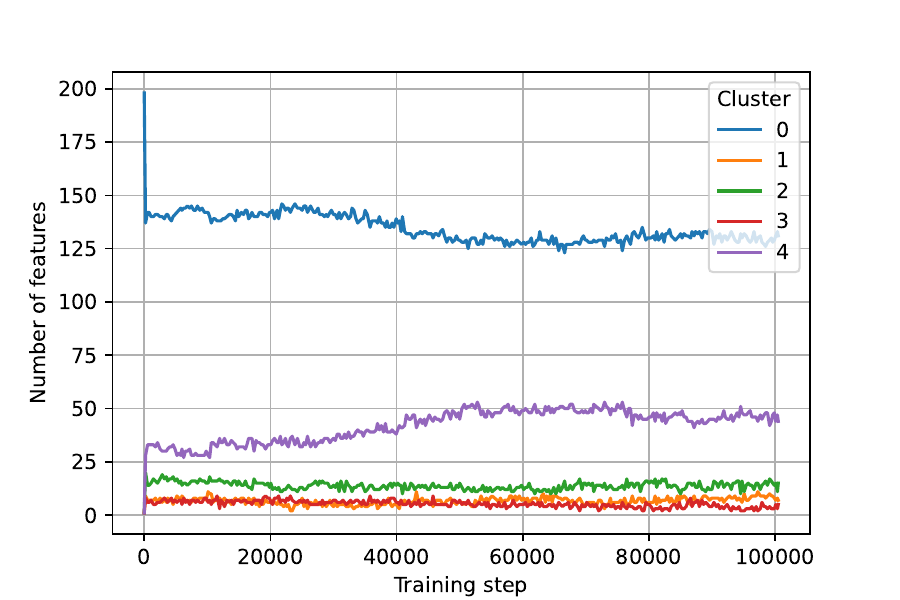}}
    \end{figure}

    \paragraph{Impact of initialization} \revB{As we discussed in }\sectionref{app:initialization}\revB{, our method can use the prior feature grouping as the initialization, as well as initialize from the k-means\texttt{++} algorithm. From the ablation studies, we see that k-means\texttt{++} performs better for Soft GMM trained on HiRID circulatory failure, K-means trained on HiRID mortality, and K-means trained on MIMIC mortality }\tableref{tab:ablation_hirid_circ_gmm-soft_initialization,tab:ablation_hirid_mort_kmeans_initialization,tab:ablation_mimic_mort_kmeans_initialization}\revB{. Priors perform better for Fuzzy K-means trained on HiRID mortality and GMM trained on decompensation }\tableref{tab:ablation_hirid_mort_kmeans-fuzzy_initialization,tab:ablation_hirid_mort_mort_gmm_initialization}\revB{. This demonstrates that our approach does not depend entirely on a good initialization from priors, but rather can leverage them to achieve better performance if algorithmic initialization does not work well.}

    \paragraph{Elbow method}\label{app:elbow} \revB{Here we investigate whether the Elbow method can be effectively used for the selection of the optimal number of clusters. We fix the hyperameters used by the best performing models on HiRID and MIMIC mortality prediction and vary the number of clusters. Resulting performance is shown in } \figureref{fig:elbow_mort}. \revB{We see that the difference in performance is comparable to the that achieved by altering other hyperparameters } \tableref{tab:ablation_hirid_mort_kmeans,tab:ablation_mimic_mort_kmeans}. \revB{This suggests that the original approach of searching for the best joint set of hyperparameters is more beneficial for the downstream performance.}

    \begin{figure}[htbp]
        \floatconts
        {fig:elbow_mort}
        {\caption{Elbow method for optimal number of clusters for mortality prediction.}}
        {%
            \subfigure[HiRID]{%
                \includegraphics[width=0.45\linewidth]{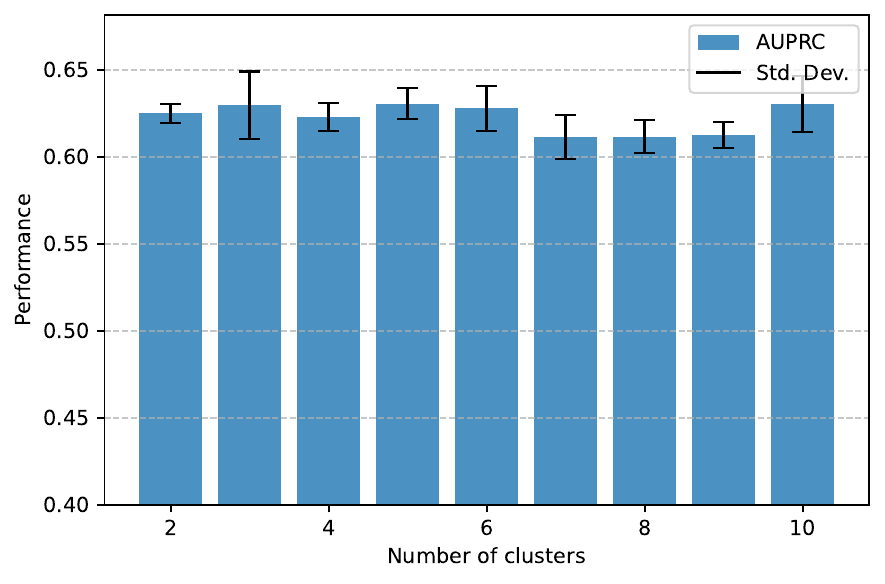}
            }
            \hfill
            \subfigure[MIMIC]{%
                \includegraphics[width=0.45\linewidth]{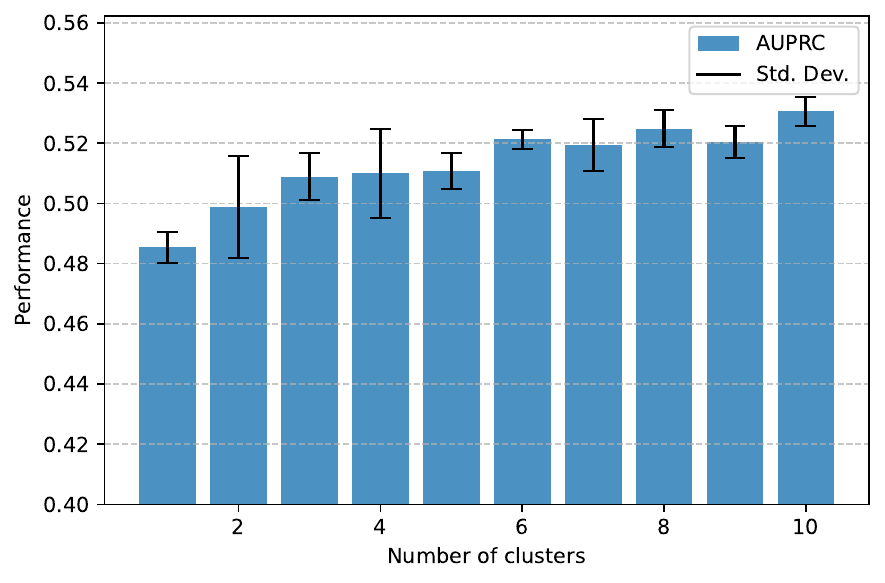}
            }
        }
    \end{figure}

    \begin{figure*}[htbp]
        \floatconts
        {fig:attention_hc}
        {\caption{Attention weights between clusters and over time before the event produced by Fuzzy K-means trained for circulatory failure prediction on HiRID.}}
        {%
            \subfigure[Sample 90]{%
                \includegraphics[width=0.3\linewidth]{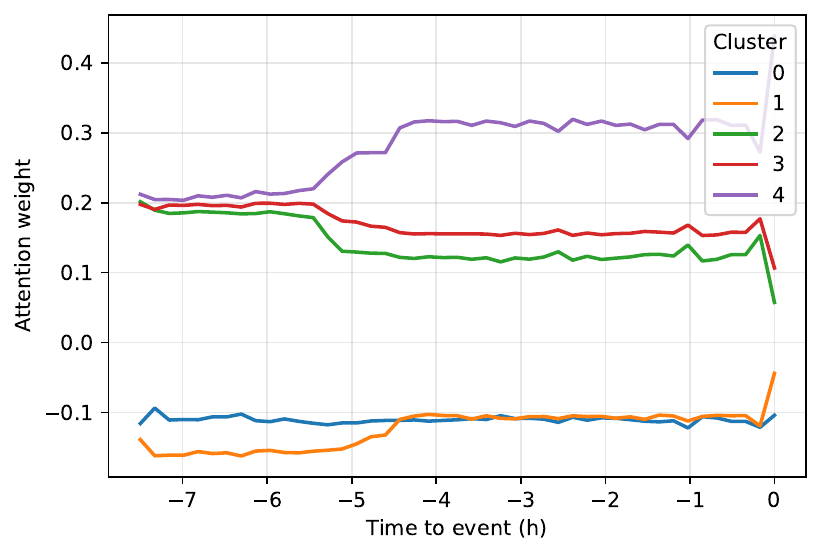}
            }
            \hfill
            \subfigure[Sample 120]{%
                \includegraphics[width=0.3\linewidth]{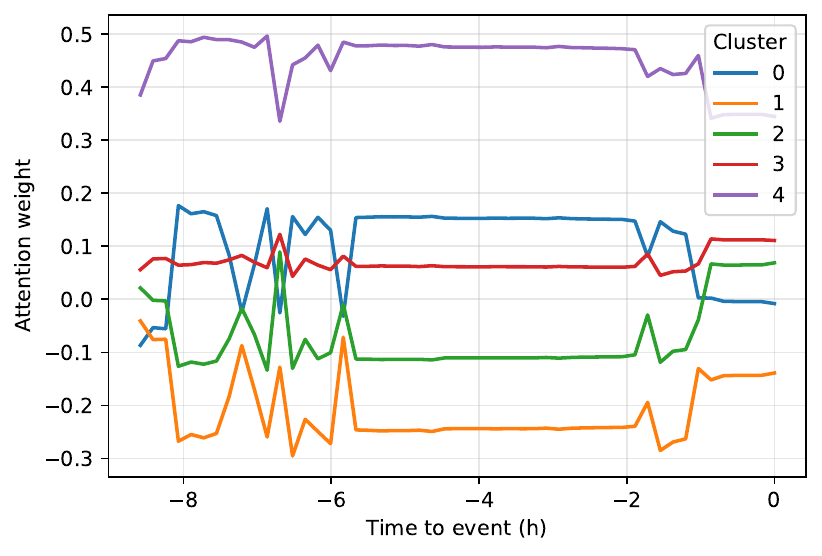}
            }
            \hfill
            \subfigure[Sample 336]{%
                \includegraphics[width=0.3\linewidth]{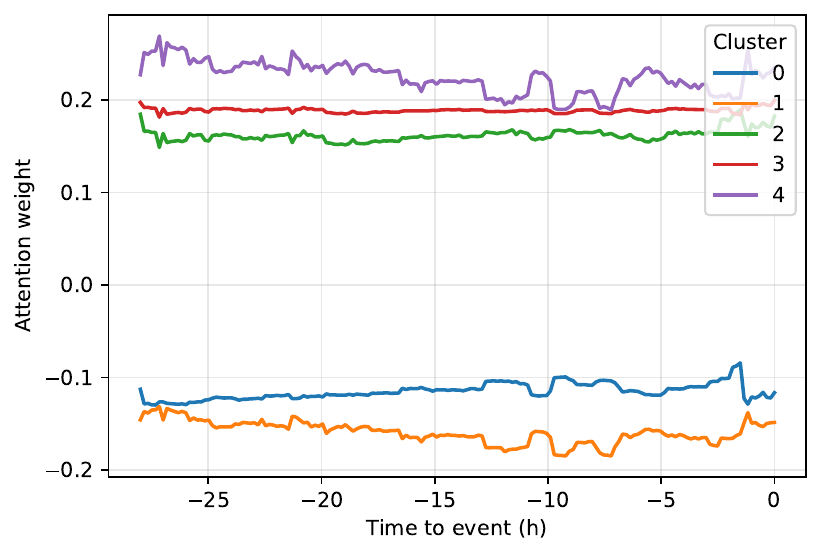}
            }
        }
    \end{figure*}

    \paragraph{Attention weights}  Our method includes an additional explainability mechanism based on attention weights, which is active when attention aggregation is used. This is the case for the best Fuzzy K-means trained on the HiRID dataset for circulatory failure prediction (other models for this task use mean aggregation, see \tableref{tab:icu_auprc_hard}). We visualize these weights over groups and the time before the events for three randomly selected samples in \figureref{fig:attention_hc}. We see that the model consistently ranks the clusters: cluster 4 has the highest weight, followed by clusters 3, 4, and 0, while cluster 1 gets the lowest. The size of the clusters suggests that this ranking does not simply give more weight to the larger clusters (see \figureref{fig:cl_sizes_att}). We analyze the feature assignments shown in \tableref{tab:grouping_hc_kf}: cluster 1 seems to combine variables on acute neurological event and resuscitation; cluster 0 is the largest and contains a mix of diverse features; cluster 2 contains a combination of variables related to sepsis and acute pancreatic conditions; cluster 3 might indicate acute respiratory distress (ARDR) if the underlying ventilator support (AWPmean) and deep sedation (deep sedation) variables are high, it also curiously seemingly unrelated coagulation variable (FVII); cluster 4 contains many dynamic variables that might indicate patient crashing (key vitals: RR, HR; key labs: a\_Lac, CRP). Attention therefore places more weight on clusters that indicate patient crash and respiratory distress, while placing less weight on a large group of mixed variables and a group specifically targeting acute neurological conditions. Although insightful, this interpretation should be viewed with caution because the model underperforms other clustering algorithms in terms of the downstream metric.

\subsubsection{Learned Groups}
\label{app:learned_groups}
    
    The feature groups learned on MIMIC for decompensation and mortality prediction are shown in \tableref{tab:grouping_md_k,tab:grouping_mm_k,tab:grouping_mm_kf}. The feature groups learned on HiRID for circulatory failure and mortality prediction in \tableref{tab:grouping_hc_gs,tab:grouping_hm_k,tab:grouping_hm_kf}. We present results for the best model if one stands out clearly, and for the top two models if multiple models are competitive (see \tableref{tab:icu_auprc_hard}). We compare (1) the learned groups with the expert-defined groups, (2) the groups learned on the same dataset but for different tasks, and (3) the groups learned by different models on the same dataset and the same task. 

    \paragraph{MIMIC decompensation} Fuzzy K-means shows the best downstream performance for this task. As discussed previously (see \sectionref{sec:med_results}), from the learning dynamics (see \figureref{fig:sankey_md_kf}) we see that the model initially splits the features into four groups and eventually learns to assign all the features to one cluster (see \tableref{tab:grouping_md_k}). This replicates the best prior grouping that also just assigns all features to a single group (see \tableref{tab:icu_auprc_hard}).

    \paragraph{MIMIC mortality} The best two models for this task are K-means and fuzzy K-means. Both models produce eight clusters, with four or five of them containing a single feature (see \tableref{tab:grouping_mm_k,tab:grouping_mm_kf})). The K-means appears to select groups that focus on identifying particular conditions such as shock (cluster 0), inflammation (cluster 1), acidoses (cluster 2), etc. or baseline state of the patient such as their constitution (cluster 7) and organ function (clusters 1 and 6) \tableref{tab:grouping_mm_k}). The Fuzzy K-means produces clusters of similar sizes. Some of them are quite similar to what K-means produces: cluster 1 is exactly the same (respiratory rate), cluster 5 (systolic blood pressure) and cluster 6 (mean blood pressure) describe blood pressure, clusters 6 (weight) and cluster 3 (height) describe patient's size. Cluster 5 could be related to septic shock: low diastolic blood pressure and high glucose could be its indicators. Other clusters seem more mixed: cluster 0 contains variables that describe vital signs and neurological status, cluster 8 describes gas exchange and hemodynamics. These clusters (and the dataset in general) do not contain many variables, so we do not see an obvious pattern here. 
    
    Surprisingly, the models divide some naturally related variables, such as respiratory and hemodynamic features, into different groups (e.g., K-means assigns oxygen saturation and respiratory rate to clusters 0 and 5; blood pressure and heart rate to clusters 0 and 7). It is not immediately clear what the benefit is, but perhaps splitting the variables allows capturing the relation between different organ systems within one cluster for multiple clusters, enabling better characterization of certain conditions. Specifically, the model might need the features characterizing the respiratory system in both groups that describe coma and hypoxia.
    
    \paragraph{MIMIC overall} We notice that the models select only one cluster for decompensation and eight for both models for mortality. We speculate that this difference in behavior occurs from the nature of the tasks. Decompensation is an online task, where the model should capture the rapid deterioration of the patient's condition. Mortality is an offline prediction task, where the model is supposed to assess the patient's general status over the first two days and relate it to the outcome after the whole, possibly multiday ICU stay. This means that models trained for decompensation prediction might not rely on the predictive groups, but rather on how quickly all features change; whereas models trained for mortality prediction would rely on identifying whether the patient has a particular condition (e.g., shock) or if they are likely to develop it based on their overall status (e.g., weight, respiratory support, etc.).

    \paragraph{HiRID circulatory failure} The soft GMM model shows the best downstream results for this task. In \tableref{tab:grouping_hc_gs} we see three clusters that single out three laboratory measurements: creatine kinase (cluster 1), alkaline phosphatase (cluster 2), and creatine kinase-MB (cluster 3). These could be indicators of specific critical events: a spike in creatine kinase could suggest rhabdomyolysis, creatine kinase-MB could indicate biliary sepsis, and alkaline phosphatase could point to acute myocardial infarction. However, other variables could also indicate these conditions, so it is not completely certain that the model uses these variables to predict these specific conditions. The other two clusters are quite large, containing the rest of over 200 variables from HiRID. Interpreting them also becomes harder, with both containing vitals, laboratory analysis, and treatments for various organ systems at the same time.
    
    \paragraph{HiRID mortality} For mortality, the best downstream results are produced by K-means and Fuzzy K-means models. Both produce a similar number of clusters (4 and 5 respectively) that have somewhat similar sizes (except for cluster 2 of Fuzzy K-means) (see \tableref{tab:grouping_hm_k,tab:grouping_hm_kf}). Here, as in the circulatory failure case, interpreting the clusters is not straightforward. Cluster 0 from K-means, for example, contains many hemodynamic variables (e.g., WBC, ferritin, Ca, etc.) but at the same time contains electrolytes (e.g., Na, K); cluster 1 contains numerous vital signs (e.g., HR, SpO2), but at the same time contains treatments (e.g., antiviral, immunosuppression, milrinone). Similarly, for Fuzzy K-means, cluster 1 contains coagulation factors (FII, FV, FVII, FX) but at the same time cardiologic variables (e.g., ABP, ZVD), treatments (e.g., antibiotics, cortisol) and other variables. The only small feature cluster here, cluster 2, could be related to cardiopulmonary management during resuscitation.

    \paragraph{HiRID overall} Across tasks, we find that models use a similar number of clusters for HiRID. Due to the large number of variables, they become difficult to access. Due to computational cost, we considered only up to 8, which seems too few to the over 200 features that HiRID has. 
    
    Consistent with MIMIC, we find that the smaller clusters seem to track specific acute conditions. This finding could help develop new feature groups tailored to specific tasks or patient cohorts.

\subsubsection{Prior Feature Groups}
\label{app:prior_groups}
    
    The prior grouping by organ system for HiRID and MIMIC-III is presented in \tableref{tab:prior_group_organ_hirid,tab:prior_group_organ_mimic}, respectively. Prior grouping by measurement type is shown in \tableref{tab:prior_group_type_hirid,tab:prior_group_type_mimic}. The tables are reproduced with minor modifications from \citet{kuznetsova2023importance}. For a description of individual variables, see \citet{hyland2020early} and \citet{MIMIC-III}.

\subsubsection{Ablation Studies}
\label{app:ablations}

    Ablation studies are shown in \tableref{tab:ablation_hirid_circ_gmm-soft,tab:ablation_hirid_mort_kmeans,tab:ablation_hirid_mort_kmeans-fuzzy,tab:ablation_mimic_decomp_gmm,tab:ablation_mimic_mort_kmeans}. We discuss them in \sectionref{sec:med_results}.

    \revD{Generally, we find that a number of clusters of 3--5, a small EMA rate of 0.25, and a regularization weight of 0.0001 generally leads to good performance and can be used as a good first set of parameters. However, the choice of the initialization, architecture functions, and regularizer type depends on the underlying dataset and task, and should be optimized. To balance flexibility and simplicity, we suggest starting with k-means\texttt{++}, bias unification, attention merging, and a hard regularizer.}

    \revE{The training time  of the best-performing runs for circulatory failure prediction on HiRID was on average 348 minutes without grouping, 403 minutes with prior organ grouping, 368 minutes for K-means, 299 minutes for Fuzzy K-means, 544 minutes for Soft Fuzzy K-means, 305 minutes for GMM, and 303 minutes for Soft GMM. For other tasks the training times of different models were similarly close, averaging to around 96 minutes for mortality prediction on HiRID, 359 for decompensation prediction on MIMIC, and 95 for mortality prediction on MIMIC (note that the number of GPUs and the batch size varies between tasks).}

    \begin{table*}[htbp]
        \floatconts
        {tab:grouping_md_k}%
        {\caption{Feature grouping learned by K-means model of MIMIC features for decompensation prediction. Interpreted as no grouping is necessary.}}%
        {
            \renewcommand{\arraystretch}{1.3}
            \footnotesize
            \begin{tabular}{@{}p{0.28\linewidth}p{0.68\linewidth}@{}}
                \toprule
                Group name & Features \\ \midrule
                Cluster 0: all features & Capillary refill rate, Glasgow coma scale eye opening, Glasgow coma scale motor response, Glasgow coma scale total, Glasgow coma scale verbal response,   Diastolic blood pressure, Fraction inspired oxygen, Glucose, Heart Rate, Mean blood pressure, Oxygen saturation, Respiratory rate, Systolic blood pressure, Temperature,  pH \\
                \bottomrule
            \end{tabular}
        }
    \end{table*}

    \begin{table*}[htbp]
        \floatconts
        {tab:grouping_mm_k}%
        {\caption{Feature grouping learned by K-means model of MIMIC features for mortality prediction. Interpreted as grouping by baseline status and monitoring of certain conditions.}}%
        {
            \renewcommand{\arraystretch}{1.3}
            \footnotesize
            \begin{tabular}{@{}p{0.28\linewidth}p{0.68\linewidth}@{}}
                \toprule
                Group name & Features \\ \midrule
                Cluster 0: ``coma status'' & Glasgow coma scale eye opening, Glasgow coma scale verbal response, Time, Mean blood  pressure, Oxygen saturation \\
                Cluster 1: ``respiratory status'' & Respiratory rate \\
                Cluster 2: ``inflammation status'' & Temperature \\
                Cluster 3: ``acidosis status'' & pH \\
                Cluster 4: ``hypoxia status'' & Glasgow coma scale motor response, Height, Fraction inspired oxygen, Glucose \\
                Cluster 5: ``systolic function status'' & Systolic blood pressure \\
                Cluster 6: ``baseline constitution'' & Weight \\
                Cluster 7: ``hemodynamic stability'' & Capillary refill rate, Glasgow coma scale total, Diastolic blood pressure, Heart Rate \\
                \bottomrule
            \end{tabular}
        }
    \end{table*}

    \begin{table*}[htbp]
        \floatconts
        {tab:grouping_mm_kf}%
        {\caption{Feature grouping learned by Fuzzy K-means model of MIMIC features for mortality prediction. Interpreted as grouping by baseline status and monitoring of certain conditions.}}%
        {
            \renewcommand{\arraystretch}{1.3}
            \footnotesize
            \begin{tabular}{@{}p{0.28\linewidth}p{0.68\linewidth}@{}}
                \toprule
                Group name & Features \\ \midrule
                Cluster 0: ``mixed vitals and neurological status'' & Capillary refill rate, Glasgow coma scale eye opening, Glasgow coma scale total, Glasgow coma scale verbal response, Time, Heart Rate, Temperature \\
                Cluster 1: ``respiratory status'' & Respiratory rate \\
                Cluster 3: ``baseline statury'' & Height \\
                Cluster 4: ``motor status'' & Glasgow coma scale motor response \\
                Cluster 5: ``septic shock'' & Diastolic blood pressure, Glucose \\
                Cluster 6: ``circulatory status'' & Mean blood pressure \\
                Cluster 7: ``respiratory support'' & Fraction inspired oxygen \\
                Cluster 8: ``mixed gas exchange and hemodynamics'' & Oxygen saturation, Systolic blood pressure, Weight, pH \\        
                \bottomrule
            \end{tabular}
        }
    \end{table*}

    \begin{table*}[htbp]
        \floatconts
        {tab:grouping_hc_gs}%
        {\caption{Feature grouping learned by Soft GMM model of HiRID features for circulatory failure prediction. Interpreted as isolated specific indicators and two large mixed feature groups.}}%
        {
            \renewcommand{\arraystretch}{1.3}
            \footnotesize
            \begin{tabular}{@{}p{0.28\linewidth}p{0.68\linewidth}@{}}
                \toprule
                Group name & Features \\ \midrule
                    Cluster 0: ``mixture class'' & OUTurine/h, Administriation of antibiotics, administration of antiviral, Acetazolamide, non-steroids, Parenteral Feeding, a-BE, Ca, Ketalar, Antiepileptica, Anti delirant medi, aPTT, Fibrinogen, Muskelrelaxans, Anexate, Cl-, Factor VII, Neutr, Amiodaron, TSH, T Central, ABPs, ABPd, GCS Antwort, Glucose Administration, PAPd, PCWP, Steroids, Others in Case of HIT, Marcoumar, Propofol, Protamin, Anti Fibrinolyticum, Kalium, Non-opioide, a\_Hb, a\_Lac, Incrys, Incolloid, packed red blood cells, platelets, coagulation factors, pH Liquor, Plateaudruck, theophyllin, C-reactive protein, AWPmean, RR set, vasodilatators, ACE Inhibitors, Adenosin, Ca2+ total, Digoxin, a\_pCO2, INR, albumin, phosphate, Ammoniak, Hb, a\_PO2, Mg\_lab, platelet count, Atropin, Urea, Mineralokortikoid, Segm. Neut., Terlipressin, BNP, urinary creatinin, urinary Na+, urinary urea, Lipase, AMYL-S \\
                    Cluster 1: ``rhabdomyolysis indicator'' & creatine kinase \\
                    Cluster 2: ``biliary sepsis indicator'' & alkaline phosphatase \\
                    Cluster 3: ``acute myocardial infarction  indicator'' & creatine kinase-MB \\ 
                    Cluster 4: ``mixture class'' & NIBPs, Administation of antimycotic, RR, K-sparend, NIBPd, NIBPm, GCS Motorik, Insuling Langwirksam, GCS Augenoffnen, PAPm, PAPs, ICP, supplemental oxygen, Aldosteron Antagonist, Enteral Feeding, ZVD, Chemotherapie, Immunoglobulin, Immunsuppression, Phosphat, Na, a\_MetHb, OUT, NSAR, v-Lac, a\_HCO3-, Peripherial Anesthesia, K+, Na+, FII, Parkinson Medikaiton, dobutamine, Sartane, Factor V, Ca Antagonists, B-Blocker, Andere, Ca2+ ionizied, Beh. Pulm. Hypertonie, Stabk. Neut., Cortisol, HR, ASAT, SpO2, ABPm, ALAT, ETCO2, RASS, Antihelmenticum, TOF, bilirubine; total, CO, SvO2(m), Loop diuretics, Thiazide, Haemofiltration, steroids, FIO2, ST1, Insulin Kurzwirksam, Peep, Thrombozytenhemmer, Heparin, NMH, Benzodiacepine, Alpha 2 Agonisten, Barbiturate, ST2, Liquor/h, ST3, Lysetherapie, Rhythmus, IN, Pankreas Enzyme, VitB Substitution, Nimodipin, Opiate, Weight, Mg, a\_COHb, Trace elements, Bicarbonate, GCSF, FFP, a\_pH, Psychopharma, norepinephrine, epinephrine, Ventilator mode, TV, Naloxon, Spitzendruck, milrinone, levosimendan, vasopressin, procalcitonin, lymphocyte, desmopressin, Laktat Liquor, AiwayCode, factor X, glucose, total white blood cell count, a\_SO2, Zentral venöse sättigung, Bilirubin; direct, Thyrotardin, Thyroxin, Thyreostatikum, MCH, gamma-GT, MCHC, Antihistaminka, Troponin-T, pH Drain, Glucose Liquor, creatinine, MCV, BSR, AMYL-Drainag, Ferritin \\
                \bottomrule
            \end{tabular}
        }
    \end{table*}

    \begin{table*}[htbp]
        \floatconts
        {tab:grouping_hc_kf}%
        {\caption{Feature grouping learned by Fuzzy K-means model of HiRID features for circulatory failure prediction.}}%
        {
            \renewcommand{\arraystretch}{1.3}
            \footnotesize
            \begin{tabular}{@{}p{0.28\linewidth}p{0.68\linewidth}@{}}
                \toprule
                Group name & Features \\ \midrule
                Cluster 0: ``mixed'' & OUTurine/h, NIBPs, Administation of antimycotic, K-sparend, NIBPd, GCS Motorik, Insuling Langwirksam, GCS Augenoffnen, administration of antiviral, PAPm, PAPs, ICP, supplemental oxygen, Aldosteron Antagonist, Enteral Feeding, ZVD, non-steroids, Chemotherapie, Parenteral Feeding, Immunoglobulin, Immunsuppression, Phosphat, Na, a-BE, a\_MetHb, Ca, OUT, NSAR, v-Lac, Ketalar, a\_HCO3-, Peripherial Anesthesia, Antiepileptica, Anti delirant medi, Fibrinogen, K+, Na+, Parkinson Medikaiton, dobutamine, Sartane, Cl-, Factor V, Ca Antagonists, B-Blocker, Andere, Beh. Pulm. Hypertonie, Neutr, Stabk. Neut., None, Cortisol, None, T Central, ABPd, ASAT, ABPm, Glucose Administration, RASS, PAPd, PCWP, Antihelmenticum, TOF, bilirubine; total, Steroids, CO, SvO2(m), Loop diuretics, Thiazide, Haemofiltration, steroids, FIO2, Thrombozytenhemmer, NMH, Others in Case of HIT, Benzodiacepine, Alpha 2 Agonisten, Barbiturate, Marcoumar, ST2, Protamin, Anti Fibrinolyticum, Liquor/h, Rhythmus, Pankreas Enzyme, VitB Substitution, Kalium, Non-opioide, Mg, a\_Hb, Trace elements, Incolloid, packed red blood cells, Bicarbonate, FFP, a\_pH, norepinephrine, epinephrine, TV, Naloxon, Spitzendruck, Plateaudruck, milrinone, levosimendan, theophyllin, vasopressin, procalcitonin, desmopressin, vasodilatators, AiwayCode, Ca2+ total, albumin, glucose, total white blood cell count, a\_SO2, Mg\_lab, None, alkaline phosphatase, Atropin, Thyrotardin, Urea, Mineralokortikoid, MCH, MCHC, Segm. Neut., Terlipressin, Troponin-T, creatinine, AMYL-Drainag, urinary Na+, AMYL-S \\
                Cluster 1: ``acute neurological status'' & ETCO2, Lysetherapie, Nimodipin, Incrys, coagulation factors, pH Liquor, Laktat Liquor, a\_pCO2, BSR, BNP \\
                Cluster 2: ``sepsis and pancreatic dysfunction'' & Administriation of antibiotics, ALAT, Heparin, Propofol, ST3, Weight, platelets, lymphocyte, phosphate, Thyroxin, Thyreostatikum, Antihistaminka, Lipase \\
                Cluster 3: ``sever ARDS + FVII'' & Factor VII, Opiate, AWPmean \\
                Cluster 4: ``dynamic shock and inflammation'' & RR, NIBPm, Acetazolamide, aPTT, Muskelrelaxans, Anexate, FII, Ca2+ ionizied, Amiodaron, TSH, HR, ABPs, SpO2, GCS Antwort, ST1, Insulin Kurzwirksam, Peep, IN, a\_COHb, a\_Lac, GCSF, Psychopharma, Ventilator mode, C-reactive protein, RR set, ACE Inhibitors, Adenosin, factor X, Digoxin, INR, Ammoniak, Hb, a\_PO2, Zentral venöse sättigung, None, platelet count, Bilirubin; direct, gamma-GT, None, pH Drain, Glucose Liquor, creatine kinase, MCV, creatine kinase-MB, Ferritin, urinary creatinin, urinary urea \\
                \bottomrule
            \end{tabular}
        }
    \end{table*}

    \begin{table*}[htbp]
        \floatconts
        {tab:grouping_hm_k}
        {\caption{Feature grouping learned by K-means model of HiRID features for mortality prediction. Interpreted as mixed grouping, with clusters generally containing large subgroups of variables relating to different systems and conditions.}}%
        {        
            \renewcommand{\arraystretch}{1.3}
            \footnotesize
            \begin{tabular}{@{}p{0.28\linewidth}p{0.68\linewidth}@{}}
                \toprule
                Group name & Features \\ \midrule
                Cluster 0: ``cardiovascular, multi-organ variables'' & Ca, dobutamine, Ca Antagonists, Andere, Ca2+ ionizied, Beh. Pulm. Hypertonie, Amiodaron, ABPd, PAPd, FIO2, Others in Case of HIT, Barbiturate, Marcoumar, VitB Substitution, a\_Hb, epinephrine, Naloxon, procalcitonin, desmopressin, Laktat Liquor, INR, phosphate, total white blood cell count, Bilirubin; direct, MCH, pH Drain, MCV, AMYL-Drainag, Ferritin, Lipase \\
                Cluster 1: ``vitals, multi-organ variables'' & RR, administration of antiviral, PAPm, ICP, Enteral Feeding, non-steroids, Parenteral Feeding, Immunsuppression, Cl-, HR, SpO2, Glucose Administration, RASS, SvO2(m), Thrombozytenhemmer, Heparin, Benzodiacepine, Anti Fibrinolyticum, Kalium, Weight, Trace elements, GCSF, a\_pH, Plateaudruck, milrinone, ACE Inhibitors, Ca2+ total, a\_pCO2, alkaline phosphatase, Thyrotardin, Urea, Thyroxin, Mineralokortikoid, gamma-GT, Troponin-T, BSR \\
                Cluster 2: infection management, sedation & Administriation of antibiotics, Administation of antimycotic, NIBPd, supplemental oxygen, Chemotherapie, Phosphat, Ketalar, Anti delirant medi, aPTT, K+, B-Blocker, TSH,  ABPs, PCWP, Antihelmenticum, Loop diuretics, Thiazide, Alpha 2 Agonisten, Propofol, ST2, Protamin, a\_COHb, AWPmean, vasodilatators, glucose, Ammoniak, Zentral venöse sättigung, Antihistaminka, creatine kinase-MB, urinary urea, AMYL-S \\
                Cluster 3: ``fluids, inflammation'' & ZVD, Immunoglobulin, a-BE, a\_MetHb, OUT, a\_HCO3-, Peripherial Anesthesia, Fibrinogen, Muskelrelaxans, Anexate, Na+, FII, Parkinson Medikaiton, Factor VII, Haemofiltration, steroids, IN, Incrys, Incolloid, Bicarbonate, FFP, Psychopharma, Ventilator mode, pH Liquor, Spitzendruck, theophyllin, C-reactive protein, RR set, factor X, Digoxin, albumin, Mg\_lab,  Atropin, Thyreostatikum, MCHC, Terlipressin, Glucose Liquor, creatinine \\
                Cluster 4: ``vasoactive drugs, shock'' & OUTurine/h, NIBPs, K-sparend, NIBPm, GCS Motorik, Insuling Langwirksam, GCS Augenoffnen, PAPs, Aldosteron Antagonist, Acetazolamide, Na, NSAR, v-Lac, Antiepileptica, Sartane, Factor V, Neutr, Stabk. Neut., Cortisol, T Central, ASAT, ABPm, ALAT, ETCO2, GCS Antwort, TOF, bilirubine; total, Steroids, CO, ST1, Insulin Kurzwirksam, Peep, NMH, Liquor/h, ST3, Lysetherapie, Rhythmus, Pankreas Enzyme, Nimodipin, Opiate, Non-opioide, Mg, a\_Lac, packed red blood cells, platelets, coagulation factors, norepinephrine, TV, levosimendan, vasopressin, lymphocyte, AiwayCode, Adenosin, Hb, a\_PO2, a\_SO2, platelet count,  Segm. Neut., creatine kinase, BNP, urinary creatinin, urinary Na+ \\
                \bottomrule
            \end{tabular}
        }
    \end{table*}

    \begin{table*}[htbp]
        \floatconts
        {tab:grouping_hm_kf}
        {\caption{Feature grouping learned by Fuzzy K-means model of HiRID features for mortality prediction. Interpreted as mixed grouping, with clusters generally containing large subgroups of variables relating to different systems and conditions.}}%
        {        
            \renewcommand{\arraystretch}{1.3}
            \footnotesize
            \begin{tabular}{@{}p{0.28\linewidth}p{0.68\linewidth}@{}}
                \toprule
                Group name & Features \\ \midrule
                    Cluster 0: monitoring, treatments, mixed & OUTurine/h, NIBPs, Administation of antimycotic, NIBPm, GCS Motorik, administration of antiviral, Aldosteron Antagonist, Immunsuppression, Phosphat, a\_MetHb, Ca, OUT, Ketalar, a\_HCO3-, Anti delirant medi, aPTT, Muskelrelaxans, FII, Parkinson Medikaiton, Cl-, Ca Antagonists, Amiodaron, GCS Antwort, Glucose Administration, PCWP, bilirubine; total, Loop diuretics, NMH, Benzodiacepine, Liquor/h, Non-opioide, Trace elements, Spitzendruck, theophyllin, RR set, Laktat Liquor, AiwayCode, Ca2+ total, a\_pCO2, INR, phosphate, total white blood cell count, a\_PO2, alkaline phosphatase, Atropin, Thyreostatikum, gamma-GT, Terlipressin, Troponin-T, Glucose Liquor, MCV, BSR, AMYL-Drainag, Ferritin, AMYL-S \\
                    Cluster 1: sefation, shock, mixed & K-sparend, Insuling Langwirksam, Parenteral Feeding, v-Lac, Neutr, Stabk. Neut., TSH, HR, RASS, TOF, steroids, Others in Case of HIT, Alpha 2 Agonisten, Barbiturate, Marcoumar, Propofol, VitB Substitution, Nimodipin, Opiate, a\_Lac, Bicarbonate, Psychopharma, milrinone, Thyrotardin, Urea \\
                    Cluster 2: cardiopulmonary management during resuscitation & CO, Peep, Incrys \\
                    Cluster 3: hemodynamics, respiratory, infection, mixed & Administriation of antibiotics, RR, NIBPd, GCS Augenoffnen, PAPm, PAPs, ICP, supplemental oxygen, Acetazolamide, Enteral Feeding, ZVD, non-steroids, Chemotherapie, Immunoglobulin, Na, a-BE, NSAR, Peripherial Anesthesia, Antiepileptica, Fibrinogen, K+, Anexate, Na+, dobutamine, Sartane, Factor V, B-Blocker, Andere, Ca2+ ionizied, Beh. Pulm. Hypertonie, Factor VII, Cortisol, T Central, ABPs, ABPd, ASAT, SpO2, ABPm, ALAT, ETCO2, PAPd, Antihelmenticum, Steroids, SvO2(m), Thiazide, Haemofiltration, FIO2, ST1, Insulin Kurzwirksam, Thrombozytenhemmer, Heparin, ST2, Protamin, Anti Fibrinolyticum, ST3, Lysetherapie, Rhythmus, IN, Pankreas Enzyme, Kalium, Weight, Mg, a\_COHb, a\_Hb, Incolloid, packed red blood cells, GCSF, FFP, a\_pH, platelets, coagulation factors, norepinephrine, epinephrine, Ventilator mode, TV, Naloxon, pH Liquor, Plateaudruck, levosimendan, vasopressin, C-reactive protein, AWPmean, procalcitonin, lymphocyte, desmopressin, vasodilatators, ACE Inhibitors, Adenosin, factor X, Digoxin, albumin, glucose, Ammoniak, Hb, a\_SO2, Zentral venöse sättigung, Mg\_lab, platelet count, Bilirubin; direct, Thyroxin, Mineralokortikoid, MCH, MCHC, Antihistaminka, Segm. Neut., pH Drain, creatine kinase, creatinine, creatine kinase-MB, BNP, urinary creatinin, urinary Na+, urinary urea, Lipase \\
                \bottomrule
            \end{tabular}
        }
    \end{table*}

    \begin{figure*}[htbp]
        \floatconts
        {fig:sankey_hc_gs}
        {\caption{Dynamics of feature assignments to clusters for Soft GMM model trained on HiRID for circulatory failure prediction. Grouping does not change from the initialization.}}
        {%
            \subfigure[Cluster sizes]{%
                \includegraphics[width=0.4\linewidth]{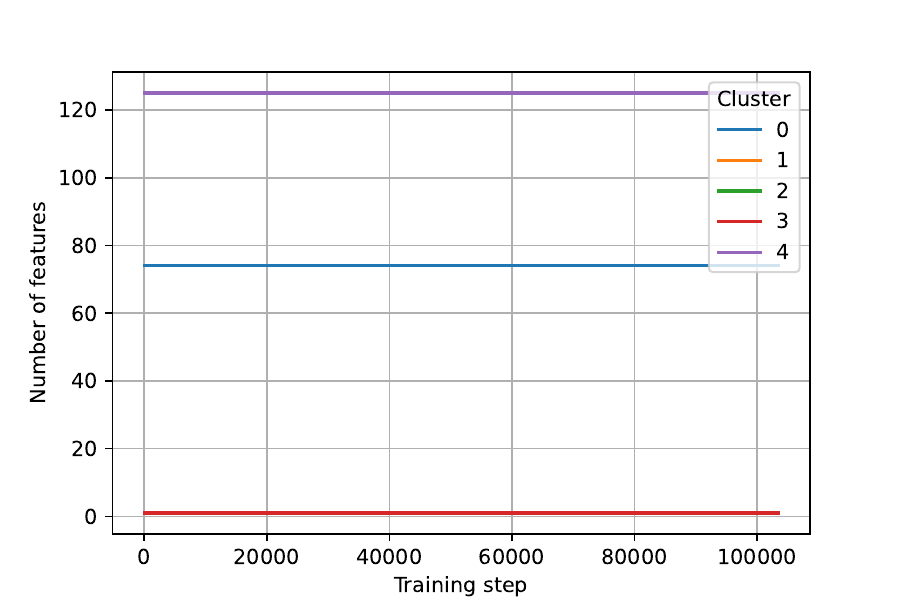}
            }
            \qquad
            \subfigure[Sankey diagram]{%
                \includegraphics[width=0.4\linewidth]{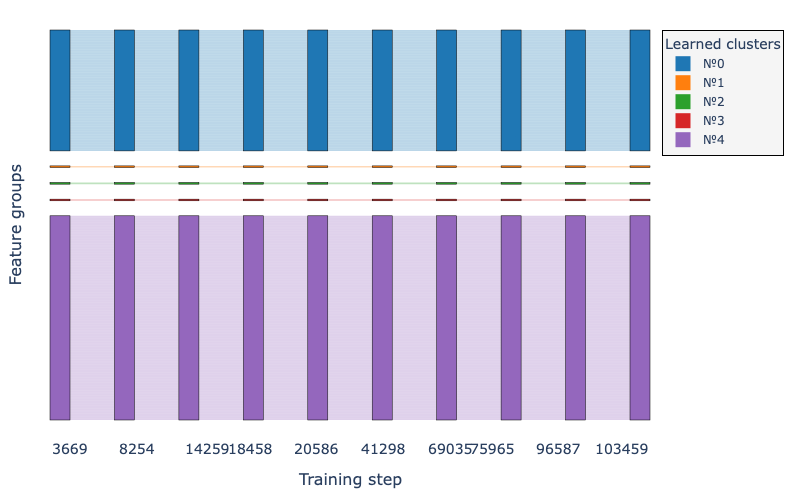}
            }
        }
        \floatconts
        {fig:sankey_hm_k}
        {\caption{Dynamics of feature assignments to clusters for K-means model trained on HiRID for mortality prediction. The groups settle in the first few iterations and stay constant throughout the rest of the training.}}
        {%
            \subfigure[Cluster sizes]{%
                \includegraphics[width=0.4\linewidth]{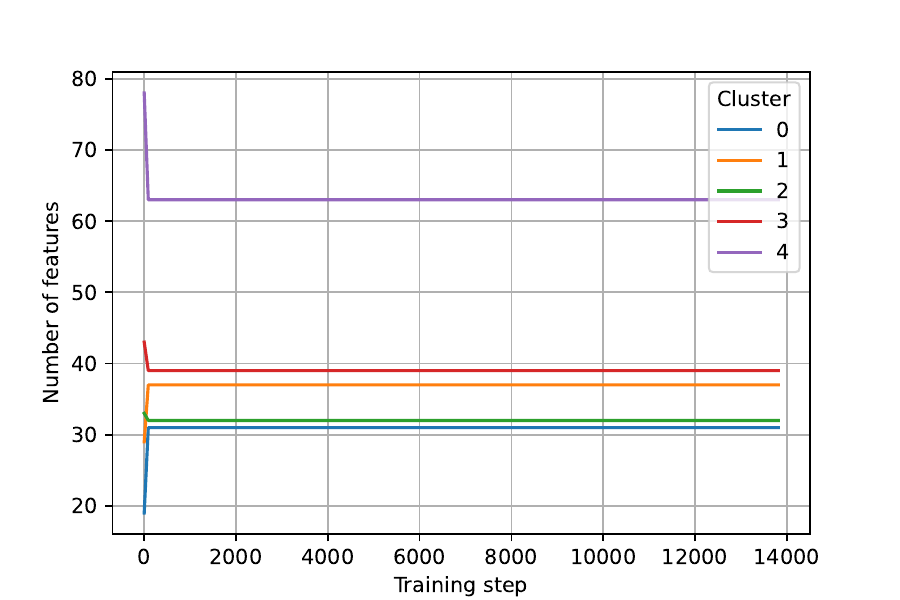}
            }
            \qquad
            \subfigure[Sankey diagram]{%
                \includegraphics[width=0.4\linewidth]{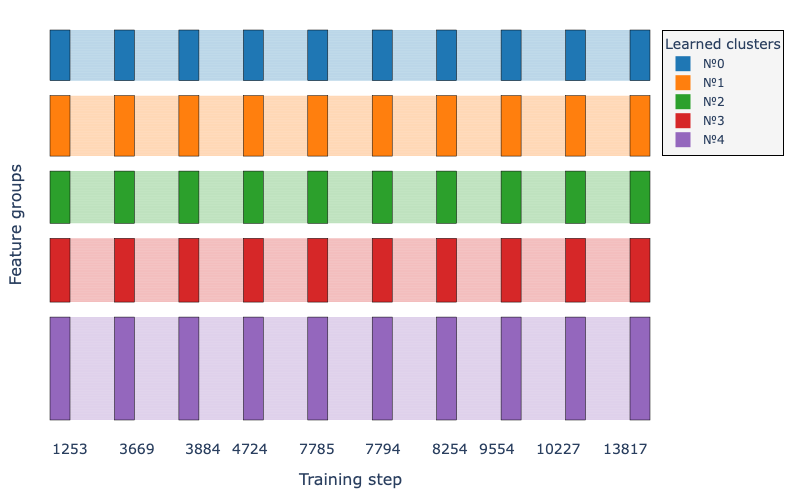}
            }
        }
        \floatconts
        {fig:sankey_hm_kf}
        {\caption{Dynamics of feature assignments to clusters for Fuzzy K-means model trained on HiRID for mortality prediction. The groups slowly evolve, with the fourth cluster gradually increasing in size.}}
        {%
            \subfigure[Cluster sizes]{%
                \includegraphics[width=0.4\linewidth]{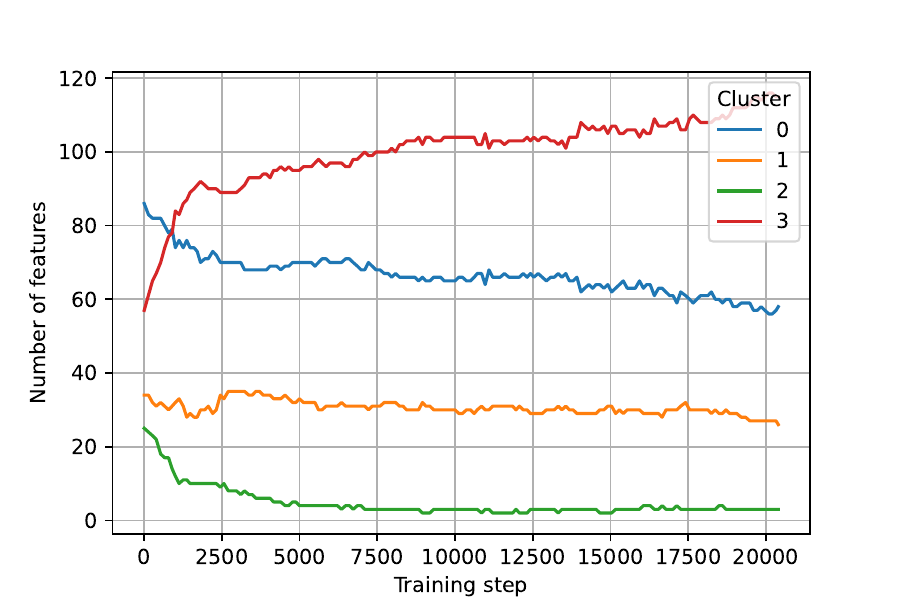}
            }
            \qquad
            \subfigure[Sankey diagram]{%
                \includegraphics[width=0.4\linewidth]{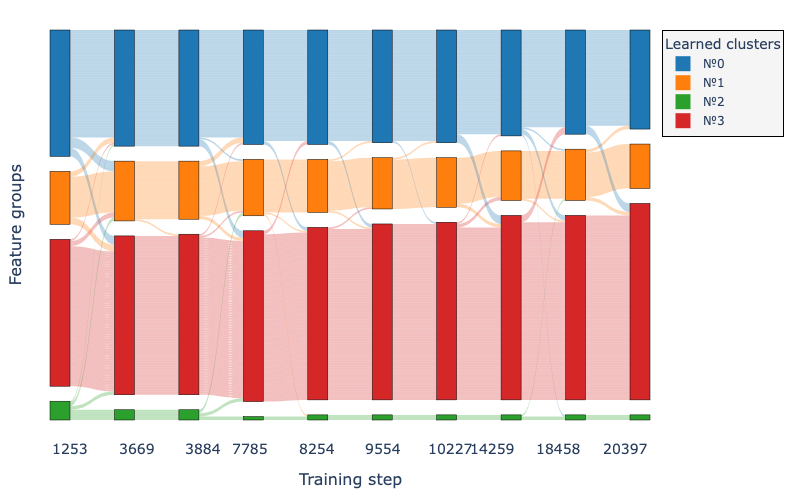}
            }
        }
    \end{figure*}

    \begin{figure*}[htbp]
        \floatconts
        {fig:sankey_md_kf}
        {\caption{Dynamics of feature assignments to clusters for Fuzzy K-means model trained on MIMIC for decompensation prediction. The features eventually end up assigned to a single cluster, replicating the best previously known grouping.}}
        {%
            \subfigure[Cluster sizes]{%
                \includegraphics[width=0.4\linewidth]{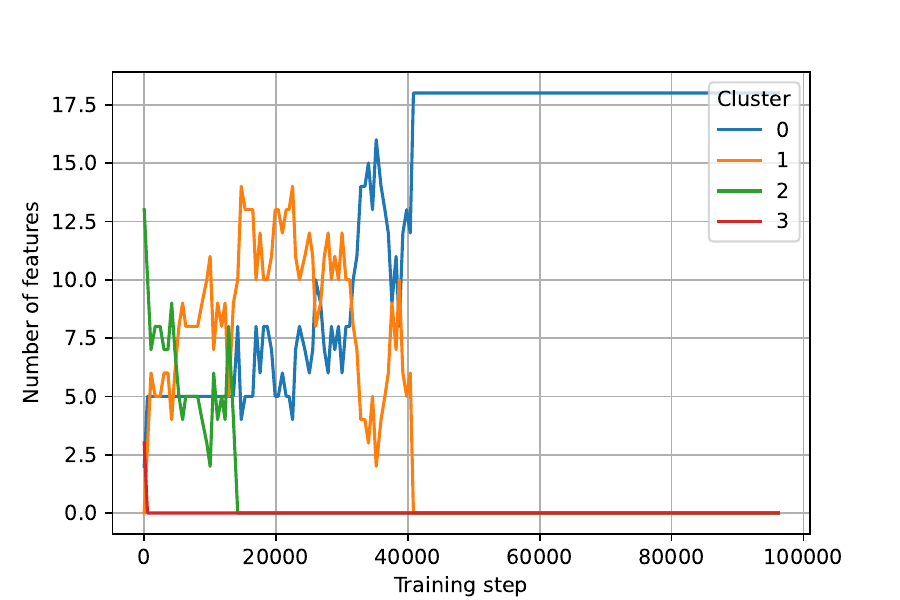}
            }
            \qquad
            \subfigure[Sankey diagram]{%
                \includegraphics[width=0.4\linewidth]{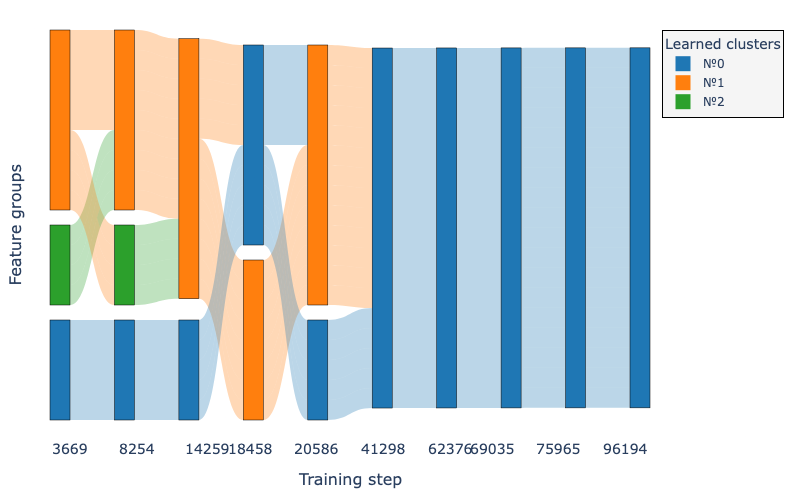}
            }
        }
        \floatconts
        {fig:sankey_mm_k}
        {\caption{Dynamics of feature assignments to clusters for K-means model trained on MIMIC for mortality prediction. Grouping does not change from the initialization.}}
        {%
            \subfigure[Cluster sizes]{%
                \includegraphics[width=0.4\linewidth]{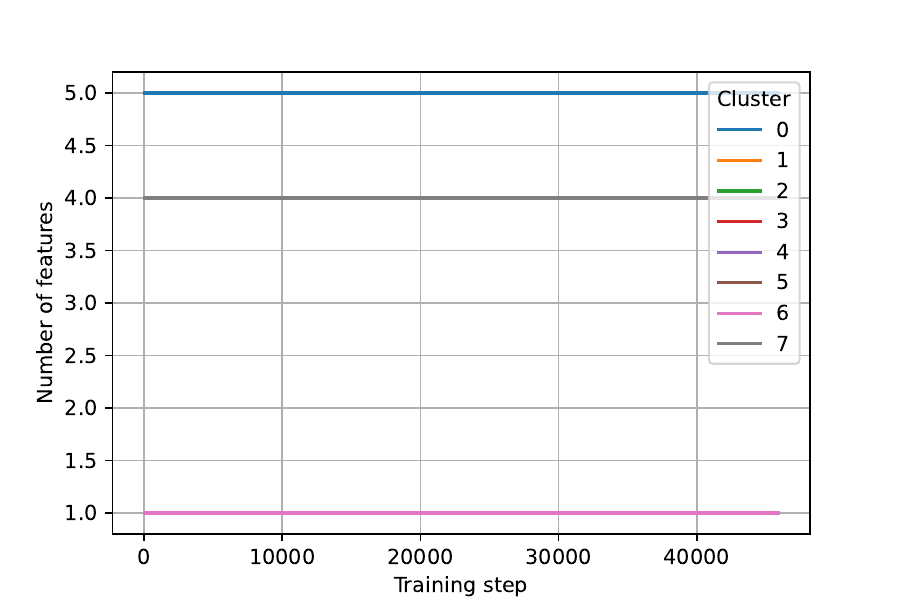}
            }
            \qquad
            \subfigure[Sankey diagram]{%
                \includegraphics[width=0.4\linewidth]{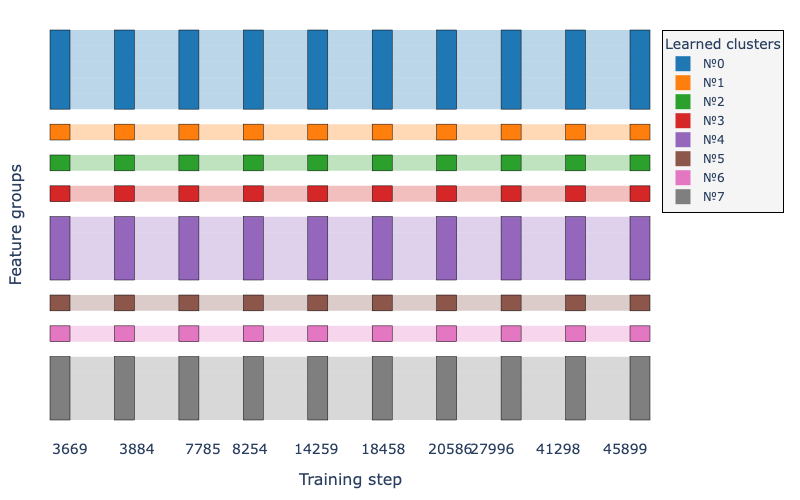}
            }
        }
        \floatconts
        {fig:sankey_mm_kf}
        {\caption{Dynamics of feature assignments to clusters for Fuzzy K-means model trained on MIMIC for mortality prediction. Grouping gradually evolves, stabilizing at around half point of the training.}}
        {%
            \subfigure[Cluster sizes]{%
                \includegraphics[width=0.4\linewidth]{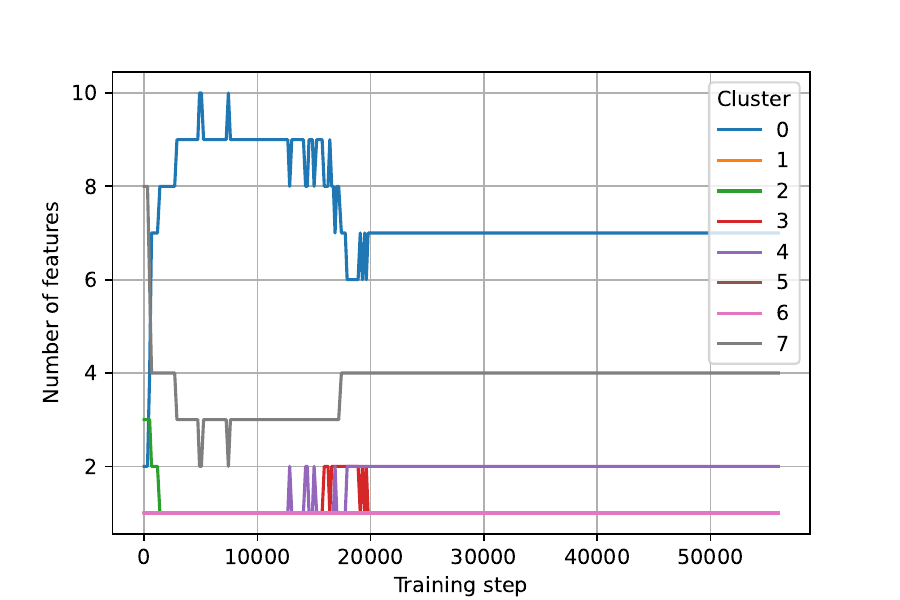}
            }
            \qquad
            \subfigure[Sankey diagram]{%
                \includegraphics[width=0.4\linewidth]{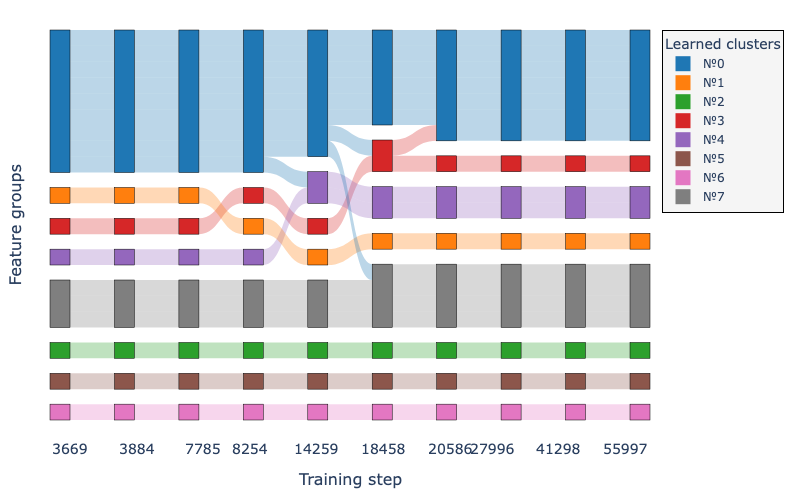}
            }
        }
    \end{figure*}

    \begin{table*}[htbp]
        \floatconts
        {tab:prior_group_organ_hirid}%
        {\caption{Prior grouping of HiRID features by organ system.}}
        {
            \small
            \renewcommand{\arraystretch}{1.3}
            \begin{tabular}{@{}p{0.2\linewidth}p{0.76\linewidth}@{}}
                \toprule
                Group name & Features \\ \midrule
                Central nervous (Cns) & GCS Antwort, GCS Motorik, GCS Augenöffnen, RASS, ICP, TOF, Benzodiacepine, Alpha 2 Agonisten, Barbiturate, Propofol, Liquor/h, Nimodipin, Opiate, Non-opioide, NSAR, Ketalar, Peripherial Anesthesia, Antiepileptica, Anti delirant medi, Psychopharma, Muskelrelaxans, Anexate, Naloxon, Parkinson Medikaiton, pH Liquor, Laktat Liquor, Glucose Liquor \\
                Circulatory & HR, T Central, ABPs, ABPd, ABPm, NIBPs, NIBPd, NIBPm, PAPm, PAPs, PAPd, PCWP, CO, SvO2(m), ZVD, ST1, ST2, ST3, Rhythmus, IN, OUT, Incrys, Incolloid, packed red blood cells, FFP, platelets, coagulation factors, norepinephrine, epinephrine, dobutamine, milrinone, levosimendan, theophyllin, vasopressin, desmopressin, vasodilatators, ACE Inhibitors, Sartane, Ca Antagonists, B-Blocker, Andere, Adenosin, Digoxin, Amiodaron, Atropin, Thyrotardin, Thyroxin, Thyreostatikum, Mineralokortikoid, Antihistaminka, Terlipressin, Troponin-T, creatine kinase, creatine kinase-MB, BNP, TSH, AMYL-S \\
                Hematologic & Glucose Administration, Insuling Langwirksam, Insulin Kurzwirksam, Thrombozytenhemmer, Heparin, NMH, Others in Case of HIT, Marcoumar, Protamin, Anti Fibrinolyticum, Lysetherapie, Pankreas Enzyme, VitB Substitution, Weight, a-BE, a COHb, a Hb, a Lac, a MetHb, v-Lac, aPTT, Fibrinogen, FII, Factor V, Factor VII, factor X, INR, albumin, glucose, Ammoniak, Hb, total white blood cell count, platelet count, MCH, MCHC, MCV, Ferritin, Lipase \\
                Immune & Administriation of antibiotics, Administation of antimycotic, administration of antiviral, Antihelmenticum, Steroids, Enteral Feeding, steroids, non-steroids, Chemotherapie, Immunoglobulin, Immunsuppression, GCSF, C-reactive protein, procalcitonin, lymphocyte, Neutr, Segm. Neut., Stabk. Neut., BSR, Cortisol \\
                Hepatic & ASAT, ALAT, bilirubine, total, Bilirubin, direct, alkaline phosphatase, gamma-GT \\
                Pulmonary & SpO2, ETCO2, RR, supplemental oxygen, FIO2, Peep, Ventilator mode, TV, Spitzendruck, Plateaudruck, AWPmean, RR set, AiwayCode, Beh. Pulm. Hypertonie, a pCO2, a PO2, a SO2, Zentral venöse sättigung, pH Drain, AMYL-Drainag \\
                Renal & OUTurine/h, K-sparend, Aldosteron Antagonist, Loop diuretics, Thiazide, Acetazolamide, Haemofiltration, Parenteral Feeding, Kalium, Phosphat, Na, Mg, Ca, Trace elements, Bicarbonate, a HCO3-, a pH, K+, Na+, Cl-, Ca2+ ionizied, Ca2+ total, phosphate, Mg lab, Urea, creatinine, urinary creatinin, urinary Na+, urinary urea \\
                \bottomrule
            \end{tabular}
        }
    \end{table*}
    
    \begin{table*}[htbp]
        \floatconts
        {tab:prior_group_organ_mimic}
        {\caption{Prior grouping of MIMIC-III features by organ system.}}       
        {
            \small
            \renewcommand{\arraystretch}{1.3}
            \begin{tabular}{@{}p{0.2\linewidth}p{0.76\linewidth}@{}}
                \toprule
                Group name & Features \\ \midrule
                Central nervous (Cns) & Glasgow coma scale eye opening, Glasgow coma scale motor response, Glasgow coma scale total, Glasgow coma scale verbal response \\
                Circulatory & Diastolic blood pressure, Heart Rate, Mean blood pressure, Systolic blood pressure, Temperature, Capillary refill rate \\
                Hematologic & Glucose \\
                Pulmonary & Fraction inspired oxygen, Oxygen saturation, Respiratory rate \\
                Renal & pH \\ 
                Other & Time, Height, Weight \\
                \bottomrule
            \end{tabular}
        }
    \end{table*}

    \begin{table*}[htbp]
        \floatconts
        {tab:prior_group_type_hirid}
        {\caption{Prior grouping of HiRID features by measurement type.}}
        {
            \small
            \renewcommand{\arraystretch}{1.3}
            \begin{tabular}{@{}p{0.2\linewidth}p{0.76\linewidth}@{}}
                \toprule
                Group name & Features \\ \midrule
                Derived from raw data & ETCO2, OUTurine/h, IN, OUT, Incrys, Incolloid \\
                Laboratory & a-BE, a COHb, a Hb, a HCO3-, a Lac, a MetHb, a pH, a pCO2, a PO2, a SO2, Zentral venöse sättigung, Troponin-T, creatine kinase, creatine kinaseMB, v-Lac, BNP, K+, Na+, Cl-, Ca2+ ionizied, Ca2+ total, phosphate, Mg lab, Urea, creatinine, urinary creatinin, urinary Na+, urinary urea, ASAT, ALAT, bilirubine, total, Bilirubin, direct, alkaline phosphatase, gamma-GT, aPTT, Fibrinogen, FII, Factor V, Factor VII, factor X, INR, albumin, glucose, Ammoniak, C-reactive protein, procalcitonin, lymphocyte, Neutr, Segm. Neut., Stabk. Neut., BSR, Hb, total white blood cell count, platelet count, MCH, MCHC, MCV, Ferritin, TSH, AMYL-S, Lipase, Cortisol, pH Liquor, Laktat Liquor, Glucose Liquor, pH Drain, AMYL-Drainag \\
                Monitored & HR, T Central, ABPs, ABPd, ABPm, NIBPs, NIBPd, NIBPm, PAPm, PAPs, PAPd, PCWP, CO, SvO2(m), ZVD, ST1, ST2, ST3, SpO2, ETCO2, RR, ICP, TOF, FIO2, Peep, Ventilator mode, TV, Spitzendruck, Plateaudruck, AWPmean, RR set \\
                Observed & ZVD, Rhythmus, supplemental oxygen, GCS Antwort, GCS Motorik, GCS Augen öffnen, RASS, ICP, AiwayCode, Haemofiltration, Liquor/h, Weight \\
                Treatment & packed red blood cells, FFP, platelets, coagulation factors, norepinephrine, epinephrine, dobutamine, milrinone, levosimendan, theophyllin, vasopressin, desmopressin, vasodilatators, ACE Inhibitors, Sartane, Ca Antagonists, BBlocker, Andere, Adenosin, Digoxin, Amiodaron, Atropin, K-sparend, Aldosteron Antagonist, Loop diuretics, Thiazide, Acetazolamide, Administriation of antibiotics, Administation of antimycotic, administration of antiviral, Antihelmenticum, Benzodiacepine, Alpha 2 Agonisten, Barbiturate, Propofol, Glucose Administration, Insuling Langwirksam, Insulin Kurzwirksam, Nimodipin, Opiate, Non-opioide, NSAR, Ketalar, Peripherial Anesthesia, Steroids, Thrombozytenhemmer, Enteral Feeding, Parenteral Feeding, Heparin, NMH, Others in Case of HIT, Marcoumar, Protamin, Anti Fibrinolyticum, Kalium, Phosphat, Na, Mg, Ca, Trace elements, Bicarbonate, Antiepileptica, Anti delirant medi, Psychopharma, steroids, non-steroids, Thyrotardin, Thyroxin, Thyreostatikum, Mineralokortikoid, Antihistaminka, Chemotherapie, Lysetherapie, Muskelrelaxans, Anexate, Naloxon, Beh. Pulm. Hypertonie, Pankreas Enzyme, Terlipressin, Immunoglobulin, Immunsuppression, VitB Substitution, Parkinson Medikaiton, GCSF \\ 
                \bottomrule
            \end{tabular}
        }
    \end{table*}

    \begin{table*}[htbp]
        \floatconts
        {tab:prior_group_type_mimic}
        {\caption{Prior grouping of MIMIC-III features by measurement type.}}
        {
            \small
            \renewcommand{\arraystretch}{1.3}
            \begin{tabular}{@{}p{0.2\linewidth}p{0.76\linewidth}@{}}
                \toprule
                Group name & Features \\ \midrule
                Laboratory & Glucose, pH \\
                Monitored & Diastolic blood pressure, Heart Rate, Mean blood pressure, Systolic blood pressure, Temperature, Fraction inspired oxygen, Oxygen saturation, Respiratory rate \\
                Observed & Glasgow coma scale eye opening, Glasgow coma scale motor response, Glasgow coma scale total, Glasgow coma scale verbal response, Capillary refill rate \\ 
                Other & Time, Height, Weight \\
                \bottomrule
            \end{tabular}
        }
    \end{table*}

    \begin{table*}[htbp]
        \centering
        \caption{Ablation results over various parameters for GMM (soft) trained on HiRID for circulatory failure prediction.}
        \setlength{\parskip}{8pt}
        \vspace{5pt}
        \label{tab:ablation_hirid_circ_gmm-soft}
        \subtable[Initialization][t]{
            \label{tab:ablation_hirid_circ_gmm-soft_initialization}
            \begin{tabular}{cc}
                \toprule
                Parameter & AUPRC \\
                \midrule
                k-means\texttt{++} & 0.403  \textpm \ 0.173 \\
                prior    & 0.366  \textpm \ 0.017 \\
                \bottomrule
            \end{tabular}
        }\hfill
        \subtable[Unification][t]{
            \label{tab:ablation_hirid_circ_gmm-soft_unification}
            \begin{tabular}{cc}
                \toprule
                Parameter & AUPRC \\
                \midrule
                bias & 0.395  \textpm \ 0.173 \\
                bias\_avg\_linear & 0.403  \textpm \ 0.005 \\
                bias\_ext\_catzero & 0.366  \textpm \ 0.017 \\
                bias\_sum\_linear & 0.396  \textpm \ 0.008 \\
                \bottomrule
            \end{tabular}
        }\hfill
        \subtable[Merge][t]{
            \label{tab:ablation_hirid_circ_gmm-soft_merge}
            \begin{tabular}{cc}
                \toprule
                Parameter & AUPRC \\
                \midrule
                attention & 0.366  \textpm \ 0.017 \\
                mean & 0.403  \textpm \ 0.173 \\
                \bottomrule
            \end{tabular}
        }
        
        \subtable[Regularization type][t]{
            \label{tab:ablation_hirid_circ_gmm-soft_regularization-type}
            \begin{tabular}{cc}
                \toprule
                Parameter & AUPRC \\
                \midrule
                hard & 0.393  \textpm \ 0.173 \\
                soft & 0.403  \textpm \ 0.008 \\
                \bottomrule
            \end{tabular}
        }\hfill
        \subtable[Regularization coefficient][t]{
            \label{tab:ablation_hirid_circ_gmm-soft_regularization-coefficient}
            \begin{tabular}{cc}
                \toprule
                Parameter & AUPRC \\
                \midrule
                0.0000 & 0.393  \textpm \ 0.008 \\
                0.0001 & 0.403  \textpm \ 0.007 \\
                0.0010 & 0.395  \textpm \ 0.173 \\
                \bottomrule
            \end{tabular}
        }\qquad
        \subtable[Regularization coefficient][t]{
            \includegraphics[width=0.3\linewidth]{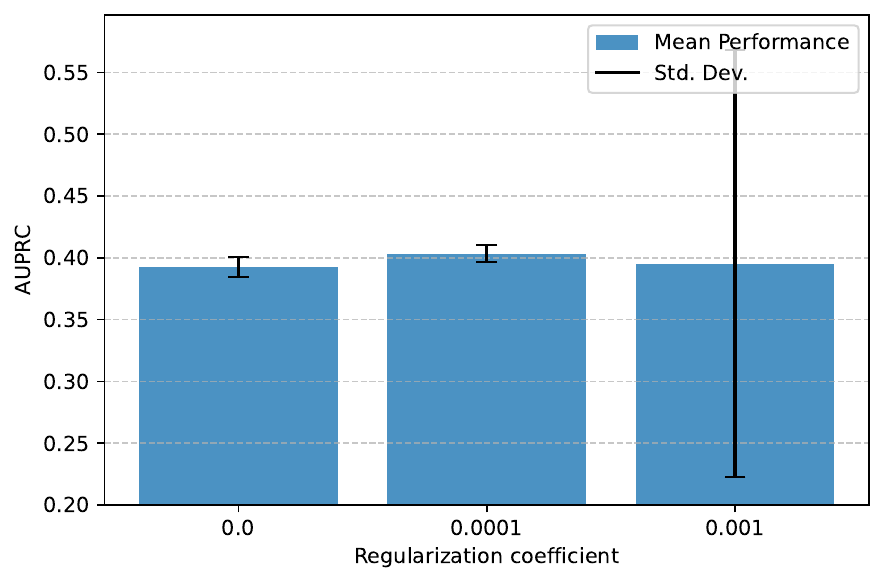}
        }
        
        \subtable[EMA rate][t]{
            \label{tab:ablation_hirid_circ_gmm-soft_ema-rate}
            \begin{tabular}{cc}
                \toprule
                Parameter & AUPRC \\
                \midrule
                0.00 & 0.393  \textpm \ 0.017 \\
                0.05 & 0.403  \textpm \ 0.004 \\
                0.25 & 0.396  \textpm \ 0.008 \\
                0.50 & 0.293  \textpm \ 0.173 \\
                0.75 & 0.393  \textpm \ 0.007 \\
                0.95 & 0.389  \textpm \ 0.008 \\
                \bottomrule
            \end{tabular}
        }\qquad
        \subtable[EMA rate][t]{
            \includegraphics[width=0.3\linewidth]{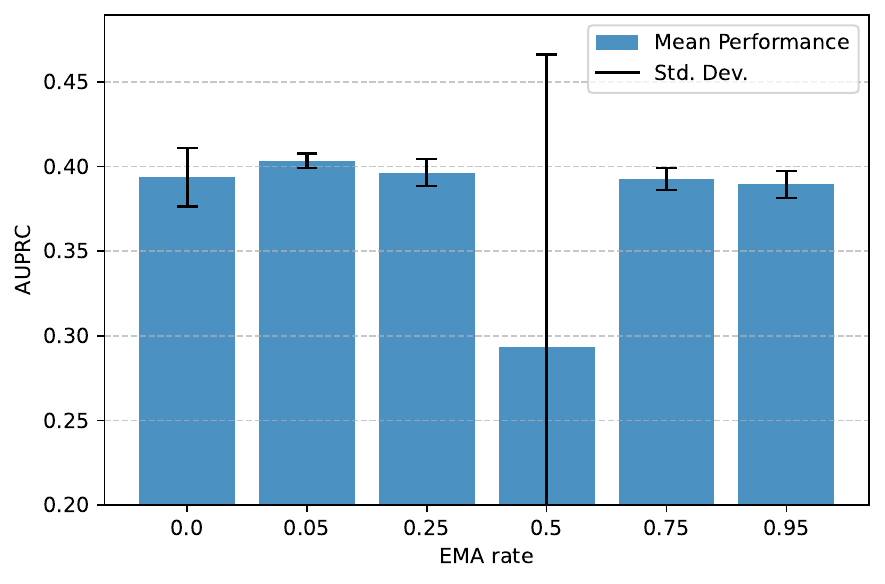}
        }
        
        \subtable[Number of clusters][t]{
            \label{tab:ablation_hirid_circ_gmm-soft_number-of-clusters}
            \begin{tabular}{cc}
                \toprule
                Parameter & AUPRC \\
                \midrule
                3 & 0.393  \textpm \ 0.007 \\
                5 & 0.403  \textpm \ 0.007 \\
                7 & 0.396  \textpm \ 0.003 \\
                9 & 0.395  \textpm \ 0.017 \\
                11 & 0.389  \textpm \ 0.173 \\
                \bottomrule
            \end{tabular}
        }\qquad
        \subtable[Number of clusters][t]{
            \includegraphics[width=0.3\linewidth]{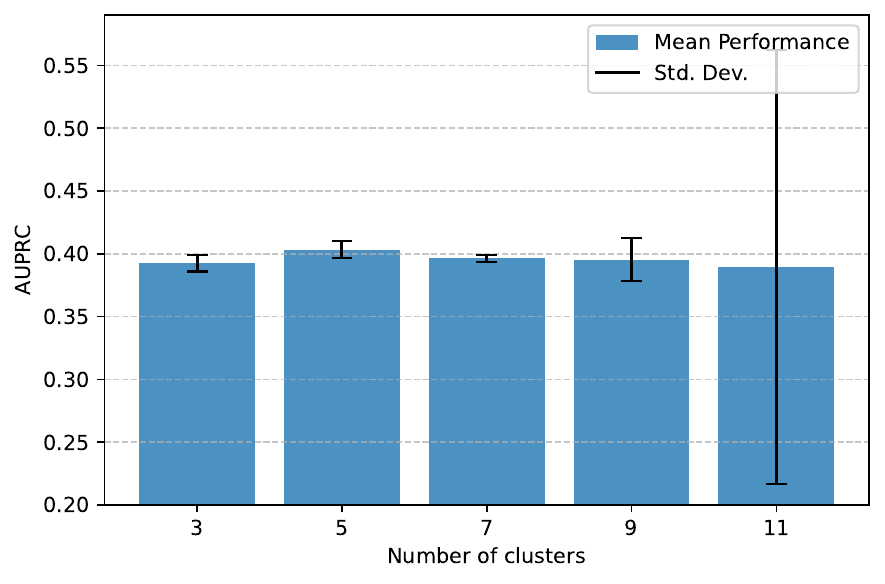}
        }
        
        \subtable[Membership threshold][t]{
            \label{tab:ablation_hirid_circ_gmm-soft_membership-threshold}
            \begin{tabular}{cc}
                \toprule
                Parameter & AUPRC \\
                \midrule
                0.80 & 0.393  \textpm \ 0.008 \\
                0.85 & 0.395  \textpm \ 0.173 \\
                0.90 & 0.403  \textpm \ 0.017 \\
                \bottomrule
            \end{tabular}
        }\qquad
        \subtable[Membership threshold][t]{
            \includegraphics[width=0.3\linewidth]{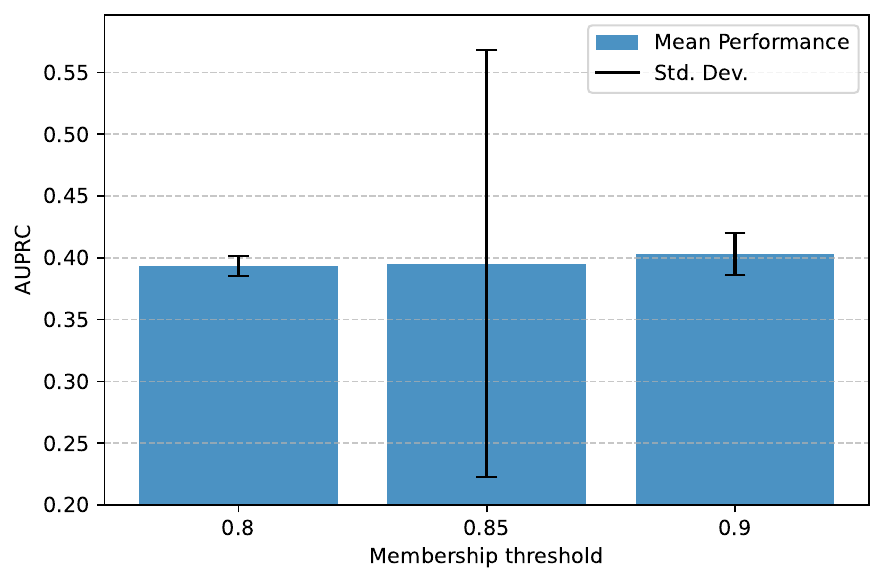}
        }
    \end{table*}
    
    \begin{table*}[htbp]
        \centering
        \caption{Ablation results over various parameters for K-means trained on HiRID for mortality prediction.}
        \setlength{\parskip}{10pt}
        \vspace{5pt}
        \label{tab:ablation_hirid_mort_kmeans}
        \subtable[Initialization][t]{
            \label{tab:ablation_hirid_mort_kmeans_initialization}
            \begin{tabular}{cc}
                \toprule
                Parameter & AUPRC \\
                \midrule
                k-means\texttt{++} & 0.631  \textpm \ 0.080 \\
                prior & 0.615  \textpm \ 0.026 \\
                \bottomrule
            \end{tabular}
        }\hfill
        \subtable[Unification][t]{
            \label{tab:ablation_hirid_mort_kmeans_unification}
            \begin{tabular}{cc}
                \toprule
                Parameter & AUPRC \\
                \midrule
                bias & 0.631  \textpm \ 0.080 \\
                bias\_avg\_linear & 0.615  \textpm \ 0.009 \\
                bias\_ext\_catzero & 0.584  \textpm \ 0.078 \\
                bias\_sum\_linear & 0.602  \textpm \ 0.014 \\
                \bottomrule
            \end{tabular}
        }\hfill
        \subtable[Merge][t]{
            \label{tab:ablation_hirid_mort_kmeans_merge}
            \begin{tabular}{cc}
                \toprule
                Parameter & AUPRC \\
                \midrule
                attention & 0.631  \textpm \ 0.080 \\
                mean & 0.615  \textpm \ 0.078 \\
                \bottomrule
            \end{tabular}
        }
        
        \subtable[Regularization type][t]{
            \label{tab:ablation_hirid_mort_kmeans_regularization-type}
            \begin{tabular}{cc}
                \toprule
                Parameter & AUPRC \\
                \midrule
                hard & 0.631  \textpm \ 0.080 \\
                soft & 0.602  \textpm \ 0.078 \\
                \bottomrule
            \end{tabular}
        }\hfill
        \subtable[Regularization coefficient][t]{
            \label{tab:ablation_hirid_mort_kmeans_regularization-coefficient}
            \begin{tabular}{cc}
                \toprule
                Parameter & AUPRC \\
                \midrule
                0.0000 & 0.626  \textpm \ 0.080 \\
                0.0001 & 0.615  \textpm \ 0.078 \\
                0.0010 & 0.631  \textpm \ 0.009 \\
                0.1000 & 0.431  \textpm \ 0.017 \\
                \bottomrule
            \end{tabular}
        }\hfill        
        \subtable[Regularization coefficient][t]{
            \includegraphics[width=0.3\linewidth]{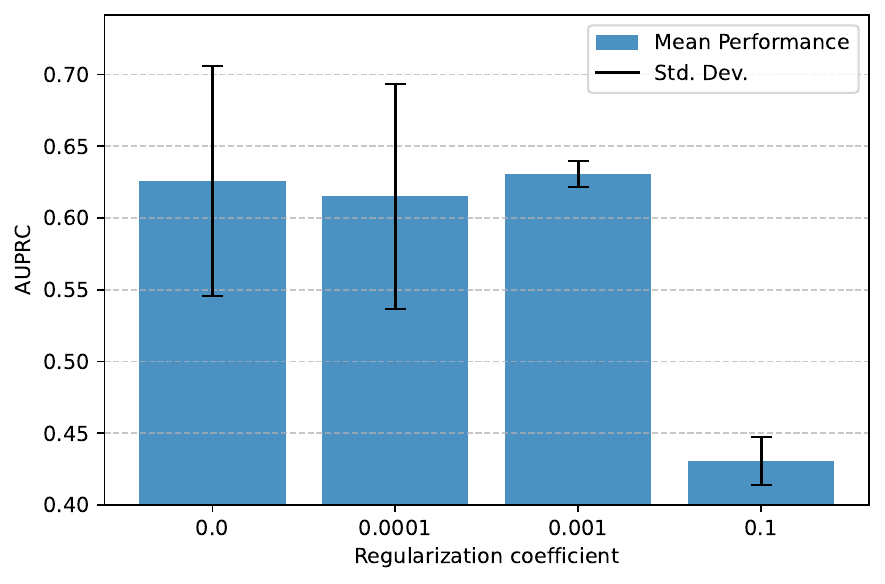}
        }

        \subtable[EMA rate][t]{
            \label{tab:ablation_hirid_mort_kmeans_ema-rate}
            \begin{tabular}{cc}
                \toprule
                Parameter & AUPRC \\
                \midrule
                0.00 & 0.604  \textpm \ 0.012 \\
                0.05 & 0.626  \textpm \ 0.015 \\
                0.25 & 0.607  \textpm \ 0.028 \\
                0.50 & 0.602  \textpm \ 0.078 \\
                0.75 & 0.564  \textpm \ 0.080 \\
                0.95 & 0.631  \textpm \ 0.026 \\
                \bottomrule
            \end{tabular}
        }\qquad        
        \subtable[EMA rate][t]{
            \includegraphics[width=0.3\linewidth]{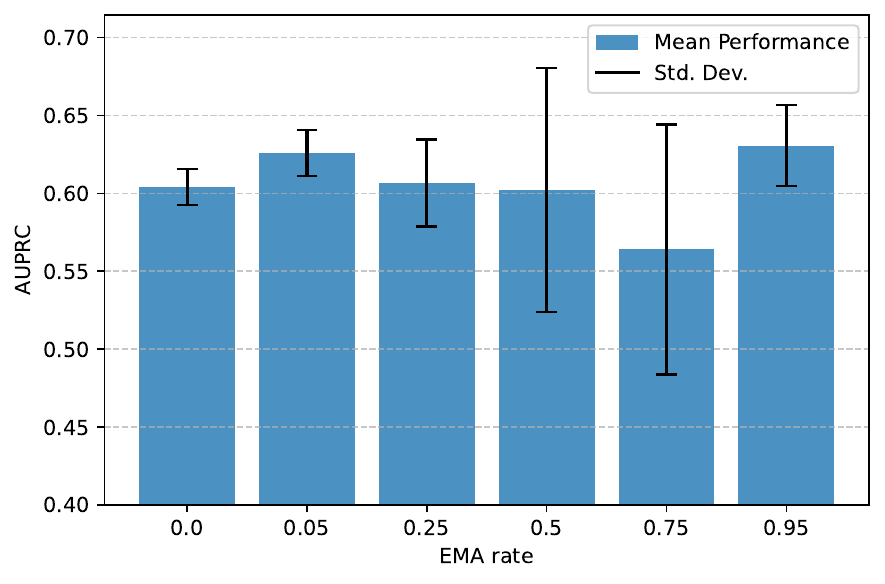}
        }
        
        \subtable[Number of clusters][t]{
            \label{tab:ablation_hirid_mort_kmeans_number-of-clusters}
            \begin{tabular}{cc}
                \toprule
                Parameter & AUPRC \\
                \midrule
                3 & 0.626  \textpm \ 0.080 \\
                5 & 0.631  \textpm \ 0.009 \\
                8 & 0.534  \textpm \ 0.078 \\
                \bottomrule
            \end{tabular}
        }\qquad
        \subtable[Number of clusters][t]{
            \includegraphics[width=0.3\linewidth]{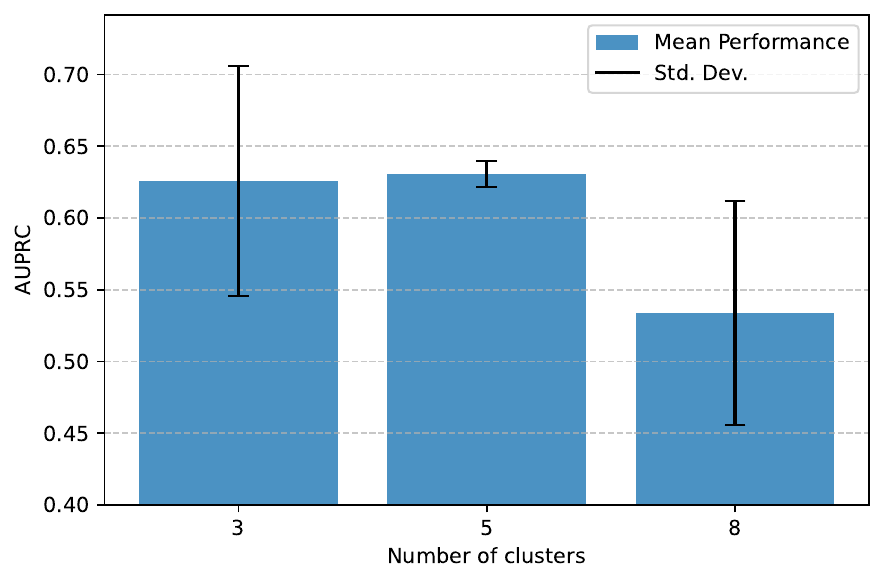}
        }
    \end{table*}

    \begin{table*}[htbp]
        \centering
        \caption{Ablation results over various parameters for Fuzzy K-means trained on HiRID for mortality prediction.}
        \setlength{\parskip}{8pt}
        \vspace{5pt}
        \label{tab:ablation_hirid_mort_kmeans-fuzzy}
        \subtable[Initialization][t]{
            \label{tab:ablation_hirid_mort_kmeans-fuzzy_initialization}
            \begin{tabular}{cc}
                \toprule
                Parameter & AUPRC \\
                \midrule
                k-means\texttt{++} & 0.597  \textpm \ 0.040 \\
                prior    & 0.629  \textpm \ 0.059 \\
                \bottomrule
            \end{tabular}
        }\hfill
        \subtable[Unification][t]{
            \label{tab:ablation_hirid_mort_kmeans-fuzzy_unification}
            \begin{tabular}{cc}
                \toprule
                Parameter & AUPRC \\
                \midrule
                bias & 0.608  \textpm \ 0.012 \\
                bias\_avg\_linear & 0.610  \textpm \ 0.040 \\
                bias\_ext\_catzero & 0.620  \textpm \ 0.059 \\
                bias\_sum\_linear & 0.629  \textpm \ 0.016 \\
                \bottomrule
            \end{tabular}
        }\hfill
        \subtable[Merge][t]{
            \label{tab:ablation_hirid_mort_kmeans-fuzzy_merge}
            \begin{tabular}{cc}
                \toprule
                Parameter & AUPRC \\
                \midrule
                attention & 0.620  \textpm \ 0.059 \\
                mean & 0.629  \textpm \ 0.040 \\
                \bottomrule
            \end{tabular}
        }
        
        \subtable[Regularization type][t]{
            \label{tab:ablation_hirid_mort_kmeans-fuzzy_regularization-type}
            \begin{tabular}{cc}
                \toprule
                Parameter & AUPRC \\
                \midrule
                hard & 0.629  \textpm \ 0.040 \\
                soft & 0.610  \textpm \ 0.059 \\
                \bottomrule
            \end{tabular}
        }\hfill
        \subtable[Regularization coefficient][t]{
            \label{tab:ablation_hirid_mort_kmeans-fuzzy_regularization-coefficient}
            \begin{tabular}{cc}
                \toprule
                Parameter & AUPRC \\
                \midrule
                0.0000 & 0.629  \textpm \ 0.016 \\
                0.0001 & 0.468  \textpm \ 0.059 \\
                0.0010 & 0.595  \textpm \ 0.040 \\
                \bottomrule
            \end{tabular}
        }\qquad
        \subtable[Regularization coefficient][t]{
            \includegraphics[width=0.3\linewidth]{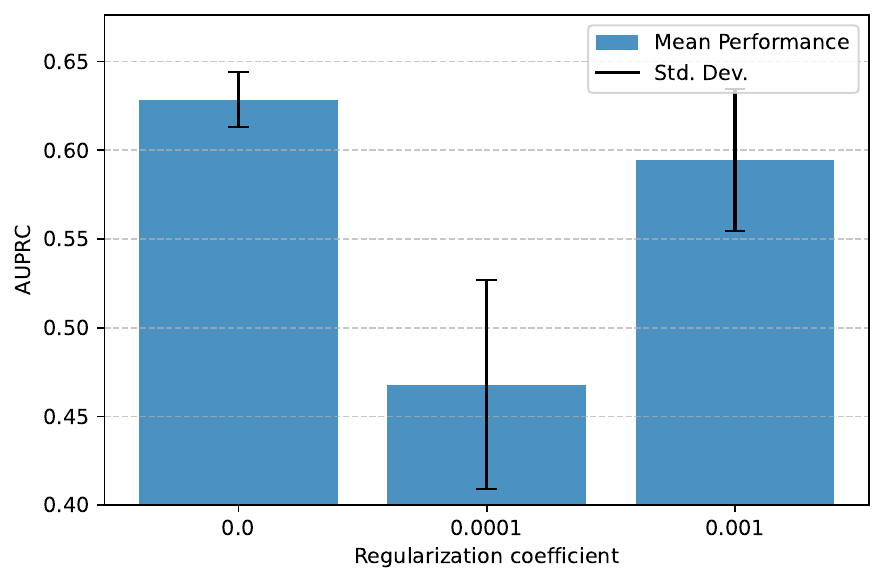}
        }
        
        \subtable[EMA rate][t]{
            \label{tab:ablation_hirid_mort_kmeans-fuzzy_ema-rate}
            \begin{tabular}{cc}
                \toprule
                Parameter & AUPRC \\
                \midrule
                0.25 & 0.629  \textpm \ 0.059 \\
                0.50 & 0.610  \textpm \ 0.040 \\
                0.75 & 0.620  \textpm \ 0.015 \\
                0.95 & 0.597  \textpm \ 0.016 \\
                \bottomrule
            \end{tabular}
        }\qquad
        \subtable[EMA rate][t]{
            \includegraphics[width=0.3\linewidth]{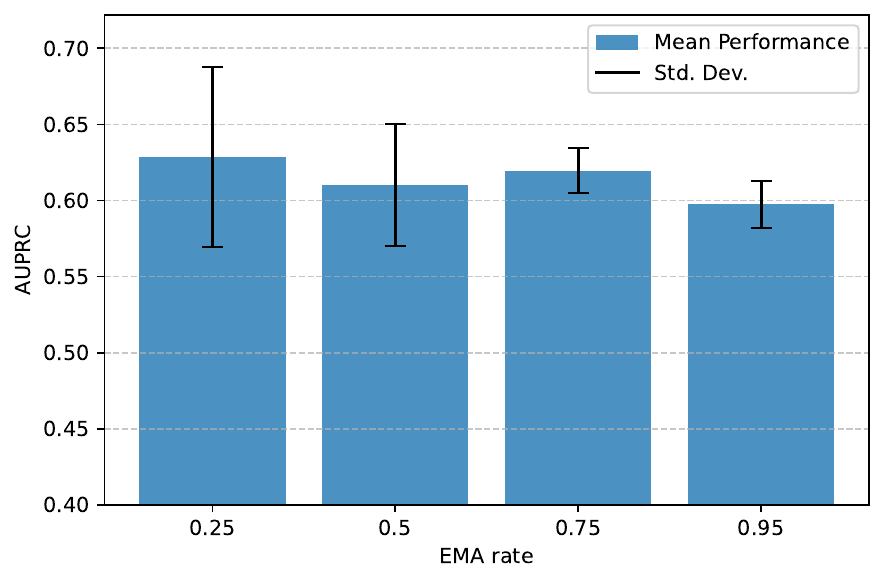}
        }
        
        \subtable[Number of clusters][t]{
            \label{tab:ablation_hirid_mort_kmeans-fuzzy_number-of-clusters}
            \begin{tabular}{cc}
                \toprule
                Parameter & AUPRC \\
                \midrule
                3 & 0.610  \textpm \ 0.012 \\
                4 & 0.608  \textpm \ 0.008 \\
                5 & 0.374  \textpm \ 0.059 \\
                6 & 0.620  \textpm \ 0.015 \\
                8 & 0.629  \textpm \ 0.040 \\
                \bottomrule
            \end{tabular}
        }\qquad
        \subtable[Number of clusters][t]{
            \includegraphics[width=0.3\linewidth]{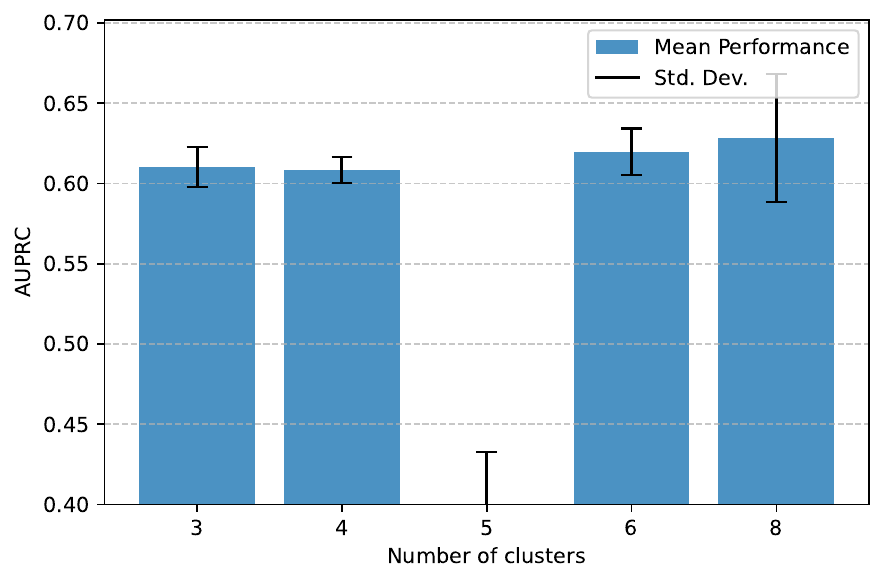}
        }
        
        \subtable[Fuzzy coefficient][t]{
            \label{tab:ablation_hirid_mort_kmeans-fuzzy_fuzzy-coefficient}
            \begin{tabular}{cc}
                \toprule
                Parameter & AUPRC \\
                \midrule
                2 & 0.610  \textpm \ 0.012 \\
                5 & 0.629  \textpm \ 0.059 \\
                10 & 0.597  \textpm \ 0.016 \\
                \bottomrule
            \end{tabular}
        }\qquad
        \subtable[Fuzzy coefficient][t]{
            \includegraphics[width=0.3\linewidth]{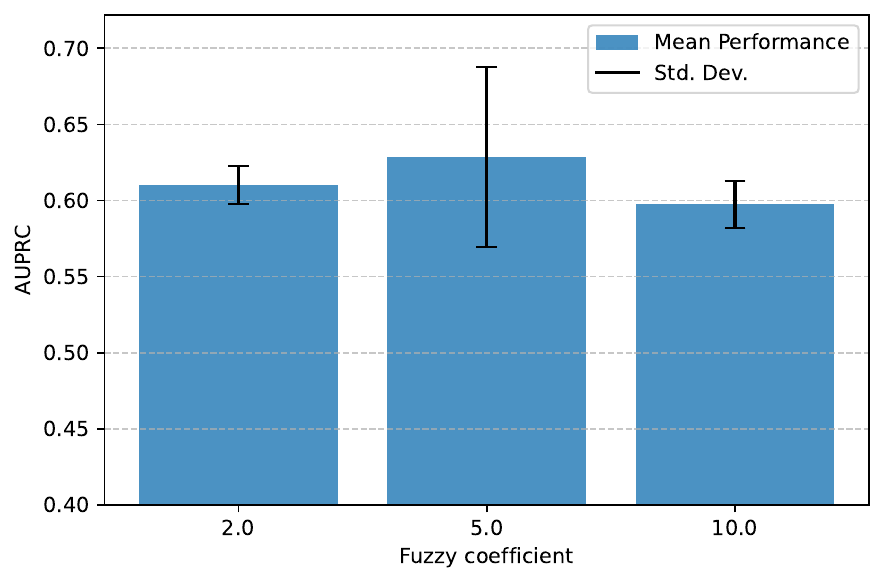}
        }
    \end{table*}

    \begin{table*}[htbp]
        \centering
        \caption{Ablation results over various parameters for GMM trained on MIMIC for decompensation prediction.}
        \setlength{\parskip}{10pt}
        \vspace{5pt}
        \label{tab:ablation_mimic_decomp_gmm}
        \subtable[Initialization][t]{
            \label{tab:ablation_hirid_mort_mort_gmm_initialization}
            \begin{tabular}{cc}
                \toprule
                Parameter & AUPRC \\
                \midrule
                k-means\texttt{++} & 0.378  \textpm \ 0.025 \\
                prior    & 0.391  \textpm \ 0.006 \\
                \bottomrule
            \end{tabular}
        }\hfill
        \subtable[Unification][t]{
            \label{tab:ablation_hirid_mort_mort_gmm_unification}
            \begin{tabular}{cc}
                \toprule
                Parameter & AUPRC \\
                \midrule
                bias & 0.385  \textpm \ 0.005 \\
                bias\_avg\_linear & 0.391  \textpm \ 0.012 \\
                bias\_ext\_catzero & 0.364  \textpm \ 0.025 \\
                bias\_sum\_linear & 0.377  \textpm \ 0.011 \\
                \bottomrule
            \end{tabular}
        }\hfill
        \subtable[Merge][t]{
            \label{tab:ablation_hirid_mort_mort_gmm_merge}
            \begin{tabular}{cc}
                \toprule
                Parameter & AUPRC \\
                \midrule
                attention & 0.391  \textpm \ 0.025 \\
                mean & 0.381  \textpm \ 0.012 \\
                \bottomrule
            \end{tabular}
        }
        
        \subtable[Regularization type][t]{
            \label{tab:ablation_hirid_mort_mort_gmm_regularization-type}
            \begin{tabular}{cc}
                \toprule
                Parameter & AUPRC \\
                \midrule
                hard & 0.380  \textpm \ 0.012 \\
                soft & 0.391  \textpm \ 0.025 \\
                \bottomrule
            \end{tabular}
        }\hfill
        \subtable[Regularization coefficient][t]{
            \label{tab:ablation_hirid_mort_mort_gmm_regularization-coefficient}
            \begin{tabular}{cc}
                \toprule
                Parameter & AUPRC \\
                \midrule
                0.0000 & 0.385  \textpm \ 0.011 \\
                0.0001 & 0.391  \textpm \ 0.007 \\
                0.0010 & 0.378  \textpm \ 0.025 \\
                0.1000 & 0.377  \textpm \ 0.012 \\
                \bottomrule
            \end{tabular}
        }\hfill
        \subtable[Regularization coefficient][t]{
            \includegraphics[width=0.3\linewidth]{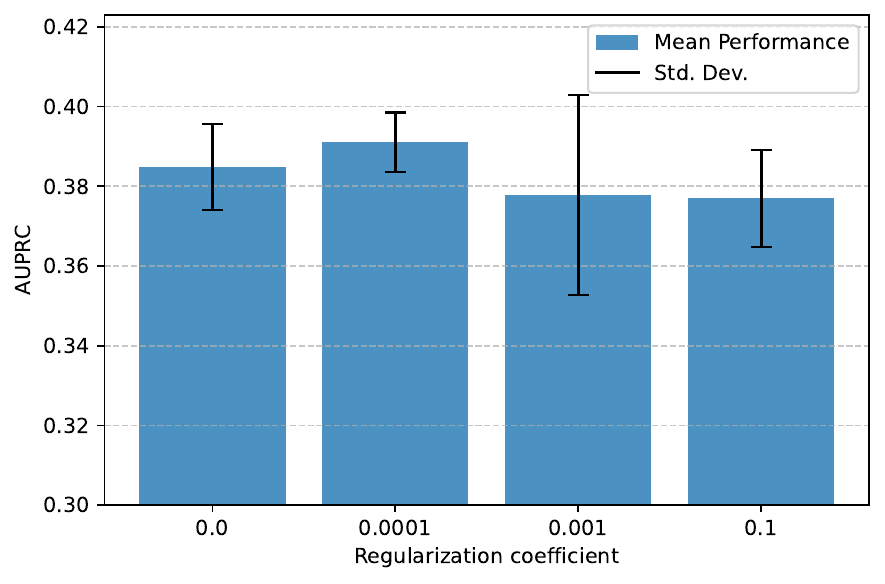}
        }
        
        \subtable[EMA rate][t]{
            \label{tab:ablation_hirid_mort_mort_gmm_ema-rate}
            \begin{tabular}{cc}
                \toprule
                Parameter & AUPRC \\
                \midrule
                0.00 & 0.385  \textpm \ 0.012 \\
                0.05 & 0.373  \textpm \ 0.006 \\
                0.25 & 0.381  \textpm \ 0.011 \\
                0.50 & 0.378  \textpm \ 0.025 \\
                0.75 & 0.374  \textpm \ 0.007 \\
                0.95 & 0.391  \textpm \ 0.003 \\
                \bottomrule
            \end{tabular}
        }\qquad
        \subtable[EMA rate][t]{
            \includegraphics[width=0.3\linewidth]{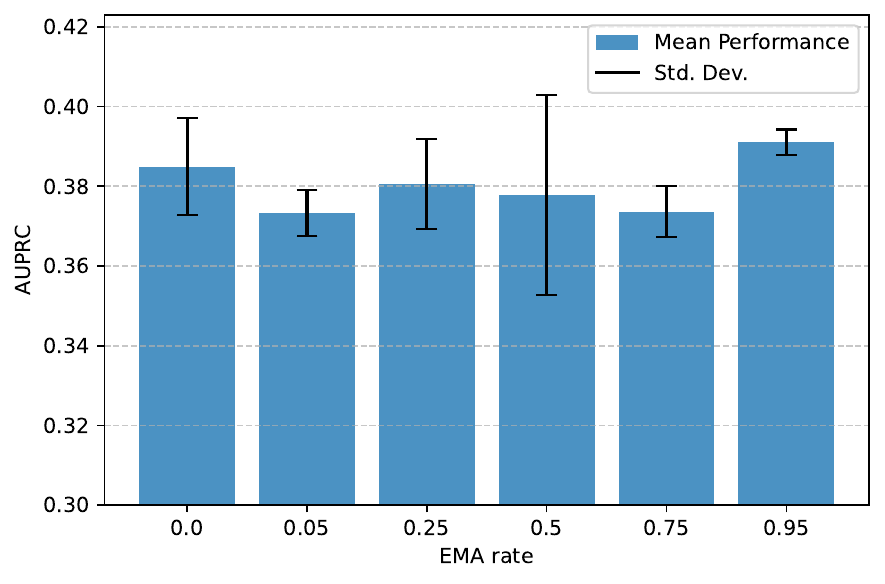}
        }
        
        \subtable[Number of clusters][t]{
            \label{tab:ablation_hirid_mort_mort_gmm_number-of-clusters}
            \begin{tabular}{cc}
                \toprule
                Parameter & AUPRC \\
                \midrule
                2 & 0.385  \textpm \ 0.025 \\
                3 & 0.381  \textpm \ 0.012 \\
                4 & 0.377  \textpm \ 0.009 \\
                5 & 0.391  \textpm \ 0.007 \\
                8 & 0.381  \textpm \ 0.011 \\
                \bottomrule
            \end{tabular}
        }\qquad
        \subtable[Number of clusters][t]{
            \includegraphics[width=0.3\linewidth]{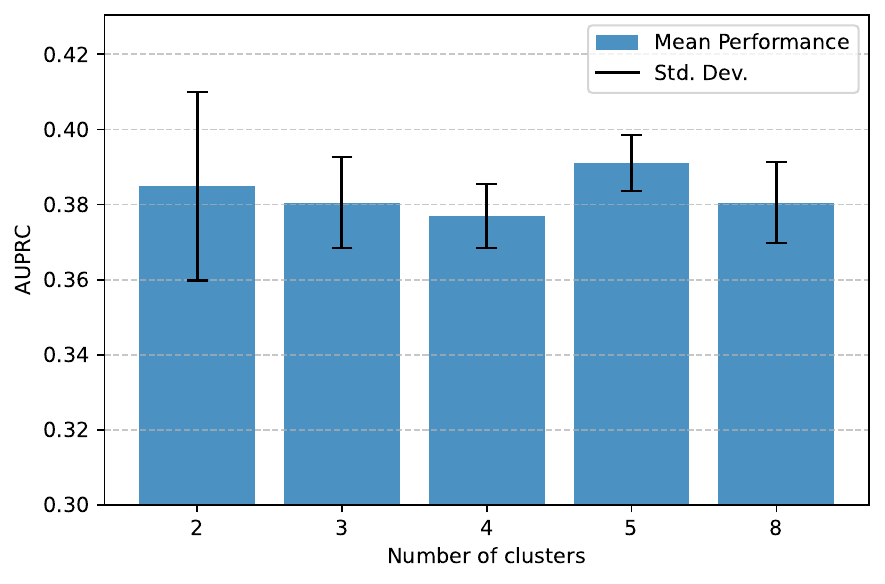}
        }
        
        \subtable[Covariance type][t]{
            \label{tab:ablation_hirid_mort_mort_gmm_covariance-type}
            \begin{tabular}{cc}
                \toprule
                Parameter & AUPRC \\
                \midrule
                diagonal  & 0.381  \textpm \ 0.011 \\
                full      & 0.391  \textpm \ 0.012 \\
                spherical & 0.385  \textpm \ 0.025 \\
                tied      & 0.381  \textpm \ 0.004 \\
                \bottomrule
            \end{tabular}
        }
    \end{table*}

    \begin{table*}[htbp]
        \centering
        \caption{Ablation results over various parameters for K-means trained on MIMIC for mortality prediction.}
        \setlength{\parskip}{10pt}
        \vspace{5pt}
        \label{tab:ablation_mimic_mort_kmeans}
        \subtable[Initialization][t]{
            \label{tab:ablation_mimic_mort_kmeans_initialization}
            \begin{tabular}{cc}
                \toprule
                Parameter & AUPRC \\
                \midrule
                k-means\texttt{++} & 0.525  \textpm \ 0.106 \\
                prior    & 0.516  \textpm \ 0.068 \\
                \bottomrule
            \end{tabular}
        }\hfill
        \subtable[Unification][t]{
            \label{tab:ablation_mimic_mort_kmeans_unification}
            \begin{tabular}{cc}
                \toprule
                Parameter & AUPRC \\
                \midrule
                bias & 0.520  \textpm \ 0.014 \\
                bias\_avg\_linear & 0.520  \textpm \ 0.015 \\
                bias\_ext\_catzero & 0.503  \textpm \ 0.068 \\
                bias\_sum\_linear & 0.525  \textpm \ 0.106 \\
                \bottomrule
            \end{tabular}
        }\hfill
        \subtable[Merge][t]{
            \label{tab:ablation_mimic_mort_kmeans_merge}
            \begin{tabular}{cc}
                \toprule
                Parameter & AUPRC \\
                \midrule
                attention & 0.515  \textpm \ 0.012 \\
                mean      & 0.525  \textpm \ 0.106 \\
                \bottomrule
            \end{tabular}
        }
        
        \subtable[Regularization type][t]{
            \label{tab:ablation_mimic_mort_kmeans_regularization-type}
            \begin{tabular}{cc}
                \toprule
                Parameter & AUPRC \\
                \midrule
                hard & 0.520  \textpm \ 0.015 \\
                soft & 0.525  \textpm \ 0.106 \\
                \bottomrule
            \end{tabular}
        }\hfill
        \subtable[Regularization coefficient][t]{
            \label{tab:ablation_mimic_mort_kmeans_regularization-coefficient}
            \begin{tabular}{cc}
                \toprule
                Parameter & AUPRC \\
                \midrule
                0.0000 & 0.525  \textpm \ 0.014 \\
                0.0001 & 0.515  \textpm \ 0.012 \\
                0.0010 & 0.520  \textpm \ 0.068 \\
                0.1000 & 0.498  \textpm \ 0.106 \\
                \bottomrule
            \end{tabular}
        }\qquad
        \subtable[Regularization coefficient][t]{
            \includegraphics[width=0.3\linewidth]{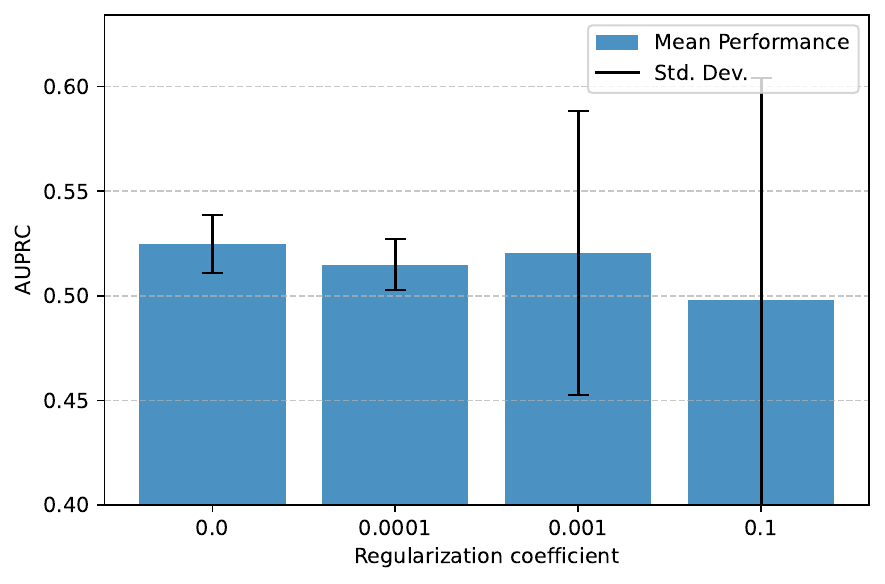}
        }
        
        \subtable[EMA rate][t]{
            \label{tab:ablation_mimic_mort_kmeans_ema-rate}
            \begin{tabular}{cc}
                \toprule
                Parameter & AUPRC \\
                \midrule
                0.00 & 0.520  \textpm \ 0.014 \\
                0.05 & 0.497  \textpm \ 0.009 \\
                0.25 & 0.525  \textpm \ 0.006 \\
                0.50 & 0.508  \textpm \ 0.015 \\
                0.75 & 0.515  \textpm \ 0.013 \\
                0.95 & 0.520  \textpm \ 0.106 \\
                \bottomrule
            \end{tabular}
        }\qquad
        \subtable[EMA rate][t]{
            \includegraphics[width=0.3\linewidth]{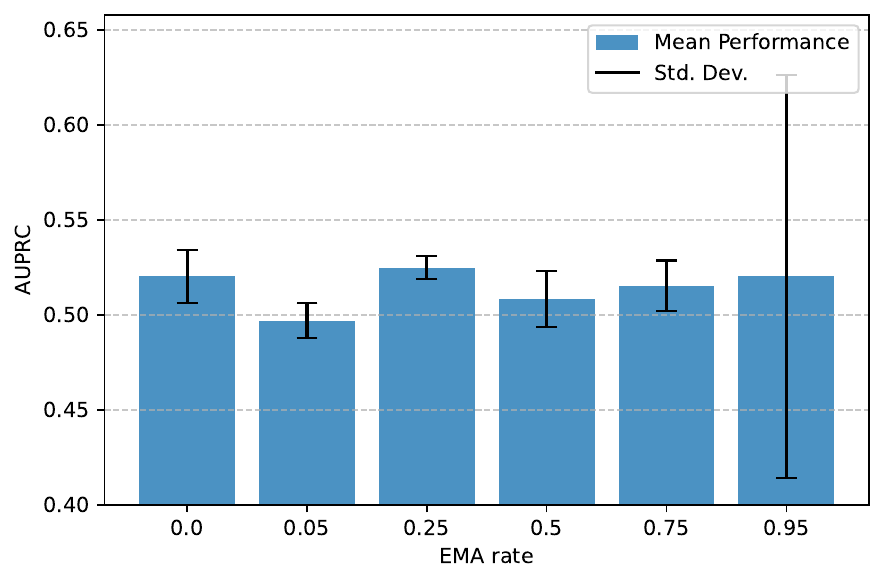}
        }
        
        \subtable[Number of clusters][t]{
            \label{tab:ablation_mimic_mort_kmeans_number-of-clusters}
            \begin{tabular}{cc}
                \toprule
                Parameter & AUPRC \\
                \midrule
                3 & 0.516  \textpm \ 0.106 \\
                5 & 0.520  \textpm \ 0.031 \\
                8 & 0.525  \textpm \ 0.014 \\
                \bottomrule
            \end{tabular}
        }\qquad
        \subtable[Number of clusters][t]{
            \includegraphics[width=0.3\linewidth]{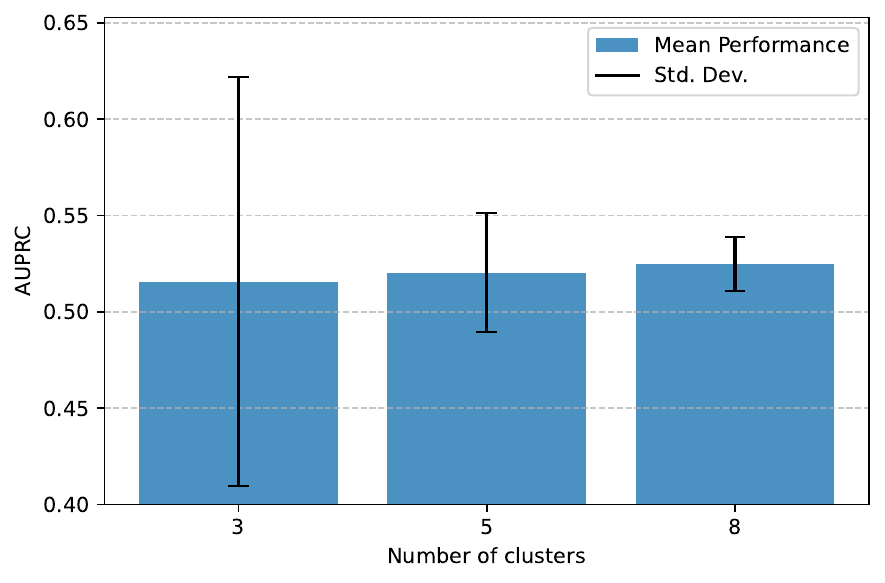}
        }
    \end{table*}

\end{document}